\documentclass[lettersize,journal]{IEEEtran}

\usepackage{subcaption}
\usepackage{adjustbox}
\usepackage{array}
\usepackage{multirow}
\usepackage{booktabs}
\usepackage{makecell}
\usepackage{amsmath,amsfonts}
\usepackage{algorithmic}
\usepackage{algorithm}
\usepackage{textcomp}
\usepackage{stfloats}
\usepackage{url}
\usepackage{verbatim}
\usepackage{graphicx}
\usepackage{cite}
\usepackage{xspace}
\usepackage{xcolor}
\usepackage[caption=false,font=normalsize,labelfont=sf,textfont=sf]{subfig}

\newcommand{\ie}{\emph{i.e.,}\xspace}

\begin{document}

\title{FLOL: Fast Baselines for Real-World Low-Light Enhancement}

\author{Juan C. Benito, Daniel Feijoo, Alvaro Garcia, Marcos V. Conde,~\IEEEmembership{Member,~IEEE,}\\
Cidaut AI, Spain 
\thanks{This paper was produced by Cidaut AI, Valladolid, Spain.}
}

\markboth{IEEE Transactions on Image Processing, January~2026}%
{Shell \MakeLowercase{\textit{et al.}}: A Sample Article Using IEEEtran.cls for IEEE Journals}


\maketitle

\begin{abstract}
Low-Light Image Enhancement (LLIE) is a key task in computational photography and imaging. The problem of enhancing images captured during night or in dark environments has been well-studied in the computer vision literature. However, current deep learning-based solutions struggle with efficiency and robustness for real-world scenarios (e.g., scenes with noise, saturated pixels). We propose a lightweight neural network that combines image processing in the frequency and spatial domains. Our baseline method, FLOL, is one of the fastest models for this task, achieving results comparable to the state-of-the-art on popular real-world benchmarks such as LOLv2, LSRW, MIT-5K and UHD-LL. Moreover, we are able to process 1080p images in real-time under 12ms. The code is available at \url{https://github.com/cidautai/FLOL}
\end{abstract}

\begin{IEEEkeywords}
Image Processing, Low-light Enhancement, Efficiency, Neural Networks
\end{IEEEkeywords}

\section{Introduction}
\label{sec:intro}

\IEEEPARstart{T}{he} performance of imaging systems can be disrupted in low-light conditions. For instance, detection and action recognition algorithms might struggle in conditions of low illumination, under-exposure, extreme noise, and especially at night~\cite{yoshimura2023dynamicisp, yoshimura2023rawgment, onzon2021neural, xu2023toward}. Therefore, it is essential in image processing to find a reliable solution that allows to improve exposure and ``lighten'' dark images correctly. The image restoration problem~\cite{elad1997restoration} for night or dark images could be formulated as:
\vspace{-1mm}
\begin{equation}\label{eq:problem}
    \mathbf{y} = \gamma (\mathbf{x}) + \mathbf{n}
\end{equation}

where $\mathbf{y}$ is the captured image, $\mathbf{x}$ is the underlying clean image, $\gamma$ is a function for the response of the camera sensor (\emph{e.g.} ISO gain, clip saturated pixels), and $\mathbf{n}$ is the sensor read-shot noise --- especially strong during night due to the lack of photon readings by the camera sensor~\cite{hasinoff2014photon}. For simplification, as previous works, we will ignore the possible blur effects and other optics-related artifacts such as glare and flare.

\begin{figure*}[h!]
  \centering
  \begin{subfigure}{0.49\textwidth}
    \includegraphics[width=\linewidth, height=4.85cm]{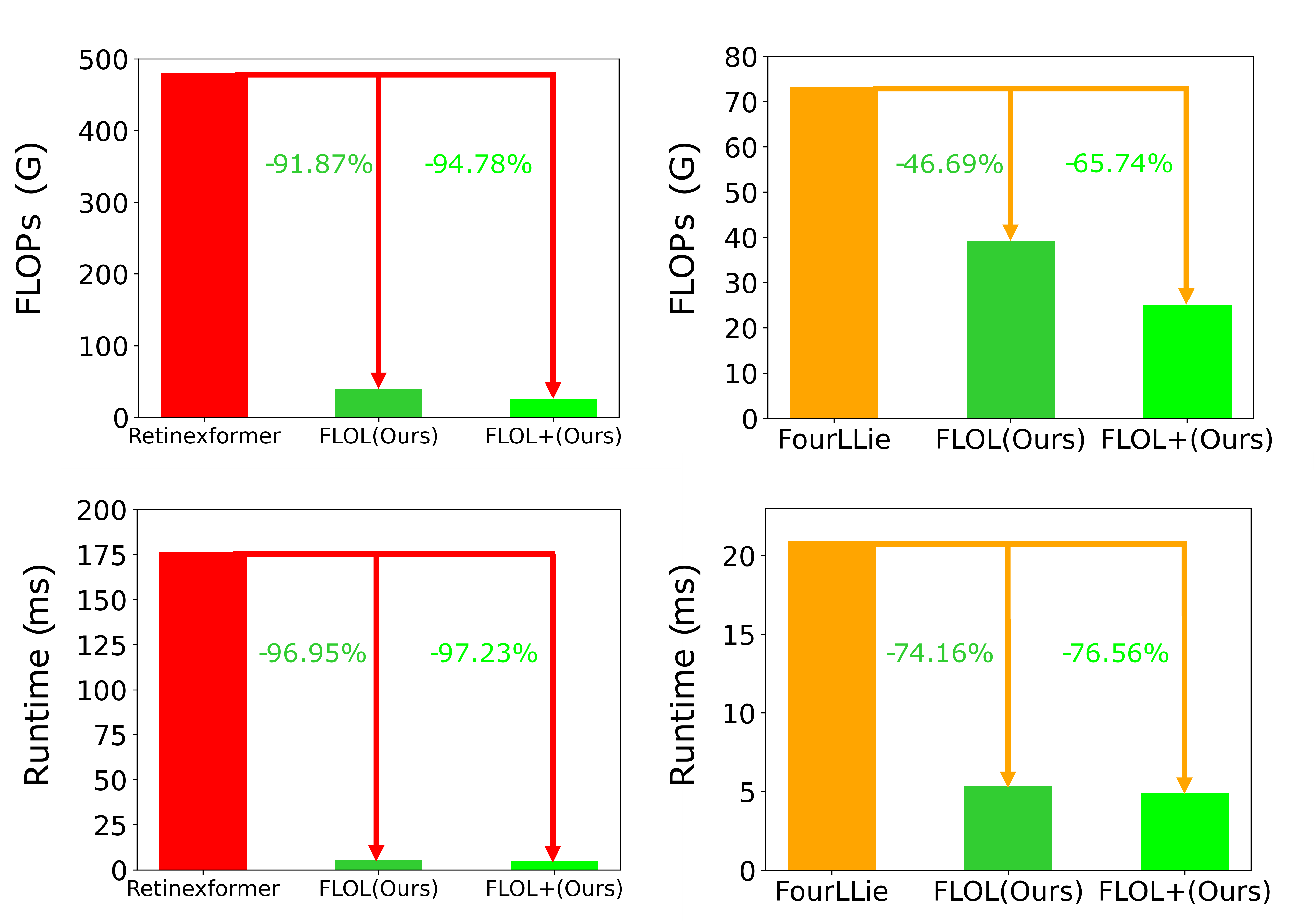}
  \end{subfigure}
  \hfill
  \begin{subfigure}{0.5\textwidth}
    \includegraphics[width=\linewidth, height=4.85cm]{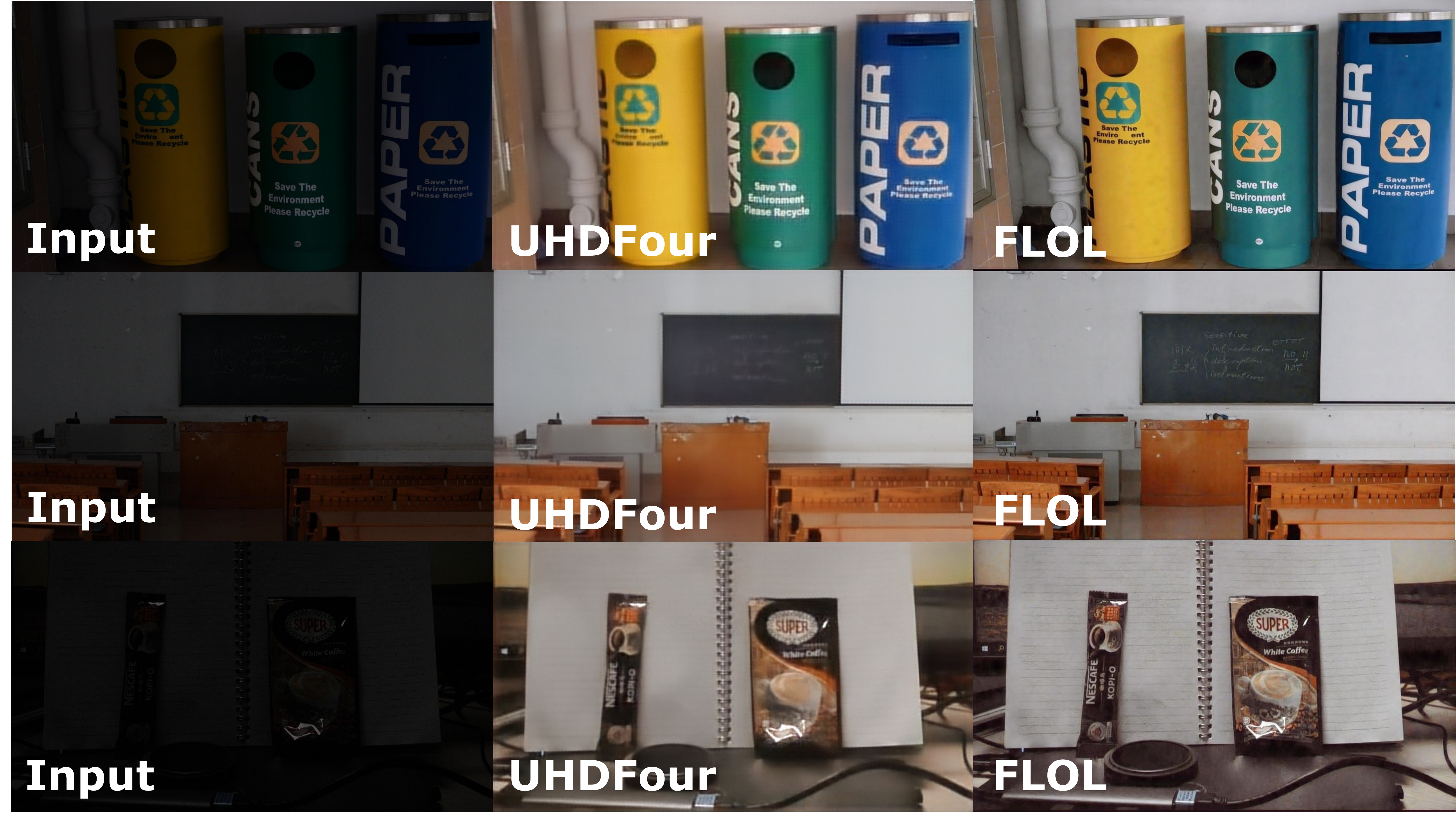}
  \end{subfigure}
  \caption{(Left) \textbf{Efficiency comparison} -FLOPs operations and runtime- of our solutions \emph{FLOL} and \emph{FLOL+} against Retinexformer~\cite{Cai_2023_ICCV} and FourLLIE~\cite{wang2023fourllie}. We calculate metrics using HD $1280\times720$ resolution. Our method achieves similar performance while requiring $7\times$ less operations and processing images \textbf{in less than 6 ms}. (Right) We show qualitative results of our method \emph{FLOL} compared with UHDFour~\cite{uhdll} on the UHD-LL~\cite{uhdll} dataset.}
  \label{fig:teaser}
\end{figure*}

Diverse techniques have been proposed to solve this inverse problem \emph{i.e.} reconstruct $\mathbf{x}$ from $\mathbf{y}$, which are classified into two groups: deep learning-based techniques \cite{retinex_4, retinex_2, retinex_net, wang2023fourllie} and classical techniques \cite{videodeeplearning2021, kind, retinex}. Nowadays, deep learning methods are widely considered the state-of-the-art for this task~\cite{Cai_2023_ICCV, sci, GLARE}. Most deep learning solutions based on convolutional neural networks (CNNs) and Transformers~\cite{vaswani2017attention,Cai_2023_ICCV} showcase great complexity in terms of parameters and operations to achieve an acceptable reconstruction performance. As a result, most neural methods lack \textbf{efficiency} --memory and runtime--, and they cannot enhance images in real-time {\emph{i.e.}} 50 ms per image or 20 FPS, for instance Retinexformer~\cite{Cai_2023_ICCV} requires 175 ms. We show these efficiency limitations in Fig. \ref{fig:teaser}.

\paragraph{\textbf{Defining real-world low-light enhancement.}}
Besides \textbf{efficiency} \ie process images under realistic GPU and time resources, low-light image enhancement solutions should \textbf{generalize} in real scenes (``in the wild''). We find several real and synthetic datasets for this task such as LOLv2~\cite{lol_v2}, and it is well-known that models trained on synthetic data fail to generalize on real cases~\cite{abdelhamed2018high, real_blur}. The reason is the domain gap between the simulated conditions and the real-world conditions; most approaches assume $\mathbf{n}$ to follow a Gaussian distribution, and $\gamma$ to be a linear function, however, the camera sensor's noise and response are more complex. LLIE models must be robust to out-of-distribution (OOD) conditions~\cite{sci}. Therefore, we focus on solving these two problems in LLIE: efficiency and robustness in real scenarios. 

Previous works such as RetinexNet~\cite{retinex_net}, UHDFour~\cite{uhdll}, FourLLIE~\cite{wang2023fourllie} and SNR-Net~\cite{snr_net}, aim to build efficient models using \emph{two components}: first, a frequency component to estimate and process the fourier amplitude of the image; second, the spatial denoising component, where the model will focus on improving details and reducing noise. This allows to tackle directly the two degradation components in the inverse problem (Eq.~\ref{eq:problem}) \ie the illumination-related $\gamma$ and the noise $\mathbf{n}$.

\noindent\textbf{Our contributions can be summarized as follows}: (1) We introduce a new lightweight and fast baseline model called \textbf{FLOL} for low-light image enhancement -see Fig.~\ref{fig:teaser}-; (2)  we achieve \emph{performance comparable to state-of-the-art (SOTA) methods} on real-world low-light enhancement benchmarks; and (3) our method requires $10\times$ less parameters than others and can process Full-HD images in \underline{real-time} at $12$ ms on regular GPUs.

\section{Related Work}

\paragraph{Low-Light Image Enhancement.}{The principal task in LLIE consists in correcting the illumination level of dark and under-exposed images properly, without adding artifacts or noise. Traditional methods considered histogram equalization~\cite{abdullah2007, ibrahim2007} or gamma correction~\cite{huang2013, rahman2016adaptive} but, most recent solutions are based on CNNs which are widely used as an end-to-end solution to this problem by implementing effective neural network architectures or new color spaces~\cite{youonlyneed}. For example, Dong \textit{et al.}~\cite{dong2025towards} introduced SG-LLIE that presents a scale-aware CNN-Transformer guided by structure priors. Other methods apply traditional image decomposition like RetinexNet \cite{retinex_net} or neural implicit representations (NIR) like CoLIE~\cite{Chobola2024}. They are capable of reaching higher quality images by estimating illumination and reflectance maps with help of retinex theory. In addition, Jiang \textit{et al.}~\cite{Jiang_2024_ECCV} also employed retinex decomposition and diffusion models in tandem in order to create a new method named LightenDiff. However, Retinexformer~\cite{Cai_2023_ICCV} is the most competitive retinex-derived method ---if we focus on balance between performance and efficiency--- based on attention mechanisms such as the Transformer~\cite{vaswani2017attention}. Further, Xu\textit{ et al.}~\cite{snr_net} introduced a method that pays attention to the \textit{Signal-to-Noise-Ratio} (SNR) in the dark (noisy) images. This makes possible to enhance ``pixel by pixel'' regions with lower SNR associated with dimmer regions of the image. Moreover, Shi \textit{et al.}~\cite{shi2024zero} used previous concepts (retinex theory and SNR) in a zero-shot model called Zero-IG to enhance images individually without need of pre-training. Another example is GLARE~\cite{GLARE}, which is a heavy model that employs generative latent features, but also it requires a long processing time as it introduces a huge amount of parameters and FLOPs~\cite{yan2025hvi}. Other procedures handle unsupervised learning, focusing on the illumination curves estimation or the improvement of illumination learning to perform a good light correction~\cite{zero_reference, sci}.

There are recent models that exploit Fourier information to solve LLIE. For example, Li \textit{et al.}~\cite{Li2023ICLR} studies how to extract lightness and noise in the Fourier space. They noticed that it is beneficial to incorporate space and frequency information into neural networks. Also, Huang\textit{ et al.}~\cite{deepfourier} and Li \textit{et al.}~\cite{uhdll} suggested before that illumination information is contained within the amplitude component, thus, this component differs from low to high images and it is related to the lightness level on them. In consequence, Wang\emph{ et al.} leveraged this idea in FourLLIE ~\cite{wang2023fourllie}. 
}

\paragraph{Fourier Theory.}Fourier information has been proved to be helpful in LLIE problem. Firstly, the Fourier theory was considered on extracting noise and lightness by using neural networks in the Fourier space~\cite{uhdll}. Also, Huang \textit{et al.}~\cite{deepfourier} and again, Li \textit{et al.}~\cite{uhdll} suggested that illumination information is contained within the amplitude component, thus, this component differs from low to high images. Another example is Fuoli\emph{ et al.}~\cite{fourierlosses2021}, which also presented a special loss based on Fourier theory that attempts to restore high-frequency information in super-resolution problems. Finally, Wang\emph{ et al.}~\cite{wang2023fourllie} applied Fourier frequency information to give a LLIE solution. They introduced a new Fourier-based network called FourLLIE which includes the phase and amplitude Fourier images along with the SNR map introduced by Xu\emph{ et al.}~\cite{snr_net}. We investigate further to design a method capable of processing high-resolution images faster and demanding less memory, all without losing performance.

\begin{figure*}[t]
    \centering
    \includegraphics[width=\textwidth]{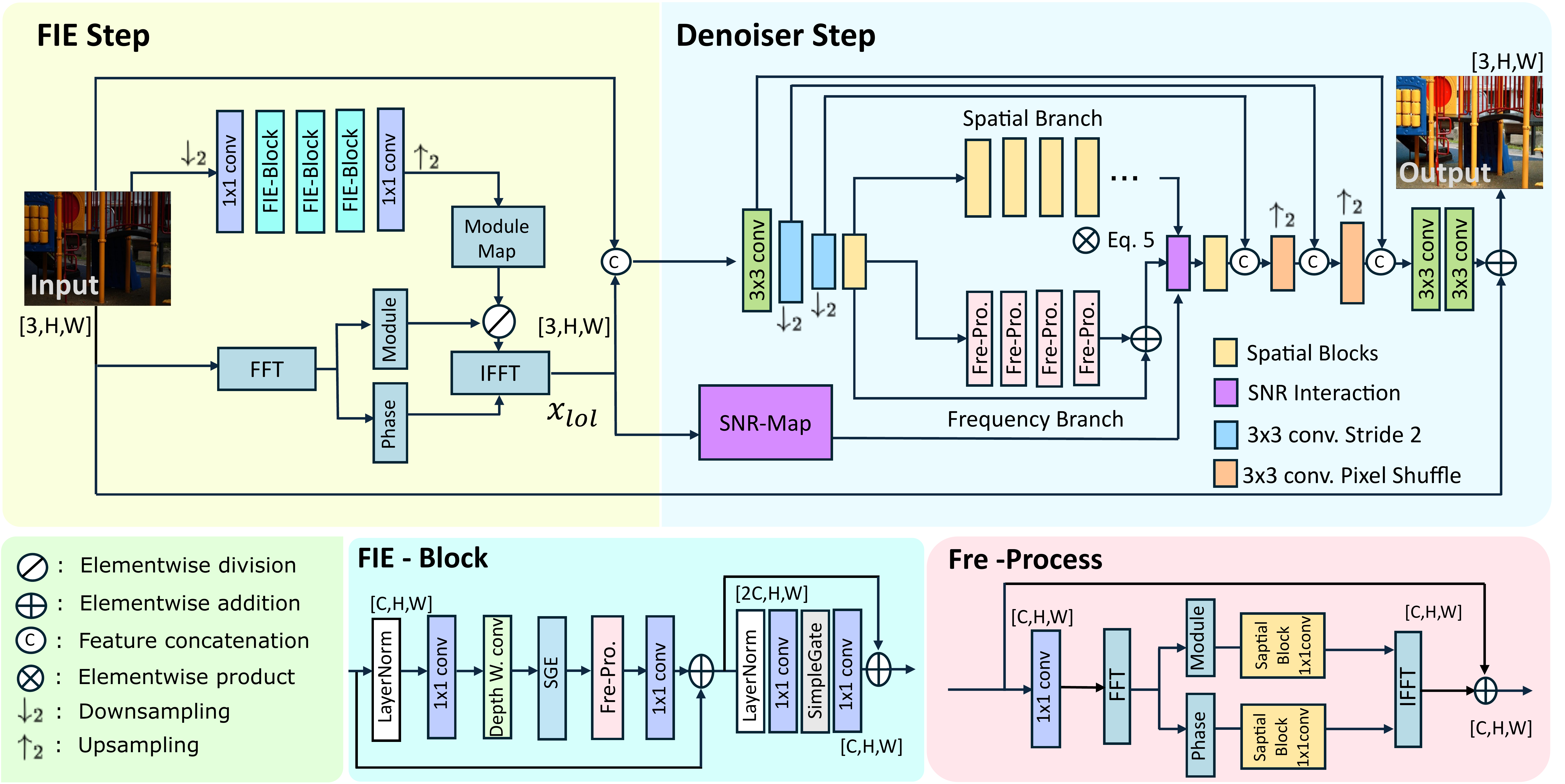}
    \caption{Overview of the base model \textbf{FLOL}. 
    In the first step, the input image is transformed into the Fourier frequency domain to enhance its illumination level. Next, during the denoiser step, we correct imperfections such as artifacts and high levels of noise. Note that our model is trained end-to-end.}
    \label{fig:scheme}
\end{figure*}
\section{Proposed Method}

Given an input image $x$ of size $(H,W)$, the discrete Fourier transform can be calculated following the relation:

\begin{equation}
    X(u,v) = \frac{1}{\sqrt{HW}} \sum_{h=0}^{H-1} \sum_{w=0}^{W-1} x(h,w) e^{-i2\pi \left(\frac{h}{H}u + \frac{w}{W}v\right)}
\end{equation}

with $h,w$ the coordinates in the spatial domain, $u,v$ the coordinates in the frequency domain and $i$ is the imaginary unit. $X(u,v)$ are complex numbers composed of a real part, $R(X(u,v))$, and an imaginary part, $I(X(u,v))$. We can put these complex values either in the binomial form or in the polar form depending on its amplitude and phase components.

As a result of other Fourier-based procedures \cite{deepfourier, uhdll, wang2023fourllie, lv2024fourier}, we know the following properties in the context of LLIE. First, the information on lightness is contained within the \emph{amplitude} component in the frequency domain. Secondly, Fourier-based models are capable of extracting \emph{global information} contained in the input image without introducing a large amount of parameters. This fact is important if we want to run our model on low memory devices.

\begin{figure*}[t]
    \centering
    \setlength{\tabcolsep}{1pt} 
    \begin{tabular}{c c c c c c}

    \includegraphics[width=0.165\linewidth, height=2cm]{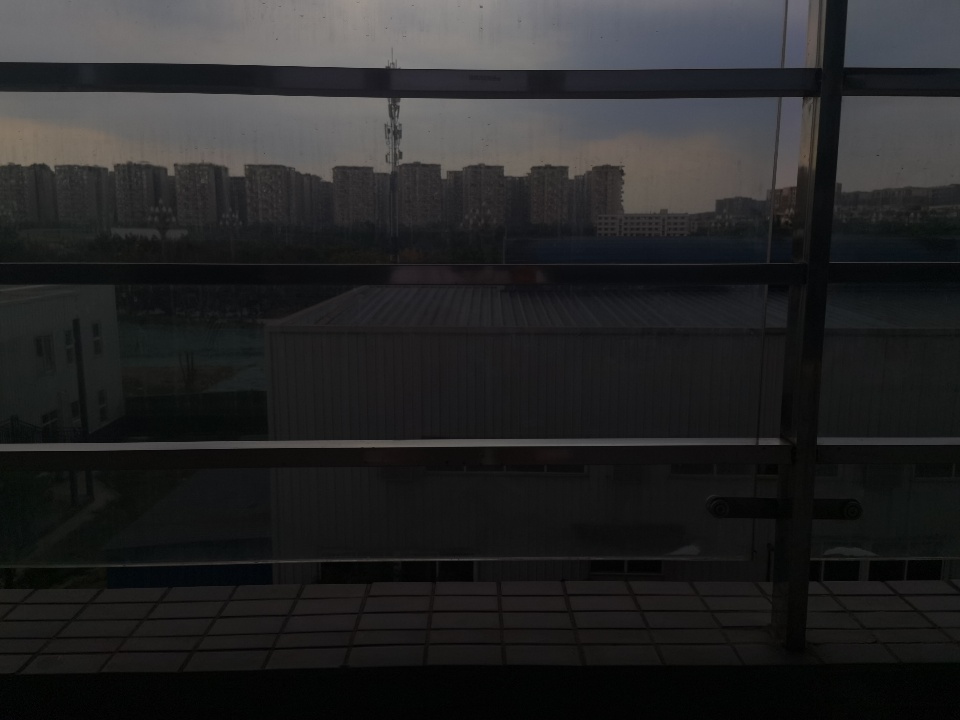} &

    \includegraphics[width=0.165\linewidth, height=2cm]{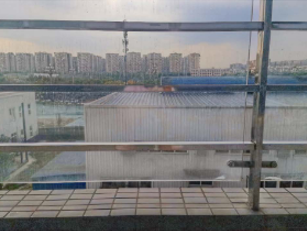} &

    \includegraphics[width=0.165\linewidth, height=2cm]{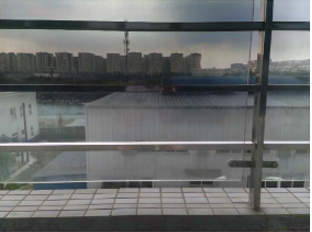} &

    \includegraphics[width=0.165\linewidth, height=2cm]{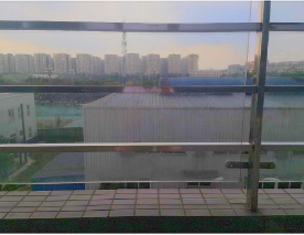} &

    \includegraphics[width=0.165\linewidth, height=2cm]{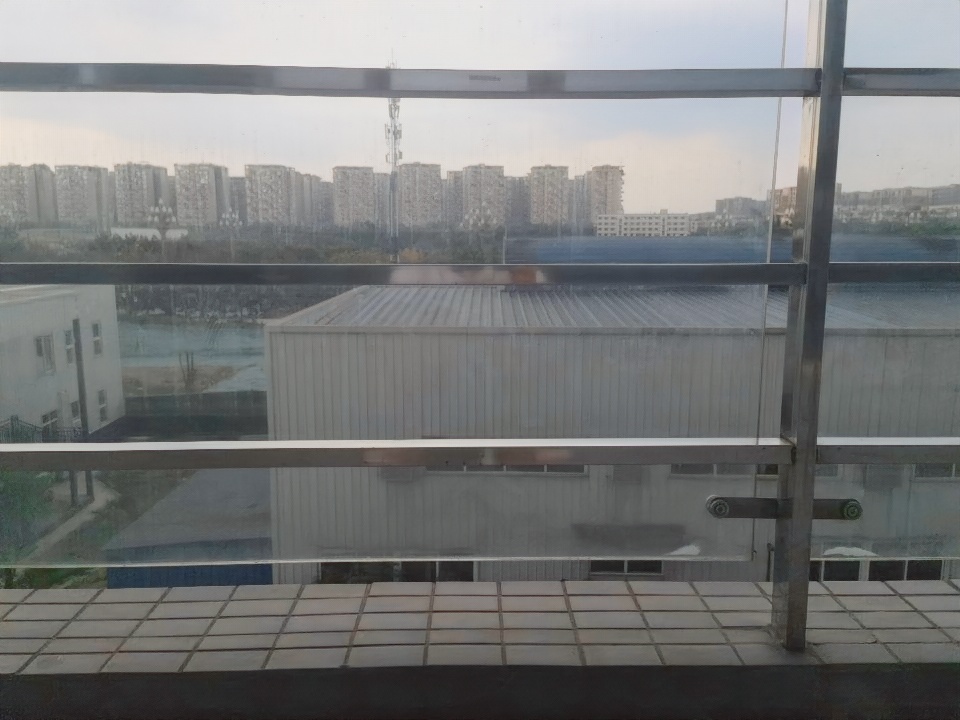} &

    \includegraphics[width=0.165\linewidth, height=2cm]{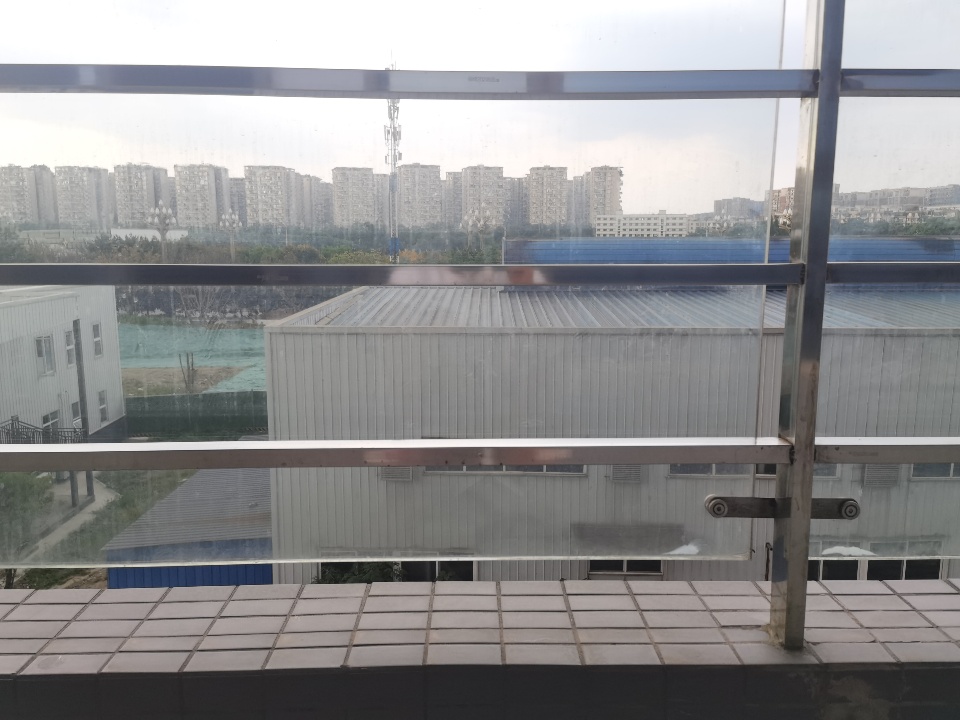} \\

    Input & KinD~\cite{kind} & DRBN~\cite{drbn} & EnGAN~\cite{enlightengan} & \bf{FLOL} & Ground Truth \\

    \includegraphics[width=0.165\linewidth, height=2cm]{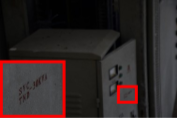} &

    \includegraphics[width=0.165\linewidth, height=2cm]{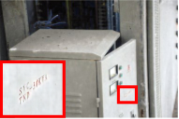} &

    \includegraphics[width=0.165\linewidth, height=2cm]{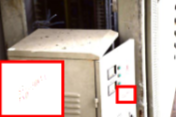} &

    \includegraphics[width=0.165\linewidth, height=2cm]{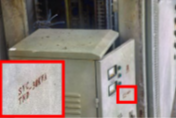} &

    \includegraphics[width=0.165\linewidth, height=2cm]{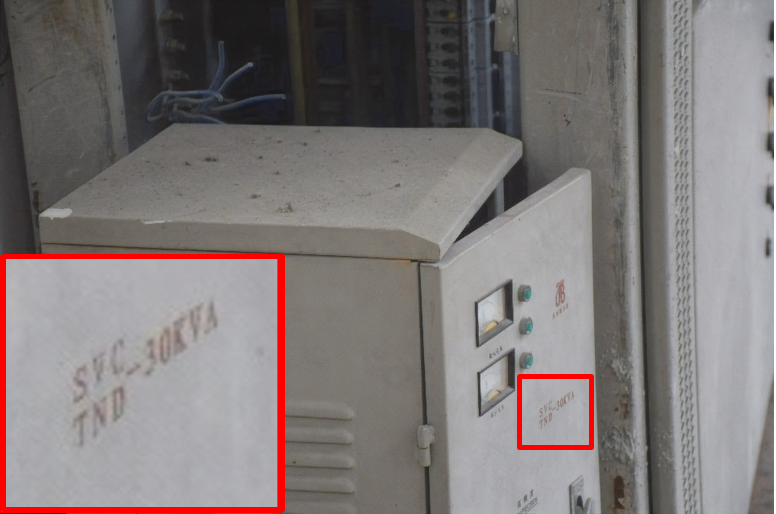} &

    \includegraphics[width=0.165\linewidth, height=2cm]{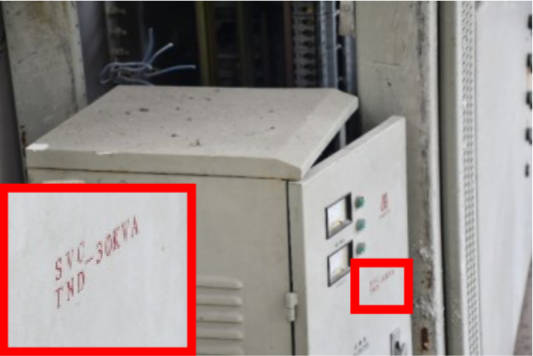} \\

    Input & MIRNet~\cite{mirnet} & RUAS~\cite{ruas} & EnGAN~\cite{enlightengan} & \bf{FLOL} & Ground Truth \\

    \end{tabular}
    \caption{Qualitative results on \textbf{LSRW-Huawei} \cite{hai2023r2rnet} (first row) and \textbf{LSRW-Nikon} \cite{hai2023r2rnet} (second row).}
    \label{fig:LSRW}
\end{figure*}

\begin{table*}[t]
\begin{center} 
\begin{adjustbox}{width=1\textwidth,center}
\centering
\begin{tabular}{c|c|c|c|c}
\Xhline{1.5pt}
Dataset &Real/Synth &Training Pairs &Testing Pairs &Notes\\
\Xhline{1.5pt}
LOLv2-Real~\cite{lol_v2} &Real &689 &100 &Images collected by
changing exposure and ISO \\

LOLv2-Synthetic~\cite{lol_v2} &Synth &900 &100 &Images obtained from RAW images \\ 

LOLv1~\cite{retinex_net} &Real &485 &15 &Images collected by
changing exposure and ISO\\

LSRW-Nikon~\cite{hai2023r2rnet} &Real &3150 &20 &Images captured with a DSLM Nikon camera\\

LSRW-Huawei~\cite{hai2023r2rnet} &Real &2450 &30 &Images obtained with a Huawei smartphone\\

UHD-LL~\cite{uhdll} &Real &2000 &150 &UHD Images collected with Sony $\alpha$7 III and Sony Alpha a6300 cameras\\

MIT-5K~\cite{mitfivek} &Real &4500 &100 &Images captured with SLR cameras\\
\Xhline{1.5pt}
\end{tabular}
\end{adjustbox}
\end{center}
\caption{Main specifications of paired datasets used for training and testing.}
\label{tab:paired_datasets}
\end{table*}

Hence, we design our end-to-end model with two different blocks (see Fig. \ref{fig:scheme}). The first step is the \emph{Fourier Illumination Enhancement (FIE)}, which increases the illumination level and estimates the clear image $\mathbf{x}$ from the dark input $\mathbf{y}$, by enhancing its amplitude component in the Fourier frequency space. The output of this part is $\mathbf{x}_{lol}$, and serves as the input of the next component. The second step, the \emph{Denoiser}, uses an SNR map~\cite{snr_net} to tackle noise, color imperfections and artifacts, emphasizing details recovery. The output is the reconstructed clean image $\hat{\mathbf{x}}$.


\subsection{Fourier Illumination Enhancement (FIE) Step}
In the first step, the input image is converted into Fourier space using the discrete 2-dimensional Fast Fourier Transform (FFT), and then separated into its phase and module components. Following previous works~\cite{uhdll, wang2023fourllie}, we focus on enhancing the amplitude component. To achieve this, we produce the amplitude transform map (Module Map) by feeding the input image into the \emph{FIE-Block}. This block follows a Metaformer~\cite{yu2022metaformer} structure: 

\vspace{-2mm}
\begin{align}
    z_1 = \text{Attention}(\text{LayerNorm} (z)) + z \\
    z_2 = \text{FFN}(\text{LayerNorm}(z_1)) + z_1
\end{align}

where $z$ are the input features and $z_2$ the output features of the block. In contrast, we apply global attention in the frequency domain, thanks to the Fourier properties. For this, we design the \emph{Fre-Process} block that works inside the FIE-Block --- a feed-forward network that operates in the frequency components (please see Fig. \ref{fig:scheme}). 
We use the same gated FFN (pixel-wise convolutions with simple gates) as NAFNet~\cite{chen2022simple}. Note that we estimate the Module Map using a low-resolution input image resized by half ($\downarrow_2$) using bilinear interpolation. This is possible since the amplitude is (partially) scale invariant~\cite{uhdll}. Moreover, this allows us to reduce the computation and the number of operations notably. The resultant amplitude map is upscaled to the original resolution using the same interpolation. We fix the number of channels of the latent features at 16. 

Finally, we continue by applying the Module Map to the original amplitude ---via element-wise division--- to enhance it. The result is the new amplitude component of the input image, and the non-modified phase. We apply an inverse Fast Fourier Transform (iFFT) to obtain the intermediate result $\mathbf{x}_{lol}$ (please see Fig. \ref{fig:scheme}). This result already presents good illumination properties, but also notable noise and artifacts. In the section ahead, we explain how to remove this noise and artifacts and how to perform a good color correction. This allows us to properly recover the clean image. The output obtained at the end of the FIE step is flawed as we mentioned above. To solve this problem, we concatenate this output and the input image and then, the result is fed into the Denoiser step. 

\begin{figure*}[t]
    \centering
    \setlength{\tabcolsep}{1pt} 
    \begin{tabular}{c c c c c c c}

    \includegraphics[width=0.16\linewidth, height = 3.5cm]{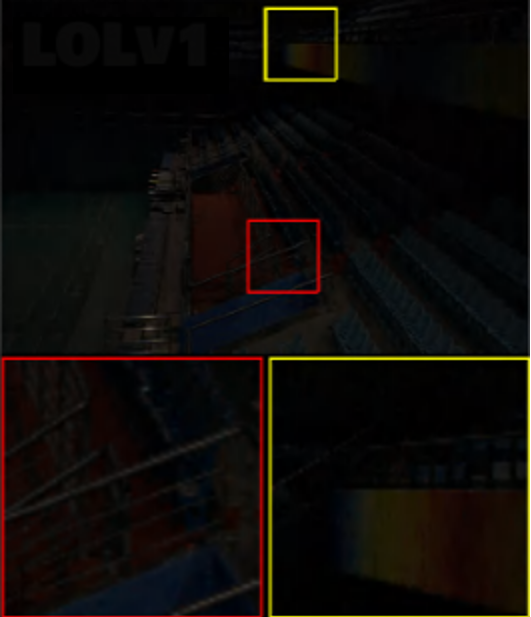} &

    \includegraphics[width=0.16\linewidth, height = 3.5cm]{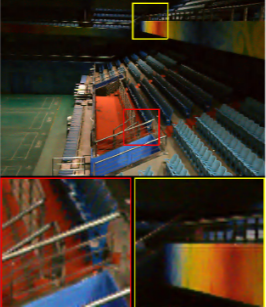} &

    \includegraphics[width=0.16\linewidth, height = 3.5cm]{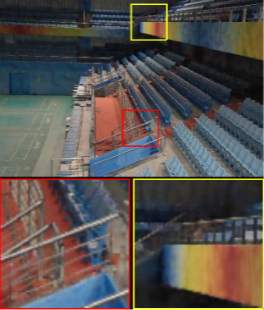} &

    \includegraphics[width=0.16\linewidth, height = 3.5cm]{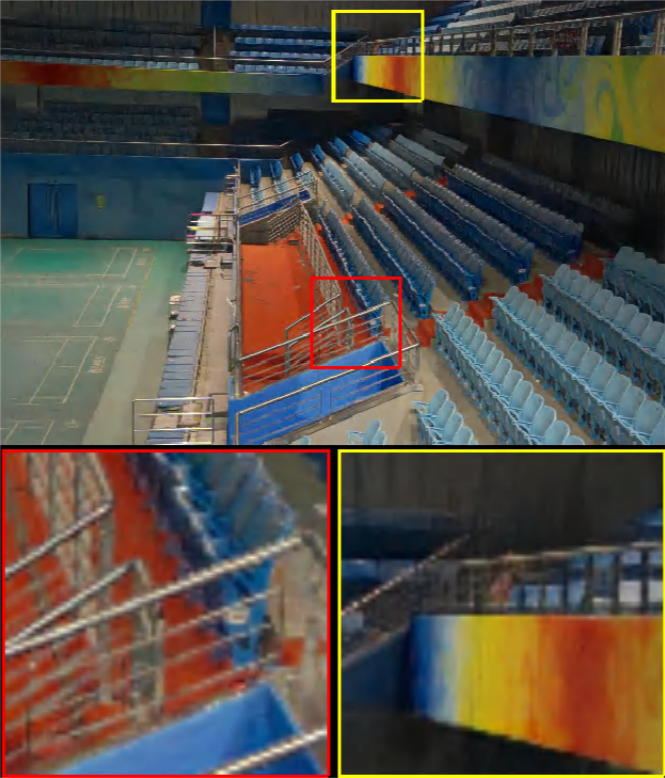} &

    \includegraphics[width=0.16\linewidth, height = 3.5cm]{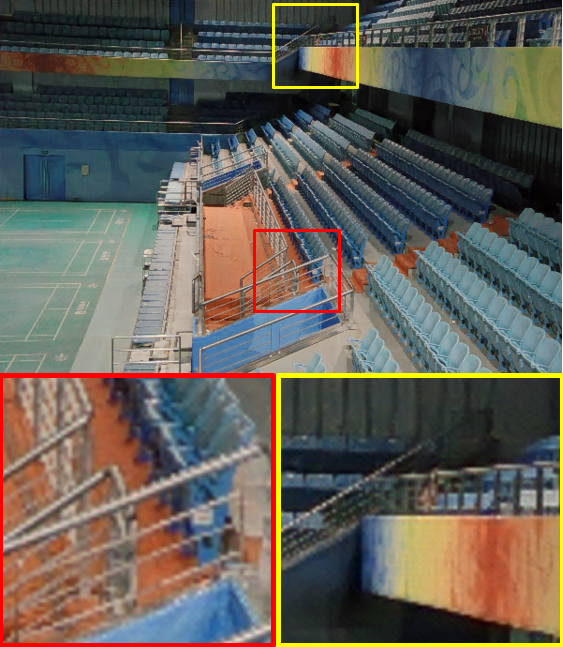} &

    \includegraphics[width=0.16\linewidth, height = 3.5cm]{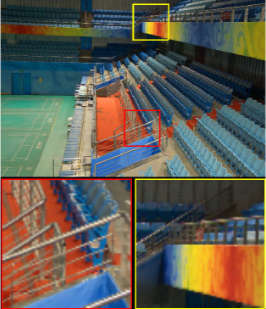} \\

    Input & RUAS~\cite{ruas} & RetFormer~\cite{Cai_2023_ICCV} & CIDNet~\cite{youonlyneed} & \bf{FLOL} &  Ground Truth \\

    \includegraphics[width=0.16\linewidth, height = 2cm]{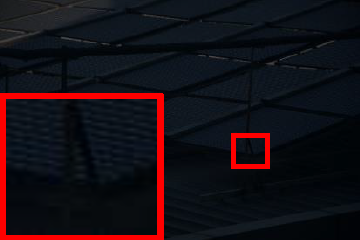} &

    \includegraphics[width=0.16\linewidth, height = 2cm]{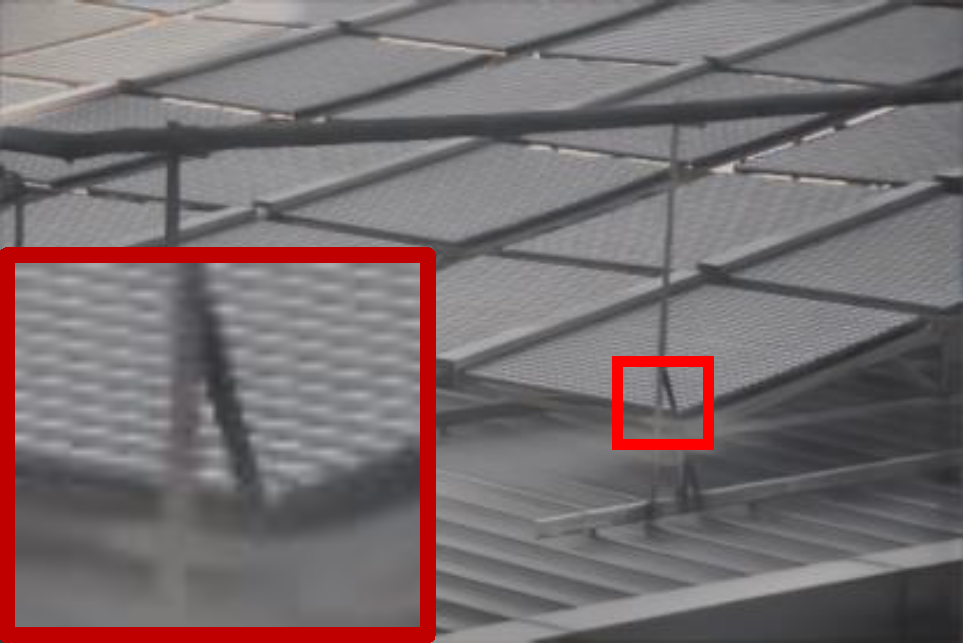} &

    \includegraphics[width=0.16\linewidth, height = 2cm]{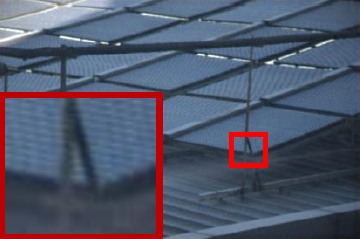} &

    \includegraphics[width=0.16\linewidth, height = 2cm]{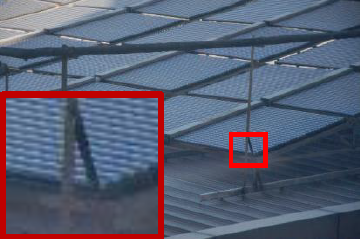} &

    \includegraphics[width=0.16\linewidth, height = 2cm]{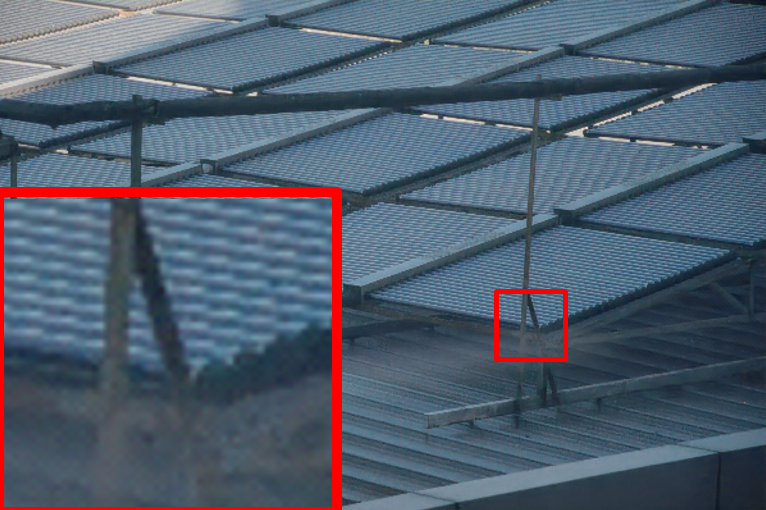} &

    \includegraphics[width=0.16\linewidth, height = 2cm]{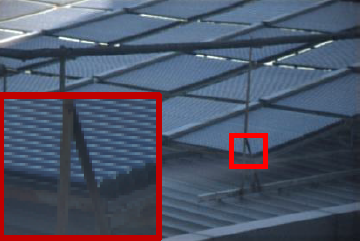} \\

    Input & Bread~\cite{guo2023low} & MIRNet~\cite{mirnet}  & FourLLIE~\cite{wang2023fourllie} & \bf{FLOL} &  Ground Truth
    
    \end{tabular}
    \caption{Qualitative results on \textbf{LOLv1}~\cite{retinex_net} (first row) and \textbf{LOLv2-Real}~\cite{lol_v2} (second row) datasets.}
    \label{fig:LOL}
\end{figure*}

\begin{table*}[]
\begin{center} 
\begin{adjustbox}{width=\textwidth,center}
\centering
\begin{tabular}{c|ccc|c|cc|c|cc}
\Xhline{1.5pt} 
\multicolumn{4}{c|}{LSRW} &\multicolumn{3}{c|}{UHD-LL} &\multicolumn{3}{c}{MIT-5K}\\
\hline
Methods &PSNR$\uparrow$ &SSIM$\uparrow$ & LPIPS$\downarrow$ &Methods &PSNR$\uparrow$ &SSIM$\uparrow$ &Methods &PSNR$\uparrow$ &SSIM$\uparrow$ \\
\Xhline{1.5pt}
RetinexNet~\cite{retinex_net} &15.906 &0.3725 &0.393  &Uformer~\cite{uformer} &19.283 &0.849 &RetinexNet~\cite{retinex_net} &13.74 &0.739\\

EnGAN~\cite{enlightengan} &16.311 &0.4697 &0.322 &Restormer~\cite{restormer} &22.597 &0.878 &EnGAN~\cite{enlightengan} &16.76 &0.834 \\ 

FIDE~\cite{fide} &\textcolor{blue}{17.669} &\textcolor{blue}{0.5485} &-  &SNR-Net(r)~\cite{snr_net} &22.717 &0.877 &FIDE~\cite{fide} &17.19 &0.785\\

ZDCE~\cite{zero_reference} &15.834 &0.4664 &\textcolor{blue}{0.315} &ZDCE~\cite{zero_reference} &17.075 &0.663 &ZDCE~\cite{zero_reference} &16.61 &0.814\\

DRBN~\cite{drbn}  &16.149 &0.5422 &0.376  &SNR-Net(s)~\cite{snr_net}  &22.170 &0.866 &DRBN~\cite{drbn}  &17.59 &0.784\\

RUAS\cite{ruas} &14.437 &0.4276 &0.455 &RUAS\cite{ruas} &13.562 &0.749 &RUAS\cite{ruas} &18.53 &0.864\\

KinD~\cite{kind} &16.472 &0.4929 &- &UHDFour~\cite{uhdll} &\textcolor{red}{26.226} &\textcolor{red}{0.900} &KinD~\cite{kind} &17.09 &0.830 \\

SCI~\cite{sci} &15.017 &0.4846 &0.321 &SCI~\cite{sci} &16.057 &0.625 &SCI~\cite{sci} &\textcolor{blue}{20.44} &\textcolor{blue}{0.893} \\

STAR~\cite{xu2020star} &14.608 &0.5039 &- &Restormer(s)~\cite{restormer} &22.252 &0.871 &STAR~\cite{xu2020star} &17.64 &0.779\\
\hline
\bf{FLOL(Ours)} &\textcolor{red}{19.10} &\textcolor{red}{0.5833} &\textcolor{red}{0.273} &\bf{FLOL(Ours)} &\textcolor{blue}{25.01} &\textcolor{blue}{0.888}&\bf{FLOL(Ours)} &\textcolor{red}{22.10} &\textcolor{red}{0.910}\\
\Xhline{1.5pt}
\end{tabular}
\end{adjustbox}
\end{center}
\caption{Quantitative comparison of the retrained state of the art on the \textbf{LSRW}~\cite{hai2023r2rnet}, \textbf{UHD-LL}~\cite{uhdll}, \textbf{MIT-5K}~\cite{mitfivek} datasets. The best result is in \textcolor{red}{red} color, second best in \textcolor{blue}{blue}. All values are adopted from \cite{Yang_2023_ICCV, uhdll, sci}. Note that our method \textbf{FLOL} achieves notable results while having more than 180$\times$ \textbf{less parameters} than UHDFour. 
}
\label{tab:LSRWUHDLLMIT5K}
\end{table*}

\subsection{Denoiser Step}
{We focus on improving the previous component result, $\mathbf{x}_{lol}$. The $\mathbf{x}_{lol}$ and $\mathbf{y}$ are concatenated and fed into the denoising block. First, we use an encoder with simple strided $3\times3$ convolutions. Then, we employ the Fourier space (Frequency branch) and the spatial domain (Spatial branch) along with the SNR map to remove imperfections -- see Fig. \ref{fig:scheme}. We calculate the SNR map from the output of the frequency step in the same way as previous works~\cite{wang2023fourllie}. The outputs of these branches are $O_{S}$ and $O_{F}$, respectively. The SNR-based interaction, shown in Eq.~\ref{eq:10}, is the combination of $O_{S}$ and $O_{F}$ with the SNR map ($R$): 

\vspace{-1mm}
\begin{equation}
    \mathcal{F} = O_{S} \times R + O_{F} \times (1-R) .
\label{eq:10}
\end{equation}

The output features $\mathcal{F}$ are fed into the decoder of the Denoiser, together with the encoder skip connections. We use sub-pixel convolutions (Pixel Shuffle)~\cite{shi2016real} to up-sample the features in the decoder. Thereby, we decode the reconstructed output image $\hat{\mathbf{x}}$ and apply a global residual connection in which we add to the output image from the decoder the original image (using $\mathbf{x}$ as a prior). Finally, we obtain the clean image without noise or artifacts and good illumination levels.

\section{Experimental Results}
\label{sec:results}



\paragraph{Datasets.} We train and evaluate our method using well-known paired datasets for the LLIE problem such as LOLv1~\cite{retinex_net}, LOLv2-Real \cite{lol_v2}, LOLv2-Synthetic \cite{lol_v2}, LSRW \cite{hai2023r2rnet}, UHD-LL~\cite{uhdll}, MIT-5K~\cite{mitfivek}. Note that we arranged the paired datasets in Tab.~\ref{tab:paired_datasets} according to the number of training/testing split pairs; specifications about the cameras used to collect the images and distinction about if each dataset is real or synthetic. We also employed unpaired datasets such as BDD100k~\cite{bdd100k}, DICM~\cite{dicm}, LIME~\cite{lime}, MEF~\cite{mef}, NPE~\cite{npe}, VV~\cite{vv} and DarkFace~\cite{darkface}.

\begin{figure*}[t]

    \centering
    \includegraphics[width=\textwidth]{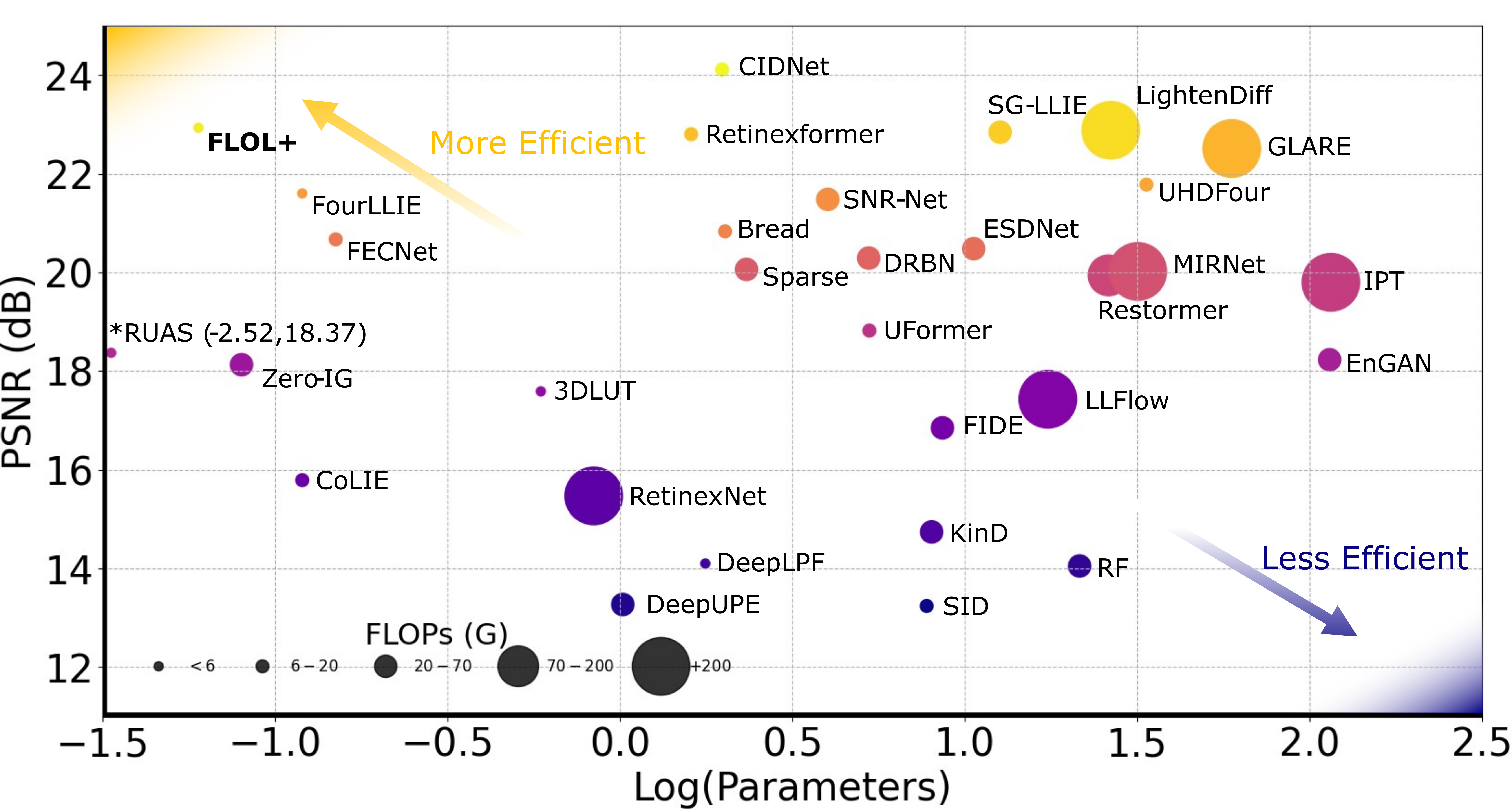}
    \caption{Comparison of our baseline model variant \textbf{FLOL+} efficiency and performance with other SOTA methods evaluated in the \textbf{LOLv2-Real}~\cite{lol_v2} dataset. We take the decimal logarithm of parameters in millions for each model to achieve better visualization. Note that RUAS~\cite{ruas} has very few parameters when compared with other methods. As a consequence, it is placed on y-axis since it is an outlier in the graph due to its low number of parameters. Even though CIDNet~\cite{youonlyneed} reaches higher values than \textbf{FLOL+} in PSNR, it lacks of efficiency in terms of FLOPs and number of parameters. All values are extracted from \cite{uhdll, Cai_2023_ICCV, wang2023fourllie, dong2025towards, yan2025hvi}.}
    \label{fig:ballgraphic}
   
\end{figure*}

\begin{table*}[]
\begin{center}
\begin{adjustbox}{width=\textwidth,center}
\begin{tabular}{c | c c c c c | c | c c c c c | c | c c c c c}
\Xhline{1.5pt}
 \multicolumn{6}{c|}{BRISQUE$\downarrow$} & \multicolumn{6}{c|}{MANIQA$\uparrow$} & \multicolumn{6}{c}{TReS$\uparrow$} \\
 
\hline

\multicolumn{1}{c|}{Methods} & \multicolumn{1}{c}{DICM} & \multicolumn{1}{c}{LIME} & \multicolumn{1}{c}{MEF} &\multicolumn{1}{c}{NPE} & \multicolumn{1}{c|}{VV}

&\multicolumn{1}{c|}{Methods} &  \multicolumn{1}{c}{DICM} & \multicolumn{1}{c}{LIME} &\multicolumn{1}{c}{MEF} &\multicolumn{1}{c}{NPE} &\multicolumn{1}{c|}{VV}

&\multicolumn{1}{c|}{Methods} &  \multicolumn{1}{c}{DICM} & \multicolumn{1}{c}{LIME} &\multicolumn{1}{c}{MEF} &\multicolumn{1}{c}{NPE} &\multicolumn{1}{c}{VV}\\

\Xhline{1.5pt}
KinD\cite{kind} & 48.72 & 39.91 & 49.94 & 36.85 & 50.56 & &  &  &  & &  & &  &  &  &  & \\

RUAS\cite{ruas} & 38.75 & 27.59 & \textcolor{red}{23.68} & 47.85 & 38.37 &Retinexformer~\cite{Cai_2023_ICCV} & \textcolor{blue}{0.31} & \textcolor{red}{0.3635} & \textcolor{red}{0.32} & \textcolor{blue}{0.33} & \textcolor{blue}{0.22} &Retinexformer~\cite{Cai_2023_ICCV} & \textcolor{blue}{60.53} & \textcolor{red}{67.51} & \textcolor{blue}{70.97} & \textcolor{blue}{69.03} &\textcolor{blue}{37.04} \\

LLFlow\cite{llflow} & 26.36 & \textcolor{blue}{27.06} &30.27 & 28.86 & 31.67 & &  &  &  &  &  & &  &  &  &  & \\

\cline{7-18}

SNR-Net\cite{snr_net} & 37.35 & 39.22 & 31.28 & 26.65 &78.72 &\multirow{4}{*}{\bf{FLOL(Ours)}} &\multirow{4}{*}{ \textcolor{red}{0.38}} &\multirow{4}{*}{\textcolor{blue}{0.3628}} &\multirow{4}{*}{\textcolor{blue}{0.30}} &\multirow{4}{*}{\textcolor{red}{0.41}} &\multirow{4}{*}{\textcolor{red}{0.39}} &\multirow{4}{*}{\textbf{FLOL(Ours)}} &\multirow{4}{*}{\textcolor{red}{63.90}} &\multirow{4}{*}{\textcolor{blue}{67.32}} &\multirow{4}{*}{\textcolor{red}{71.24}} &\multirow{4}{*}{\textcolor{red}{69.84}} &\multirow{4}{*}{\textcolor{red}{68.64}} \\

PairLIE\cite{pairlie} &33.31 &\textcolor{red}{25.23} &27.53 &28.27 &39.13 & & & & & & & &  &  &  &  & \\

Retinexformer\cite{Cai_2023_ICCV} & \textcolor{blue}{23.27} & 59.55 & \textcolor{blue}{24.91} & \textcolor{blue}{22.69} & \textcolor{blue}{24.89} & & &  &  & &  & & & &  &  & \\
\cline{1-6}
\bf{FLOL(Ours)} & \textcolor{red}{23.21} & \textbf{30.08} & \textbf{30.04} & \textcolor{red}{16.79} & \textcolor{red}{23.59} & & & & & & & & & & & & \\
\Xhline{1.5pt}
\end{tabular}
\end{adjustbox}
\end{center}
\caption{Quantitative comparison on five unpaired datasets using various IQA metrics, such as BRISQUE~\cite{brisque}, MANIQA~\cite{yang2022maniqa} or TReS~\cite{golestaneh2021tres}. BRISQUE values are adopted from \cite{youonlyneed}.}
\label{tab:unpaired}
\end{table*}

\begin{figure*}[]
    \centering
    \setlength{\tabcolsep}{1pt} 
    \begin{tabular}{c c c c c c c c c}

    \includegraphics[width=0.1\linewidth, height=2.2cm]{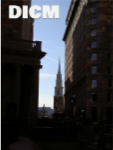} &

    \includegraphics[width=0.1\linewidth, height=2.2cm]{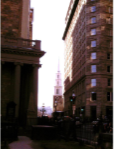} &

    \includegraphics[width=0.1\linewidth, height=2.2cm]{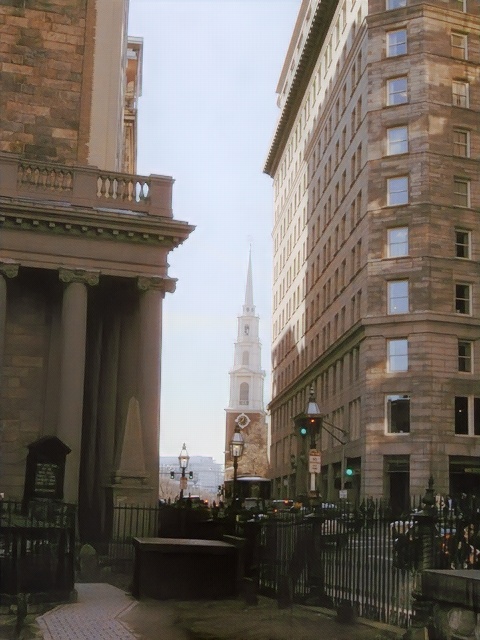} &

    \includegraphics[width=0.1\linewidth, height=2.2cm]{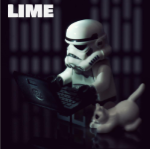} &

    \includegraphics[width=0.1\linewidth, height=2.2cm]{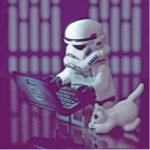} &

    \includegraphics[width=0.1\linewidth, height=2.2cm]{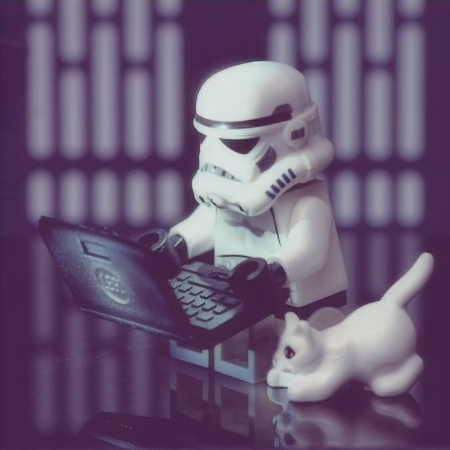}&

    \includegraphics[width=0.1\linewidth, height=2.2cm]{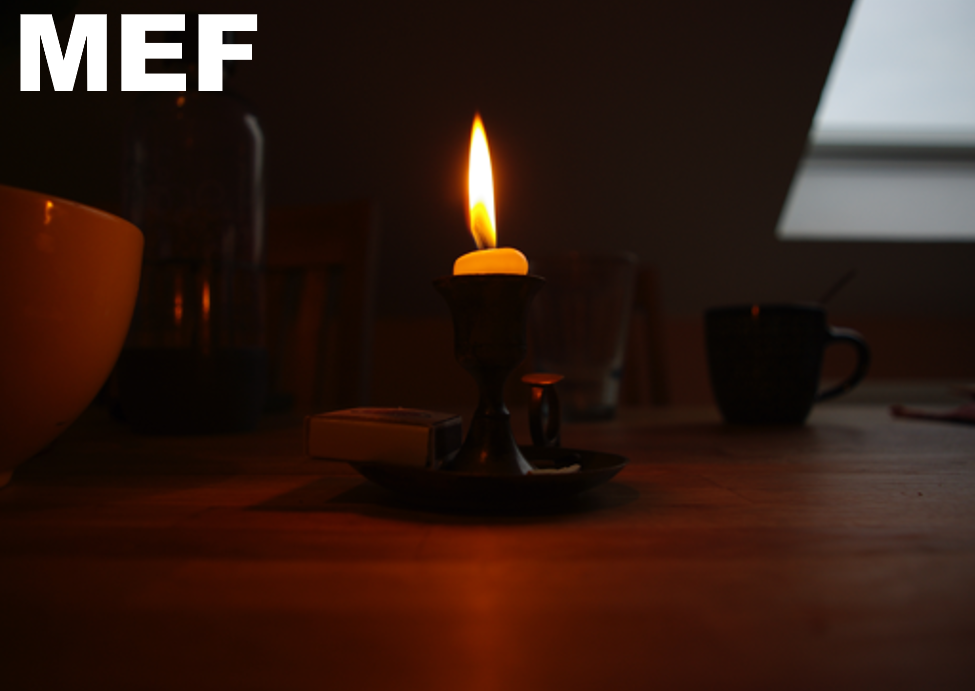} &

    \includegraphics[width=0.1\linewidth, height=2.2cm]{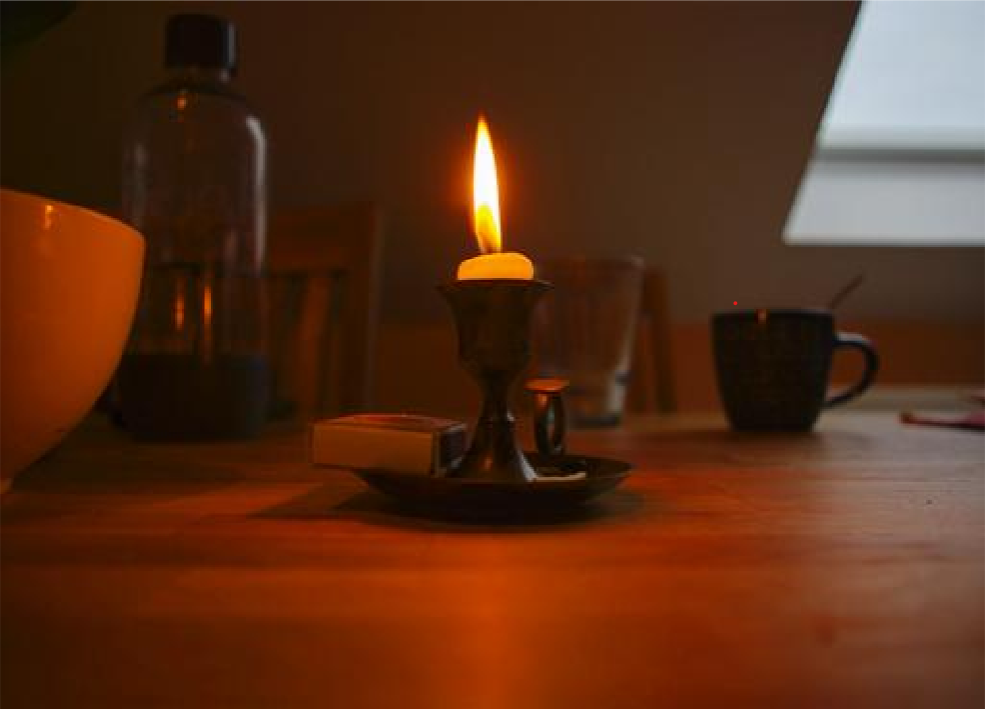} &

    \includegraphics[width=0.1\linewidth, height=2.2cm]{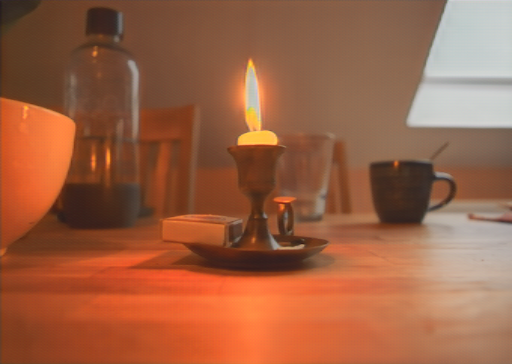} \\
    
    Input & \small RUAS~\cite{ruas} & \bf{FLOL} & Input & \small ZDCE~\cite{zero_reference} & \bf{FLOL} & Input & \small CIDNet~\cite{youonlyneed} & \bf{FLOL}

    \end{tabular}
     \setlength{\tabcolsep}{1pt} 
    \begin{tabular}{c c c c c c}

    \includegraphics[width=0.16\linewidth, height=2.2cm]{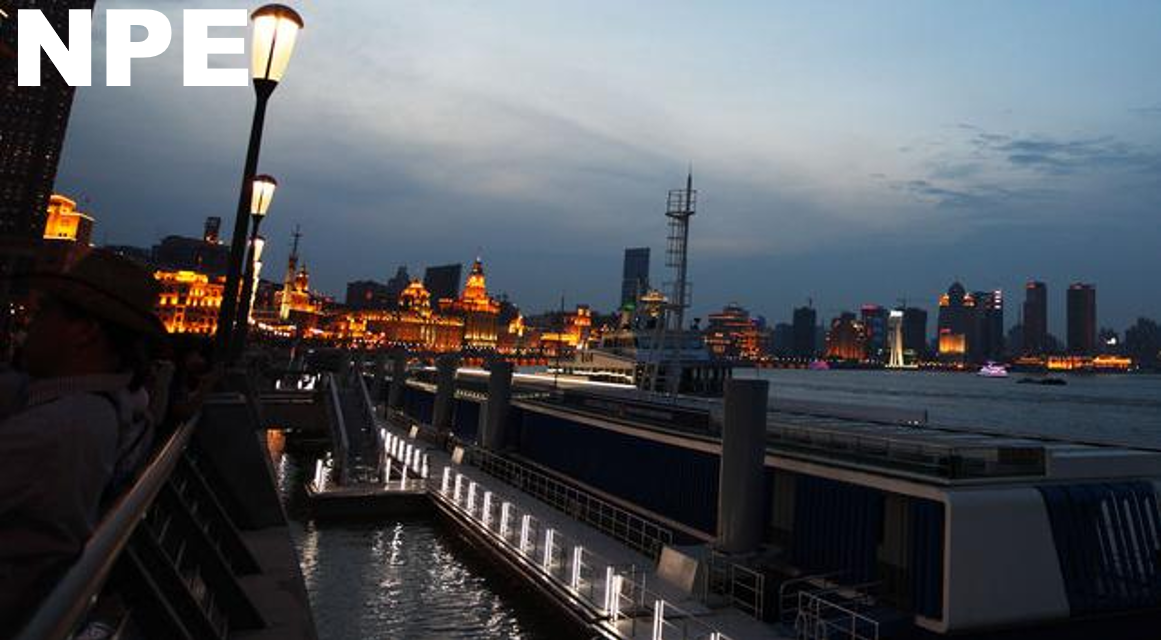} &

    \includegraphics[width=0.16\linewidth, height=2.2cm]{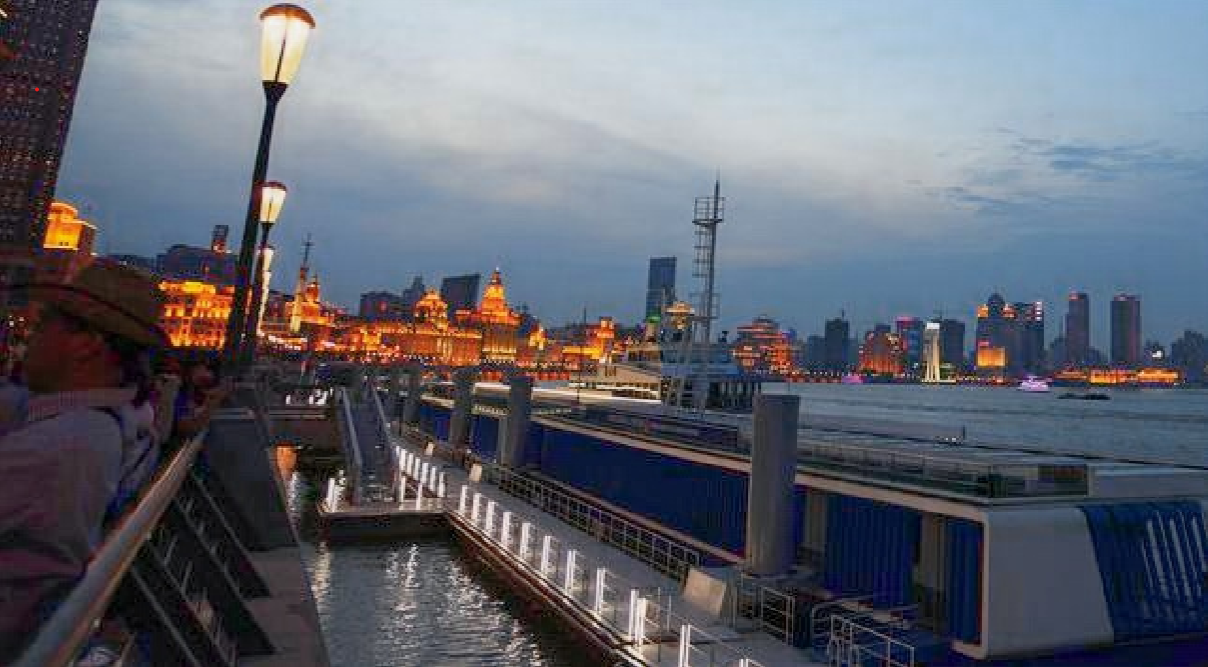}&

    \includegraphics[width=0.16\linewidth, height=2.2cm]{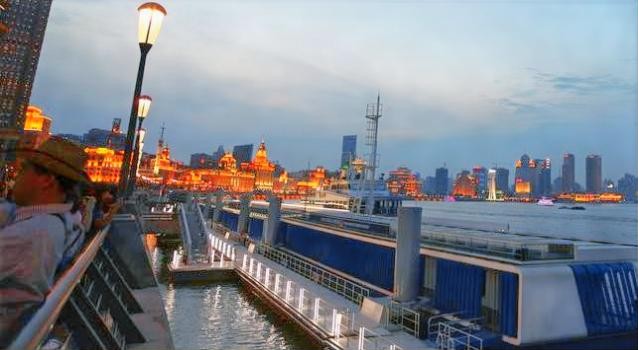} &

    \includegraphics[width=0.16\linewidth, height=2.2cm]{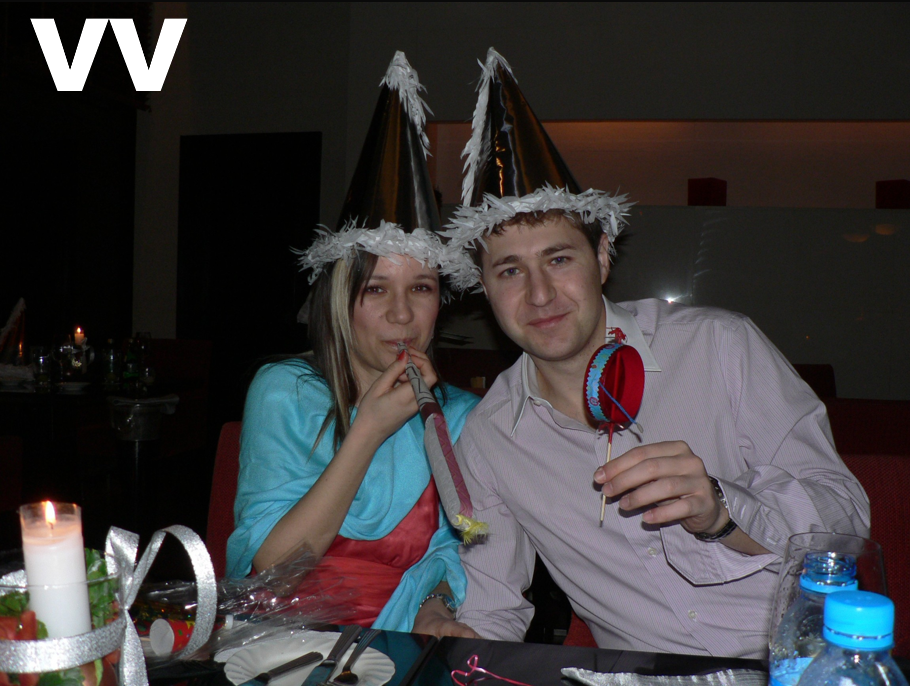} &

    \includegraphics[width=0.16\linewidth, height=2.2cm]{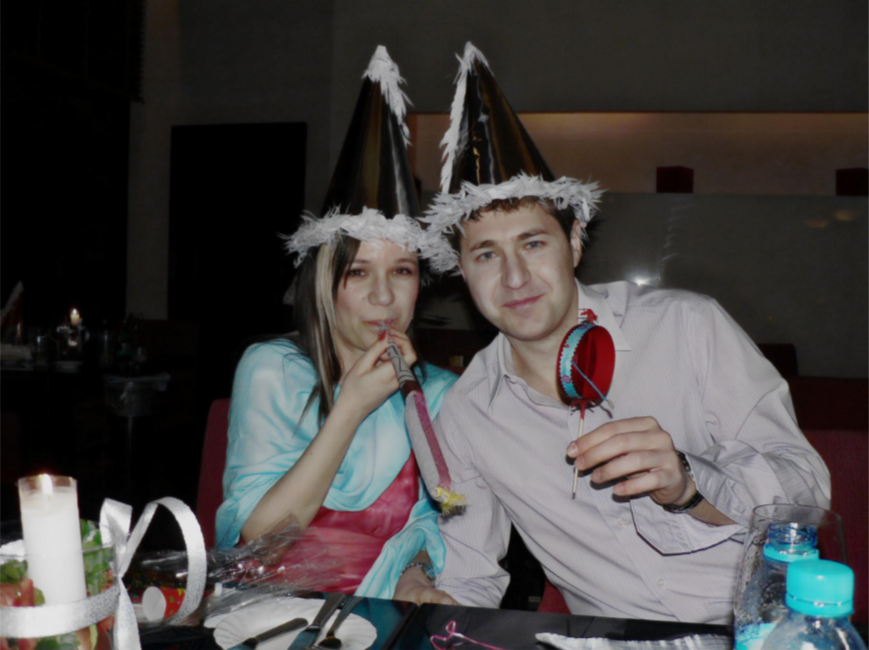} &

    \includegraphics[width=0.16\linewidth, height=2.2cm]{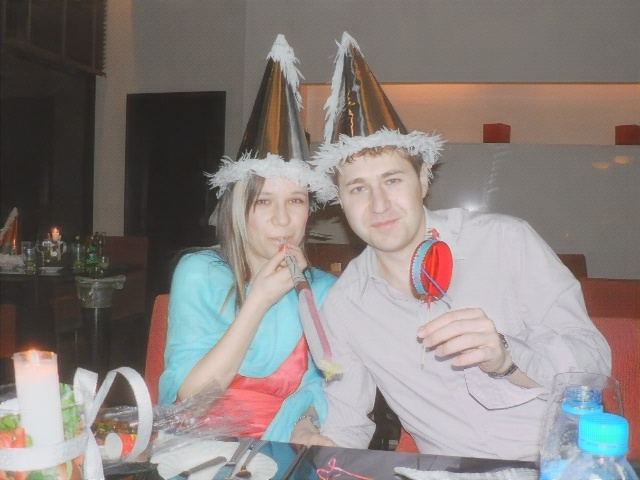} \\

     Input & Retinexformer~\cite{Cai_2023_ICCV} & \bf{FLOL} & Input & Kind~\cite{kind} & \bf{FLOL}
    
    \end{tabular}
    \caption{Qualitative results on the unpaired datasets, \textbf{DICM}~\cite{dicm}, \textbf{LIME}~\cite{lime}, \textbf{MEF}~\cite{mef}, \textbf{NPE}~\cite{npe} and \textbf{VV}~\cite{vv}, for different methods and \textbf{FLOL}. Zoom in for best view.}
    \label{fig:unpaired}
\end{figure*}

\begin{figure*}[t]
    \centering
    \setlength{\tabcolsep}{1pt} 
    \begin{tabular}{c c c c c c}

    \includegraphics[width=0.165\linewidth,height=2.15cm]{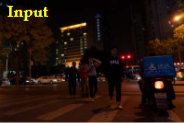} &

    \includegraphics[width=0.165\linewidth,height=2.15cm]{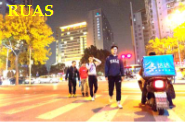} &

    \includegraphics[width=0.165\linewidth,height=2.15cm]{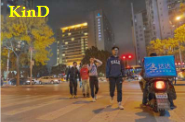} &

    \includegraphics[width=0.165\linewidth,height=2.15cm]{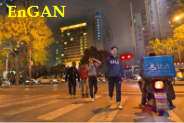} &

    \includegraphics[width=0.165\linewidth,height=2.15cm]{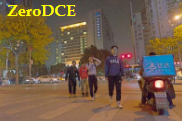} &

    \includegraphics[width=0.165\linewidth, height=2.15cm]{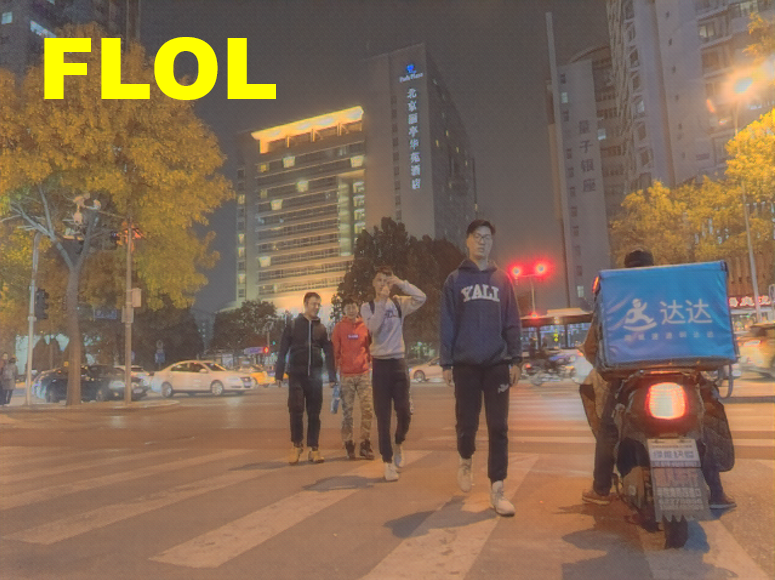} \\

    Input & RUAS~\cite{ruas} & KinD~\cite{kind} & EnGAN~\cite{enlightengan} & ZDCE~\cite{zero_reference} & \bf{FLOL} \\
    
    \end{tabular}
    \caption{Qualitative results on \textbf{DarkFace}~\cite{darkface} dataset.}
    \label{fig:darkface}
\end{figure*}

\begin{table}[h!]
\begin{center}
\begin{adjustbox}{width=1\columnwidth,center}
\begin{tabular}{c | c c | c c | c c}
\Xhline{1.5pt}
\multirow{2}{*}{Methods} & \multicolumn{2}{c|}{Complexity} & \multicolumn{2}{c|}{LOLv2-Real} & \multicolumn{2}{c}{LOLv2-Syn} \\

\cline{2-7} & \multicolumn{1}{c}{FLOPs(G)$\downarrow$} & \multicolumn{1}{c|}{Params(M)$\downarrow$} & \multicolumn{1}{c}{PSNR$\uparrow$} & \multicolumn{1}{c|}{SSIM$\uparrow$} &  \multicolumn{1}{c}{PSNR$\uparrow$} & \multicolumn{1}{c}{SSIM$\uparrow$}  \\

\Xhline{1.5pt}
SID~\cite{sid} &13.73 &7.76 &13.24  &0.442  &15.04 &0.610 \\ DeepUPE~\cite{deep_upe}  & 21.10    &1.02   &13.27 & 0.452   &15.08 &0.623\\

RF~\cite{rf} &46.23 &21.54  &14.05  &0.458   &15.97 &0.632 \\ DeepLPF~\cite{deep_lpf}  &5.86 &1.77  &14.10 &0.480 &16.02 &0.587\\

IPT~\cite{ipt}    &6887 &115.31       & 19.80        & 0.813        &18.30   &0.811 \\ 
UFormer~\cite{uformer}   &12.00 &5.29        & 18.82        & 0.771        & 19.66     &0.871\\

RetinexNet~\cite{retinex_net}  & 587.47    & 0.84   & 15.47   & 0.567  &17.13  &0.798 \\
Sparse~\cite{lol_v2}     &53.26 &2.33  &20.06  &0.816   &22.05    &0.905\\

EnGAN~\cite{enlightengan}  &61.01  &114.35    &18.23  &0.617   &16.57   &0.734 \\
RUAS~\cite{ruas}     &\textcolor{red}{0.83} &\textcolor{red}{0.003}  &18.37  &0.723   &16.55    &0.652\\

3DLUT~\cite{3dlut}     &0.075 &0.59  &17.59  &0.721   &18.04    &0.800\\

FIDE~\cite{fide}     &28.51 &8.62  &16.85  &0.678   &15.20    &0.612 \\ DRBN~\cite{drbn} &48.61  &5.27    &20.29   & 0.831    & 23.22     & 0.927\\

KinD~\cite{kind} &34.99    &8.02 &14.74 &0.641  &13.29 &0.578 \\ Restormer~\cite{restormer}   &144.25 &26.13       &19.94         &0.827         &21.41      &0.830\\

FECNet~\cite{deepfourier}   &11.84 &0.15  &20.67   &0.795  &22.57  &0.893 \\

CoLIE~\cite{Chobola2024}   &8.06 &0.120  &15.79   &0.497  &18.98  &0.685 \\ 

Zero-IG~\cite{shi2024zero}   &30.19 &0.08  &18.13   &0.746  &15.77  &0.762 \\

GLARE~\cite{GLARE}   &508.42 &59.48  &22.51   &0.87  & -  &- \\

MIRNet~\cite{mirnet}   &785 &31.76  &20.02   &0.820  &21.94  &0.876\\

SNR-Net~\cite{snr_net}   &26.35   &4.01   &21.48  &0.849  &24.14 &\textcolor{blue}{0.928} \\ 

Bread~\cite{guo2023low}   &19.85   &2.02   &20.83  &0.847  &17.63 &0.919 \\ 

LLFlow~\cite{wang2022low}      &1050     &17.42    &17.43   &0.831    &-      &-\\

ESDNet~\cite{yu2022towards}      &28.82     &10.62    &20.48   &0.841    &-      &-\\

FourLLIE~\cite{wang2023fourllie}      &5.8     &0.120    &21.60   &0.847    &24.17      &0.917\\

Retinexformer~\cite{Cai_2023_ICCV}      &15.57  &1.61  &22.80      &0.840    &\textcolor{red}{25.67} &\textcolor{red}{0.930} \\

SG-LLIE~\cite{dong2025towards}      &35.80  &12.67  & 22.84      &\textcolor{blue}{0.859}    &- &- \\

UHDFour~\cite{uhdll}   &9.58  &33.64   &21.78  &\textcolor{red}{0.87} &-   &-\\

LightenDiff~\cite{Jiang_2024_ECCV}   &2257.42  &26.54   &\textcolor{blue}{22.88}  &0.855 &21.58   &0.869\\
\hline
\textbf{FLOL(Ours)}     &\textbf{2.08}   &\textbf{0.094}   &\textbf{21.75}   &\textbf{0.847} &\textcolor{blue}{24.34}      & \textbf{0.906} \\ \textbf{FLOL+(Ours)}     &\textcolor{blue}{1.78}     &\textcolor{blue}{0.06}   &\textcolor{red}{22.93}   &\textbf{0.832} &\textbf{22.11}      & \textbf{0.901}\\
\Xhline{1.5pt}
\end{tabular}
\end{adjustbox}
\end{center}
\caption{Quantitative comparisons on the \textbf{LOLv2-Real} and \textbf{LOLv2-Synthetic} datasets~\cite{lol_v2}. Our models, \textbf{FLOL} and \textbf{FLOL+}, obtain a comparable performance with the best SOTA methods, while being notably smaller (less parameters and operations) and more efficient. All values are  collected from \cite{Cai_2023_ICCV, uhdll, wang2023fourllie, yan2025hvi, dong2025towards}. FLOPs were calculated using input image of size $256\times256$.}
\label{tab:quantitative}
\end{table}

\paragraph{Implementation Details.} 
We train our method end-to-end using Adam optimizer~\cite{Kingma2014AdamAM} (with $\beta_{1}=0.9$ and $\beta_{2}=0.999$) and a learning rate $4\times10^{-4}$ that gradually decreases to $1\times10^{-6}$ using cosine annealing. We use a mini-batch size of 32, considering that we take random crops of size $256 \times 256$ from the original low/high pair of images. We perform simple augmentations that include flips and rotations. An NVIDIA 4090Ti is used to train our model in $\approx6$ hours. We train by using the LOLv2-Real~\cite{lol_v2} and LSRW~\cite{hai2023r2rnet} datasets combined, such that the model can learn from more diverse scenes, sensors and conditions. Our model is evaluated using the standard quality metrics PSNR and SSIM~\cite{ssim}. We train the model using the following loss function that combines distortion and perception terms:

\begin{equation}
    \mathcal{L} = \lVert x - \hat{x} \rVert_1 + \lVert x - x_{lol} \rVert_1 + \lambda \cdot \text{LPIPS}(x, \hat{x})
\end{equation}

The distortion term, $l_1$ loss, enforces high fidelity. Also, the intermediate loss using $\mathbf{x}_{lol}$ enforces a good intermediate result \emph{i.e.}, good illumination correction. We use LPIPS~\cite{zhang2018perceptuallpips} (pre-trained VGG-19 \cite{VGG2014}) as a perceptual loss, with a weight $\lambda=0.1$ found empirically.

\begin{table*}[]
\begin{center}
\begin{adjustbox}{width=\textwidth,center}
\begin{tabular}{c | c c c c | c c c c}
\Xhline{1.5pt}
\multicolumn{1}{c|}{\multirow{2}{*}{Spatial Resolution (px)}} & \multicolumn{4}{c|}{FLOPs (G)} & \multicolumn{4}{c}{Runtime (ms)} \\
\cline{2-9} & \multicolumn{1}{c}{Retinexformer\cite{Cai_2023_ICCV}} & \multicolumn{1}{c}{FourLLIE\cite{wang2023fourllie}} & \multicolumn{1}{c}{\bf{FLOL}} &\multicolumn{1}{c|}{\bf{FLOL+}} & \multicolumn{1}{c}{Retinexformer\cite{Cai_2023_ICCV}} &  \multicolumn{1}{c}{FourLLIE\cite{wang2023fourllie}} & \multicolumn{1}{c}{\bf{FLOL}} &\multicolumn{1}{c}{\bf{FLOL+}}\\
\Xhline{1.5pt}
640$\times$480 & 160.4 & 24.46 & \bf{9.7} & \bf{8.38} & 55.2 & 6.8 & \bf{3.9} & \bf{3.7} \\

1280$\times$720 & 481.2 & 73.38 & \bf{39.12} & \bf{25.14} & 176.8 & 20.9 & \bf{5.4} & \bf{4.9}\\

1920$\times$1080 & OOM & 165.08 & \bf{65.52} & \bf{56.56} & 409.6 & 64.9 & \bf{12.4} & \bf{9.9}\\

2560$\times$1440 & OOM & 293.48 & \bf{116.46} & \bf{100.54} & 751.3 & 123.9 & \bf{24.2} & \bf{20.8}\\
\Xhline{1.5pt}
\end{tabular}
\end{adjustbox}
\end{center}
\caption{\textbf{Efficiency Study}. We report the FLOPs and runtime of our solutions and other methods at distinct sizes. Runtimes are calculated using a NVIDIA GeForce RTX 4090. Note that Retinexformer~\cite{Cai_2023_ICCV} fails (OOM, Out-Of-Memory) with high resolution images.}
\label{tab:runtime}
\end{table*}

\paragraph{Quantitative Results.} We present our model performance on several datasets such as UHD-LL, LSRW and MIT-5k, in Tab.~\ref{tab:LSRWUHDLLMIT5K}. In addition, we show our method performance compared with a wide range of SOTA models in Tab. \ref{tab:quantitative}. We include an ablation variant of FLOL (our reference model) named \textbf{FLOL+} and we compare its performance in Figs. \ref{fig:teaser}, \ref{fig:ballgraphic} and Tab. \ref{tab:quantitative} -- the difference between FLOL+ and FLOL is explained in more detail in the ablation study in Sec.~\ref{subsec:efficiency}. Also in Tab. \ref{tab:unpaired}, we calculate some perceptual metrics of image quality assessment (IQA) and compare them with other SOTA methods on the unpaired datasets mentioned above. Our solution accomplishes good results if we compare it with other procedures such as SNR-Net~\cite{snr_net} or MIRNet~\cite{mirnet}. In particular, our model demands $37\times$ and almost $300\times$ less parameters, respectively, to reach similar or better performance. 
Compared to the most recent method, Retinexformer~\cite{Cai_2023_ICCV}, our model has a comparable performance with \textbf{$10\times$ fewer parameters}, and \textbf{$7\times$ fewer FLOPs}. Then, we can consider RUAS~\cite{ruas} which employs only 3,000 parameters, but our model achieves almost \textbf{4 dB} and almost \textbf{8 dB} more in LOLv2-Real and LOLv2-Synthetic respectively. 
Furthermore, our procedure outperforms FourLLIE~\cite{wang2023fourllie}, the best previous method in terms of efficiency-performance trade-off,  by \textbf{+0.6 dB} in LOLv2-Real. Note that we report quantitative results for some methods based on a previous analysis~\cite{Cai_2023_ICCV}, and we use the open-source weights and models of FourLLIE. Our FLOL method outperforms in terms of image quality and/or efficiency all the existing solutions. Moreover, our single model generalizes on LOLv2-Real~\cite{lol_v2} and LSRW~\cite{hai2023r2rnet} \emph{i.e.} we do not need a dataset or sensor-specific training.

\paragraph{Qualitative Results.} The visual comparison between our algorithm FLOL and other SOTA methods in real scenes is shown in Figs. \ref{fig:teaser}, \ref{fig:LSRW} and \ref{fig:LOL}, corresponding to paired datasets. For instance, methods like KinD and RUAS introduce color degradation and overexposure in Fig.~\ref{fig:LSRW}, while MIRNet returns blurry images. Also, Bread~\cite{guo2023low} predicts bad images that suffer from color degradation as shown in Fig. \ref{fig:LOL}. In addition, we also see the remarkable performance of FLOL when compared with UHDFour~\cite{uhdll} in Fig.~\ref{fig:uhd2}. In particular, UHDFour fails when it tries to recover details like color, edges and textures. As a result, it produces blurry images in which it is impossible to distinguish small letters or items correctly. Nevertheless, FLOL retrieves those elements and achieves a good illumination level as well. These images have a resolution of $960\times540$ and FLOL can process them at $\approx4$ ms.
Our method produces high-quality results with good color correction and details, without adding noise or artifacts to the image, and effectively increases the ``light" and brightness of the image, improving its visibility as a result. To prove the robustness of our approach, we also apply our model to random real scenes from unpaired datasets like DICM~\cite{dicm}, LIME~\cite{lime}, MEF~\cite{mef}, NPE~\cite{npe} and VV~\cite{vv} (please see Fig. \ref{fig:unpaired}). Also we test our model on DarkFace~\cite{darkface} dataset in Fig.~\ref{fig:darkface}. Our model is robust and can generalize to these unseen and challenging datasets. We can observe a significant improvement after applying our method.

\begin{figure*}[]
    \setlength{\tabcolsep}{0.3pt} 
    \begin{tabular}{c c c c c}
    \raggedleft

    \includegraphics[width=0.2\linewidth]{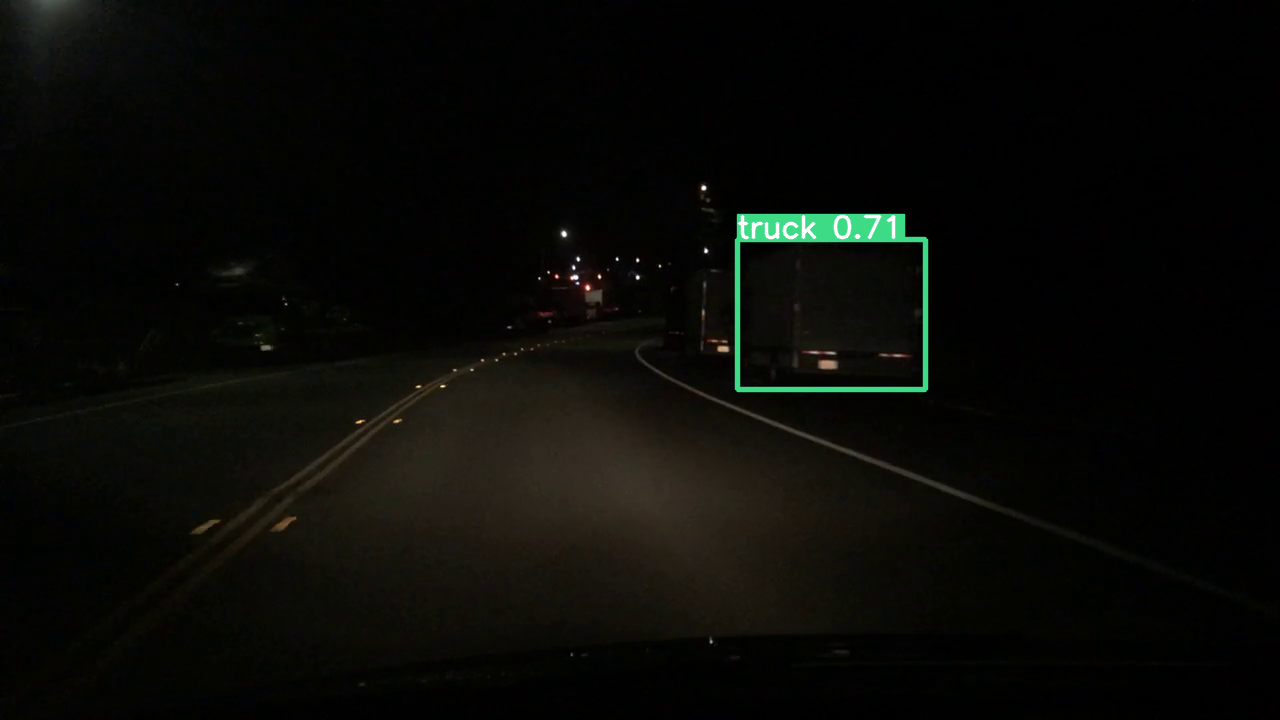} &

    \includegraphics[width=0.2\linewidth]{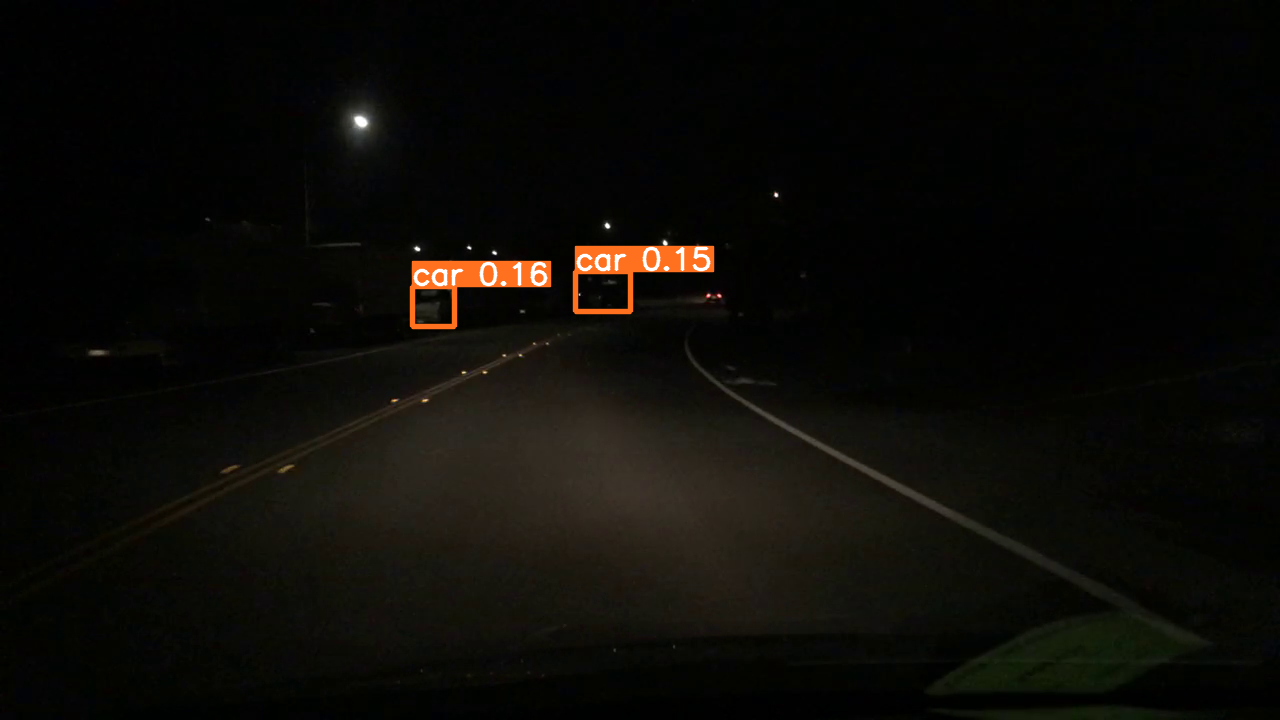} &

    \includegraphics[width=0.2\linewidth]{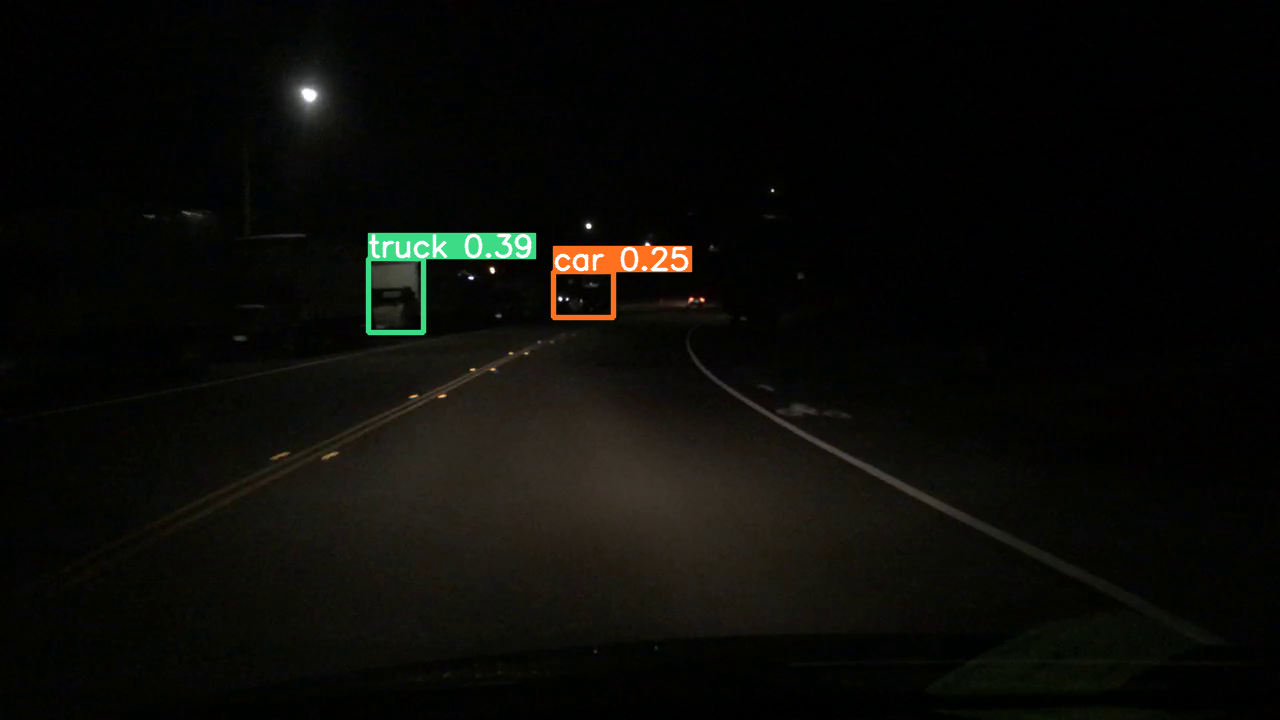} &

    \includegraphics[width=0.2\linewidth]{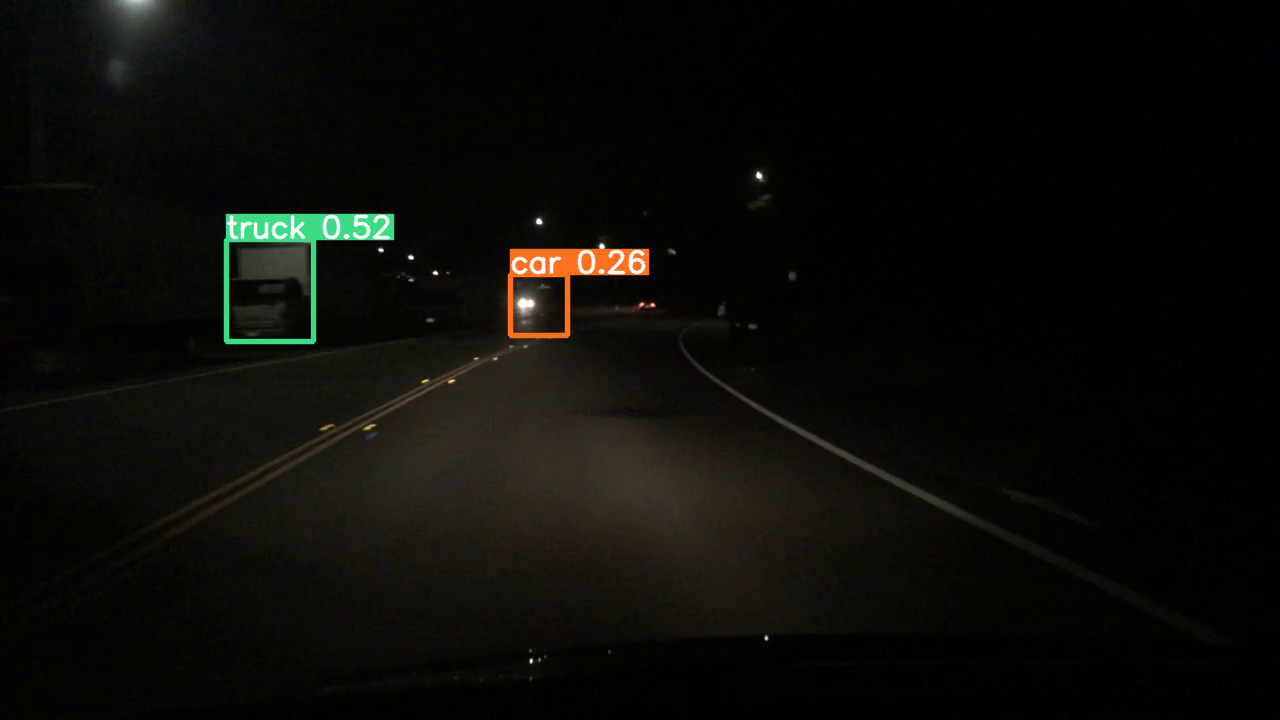} &
    
    \includegraphics[width=0.2\linewidth]{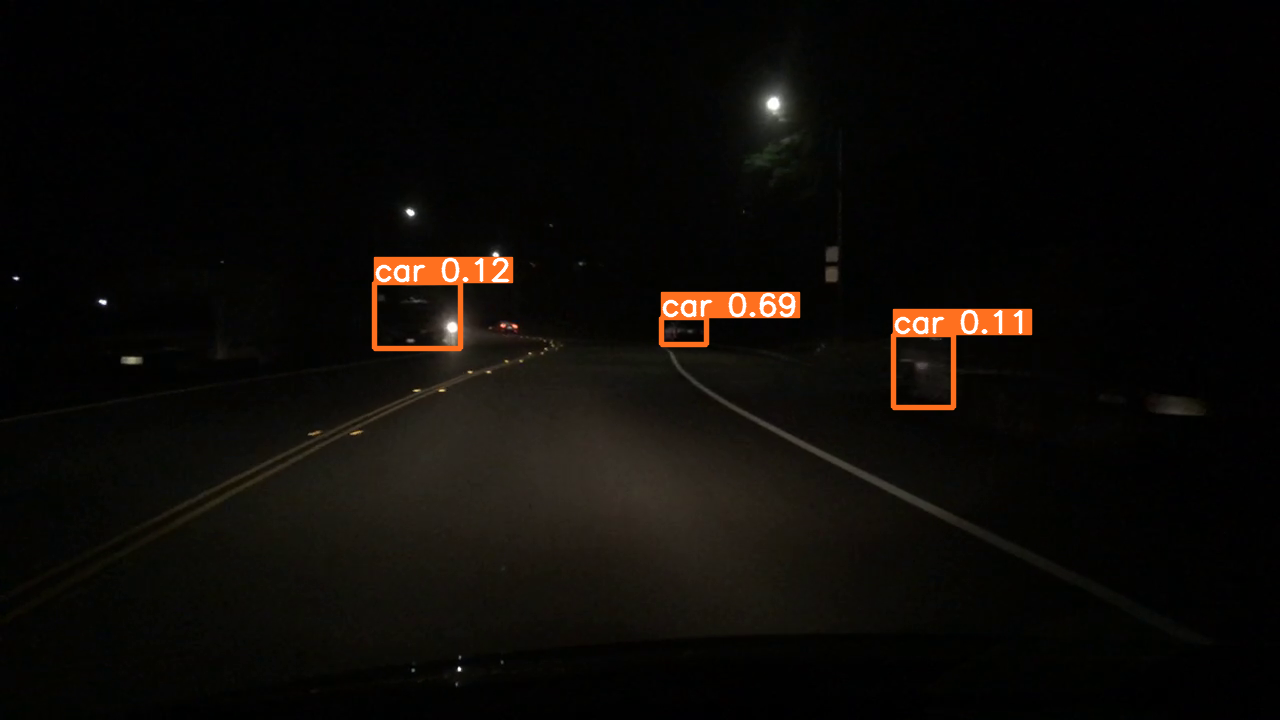} \\

    \multicolumn{5}{c}{Real low-light images from the driving scenario} \\
    \includegraphics[width=0.2\linewidth]{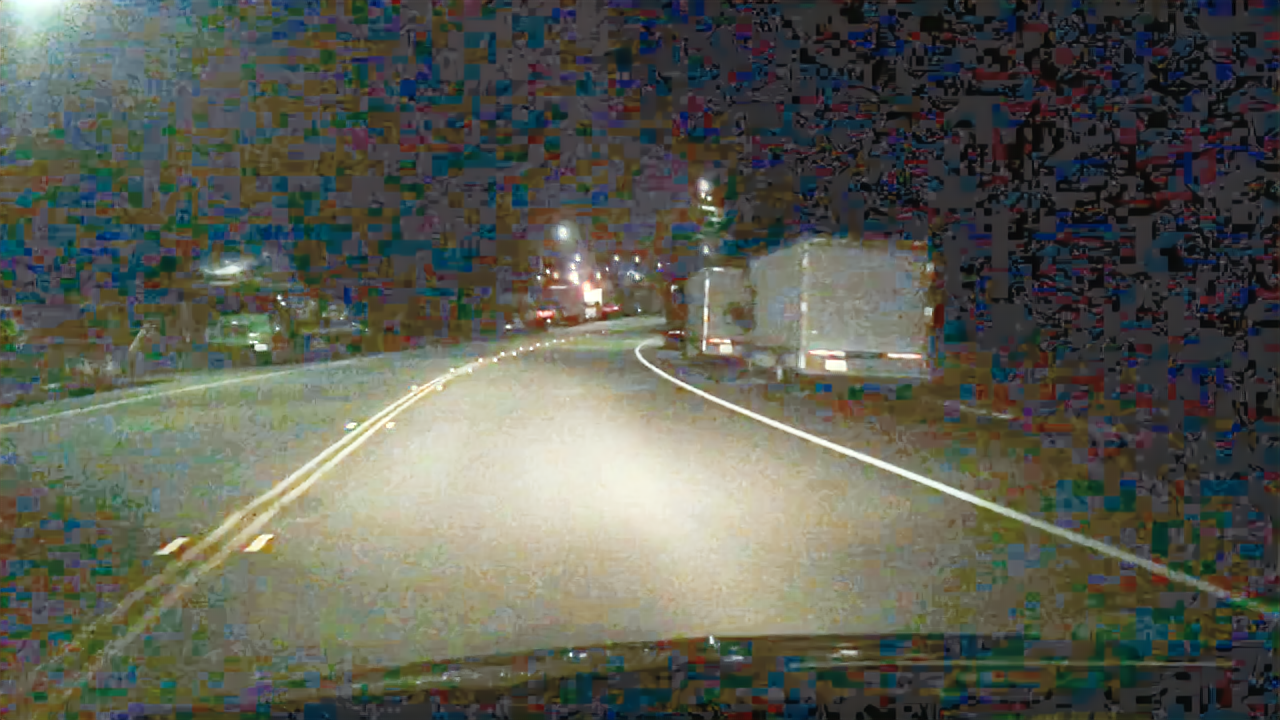} &

    \includegraphics[width=0.2\linewidth]{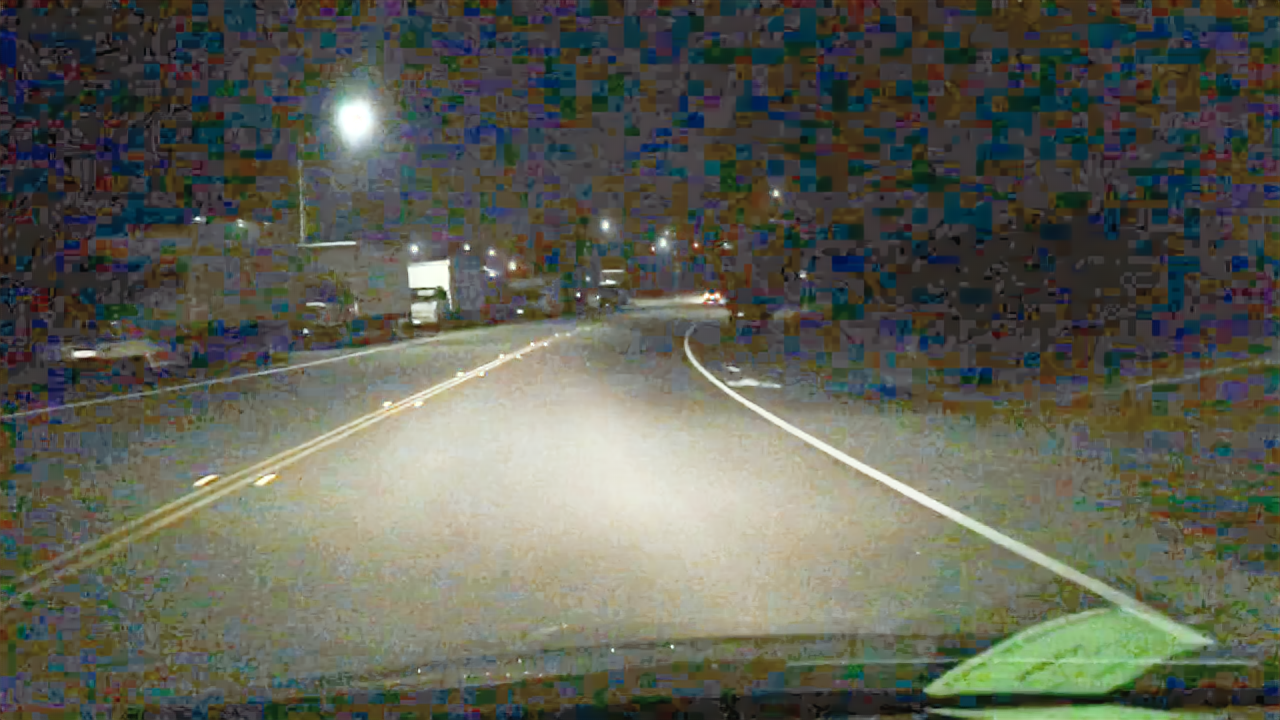} &

    \includegraphics[width=0.2\linewidth]{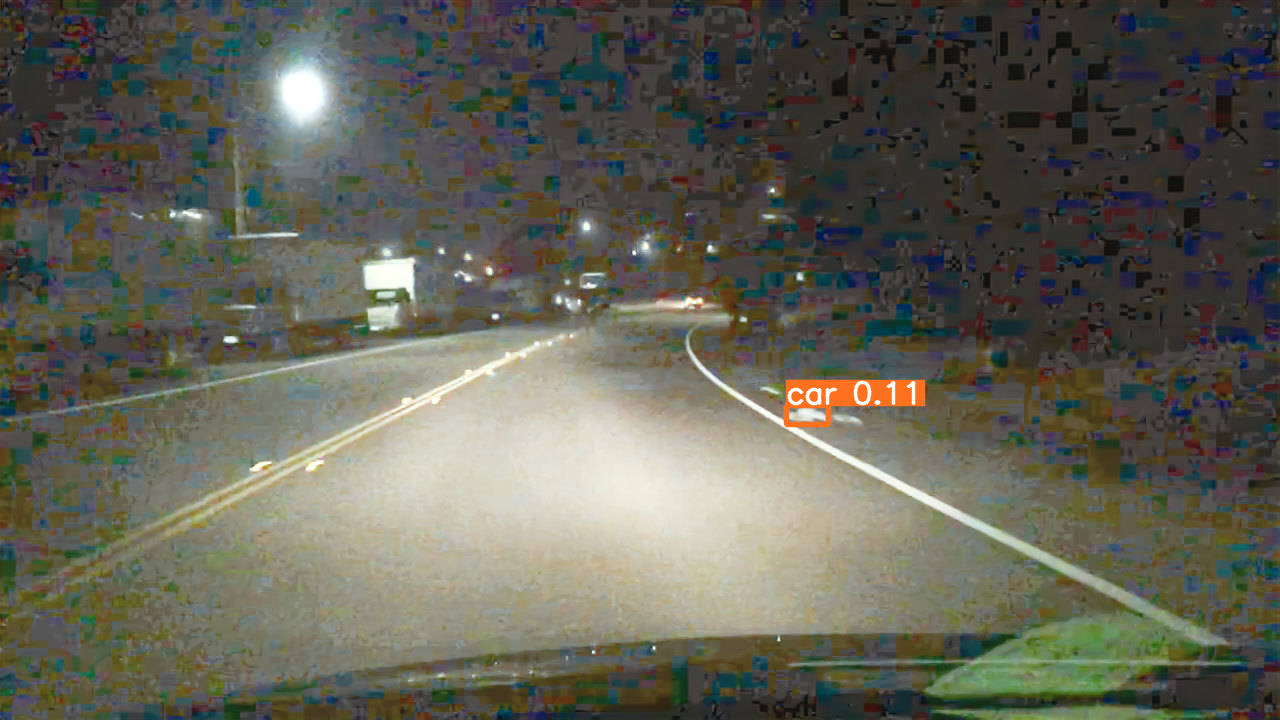} &

    \includegraphics[width=0.2\linewidth]{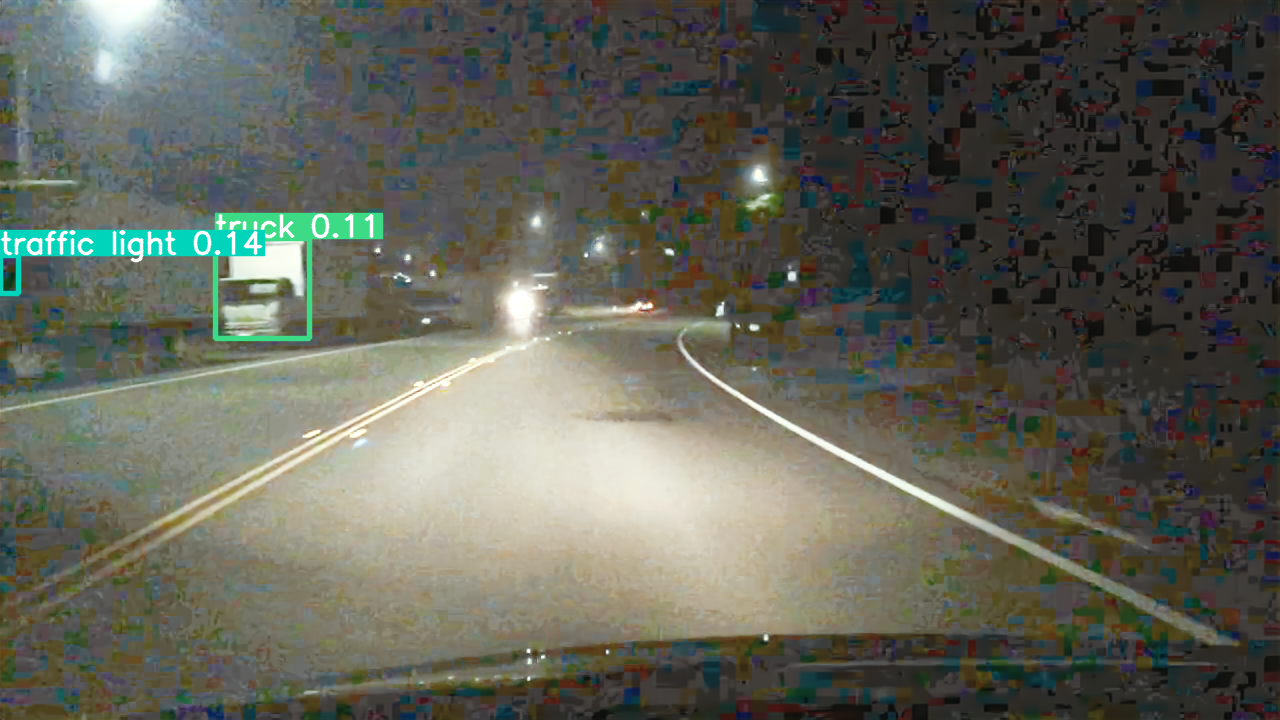} &
    
    \includegraphics[width=0.2\linewidth]{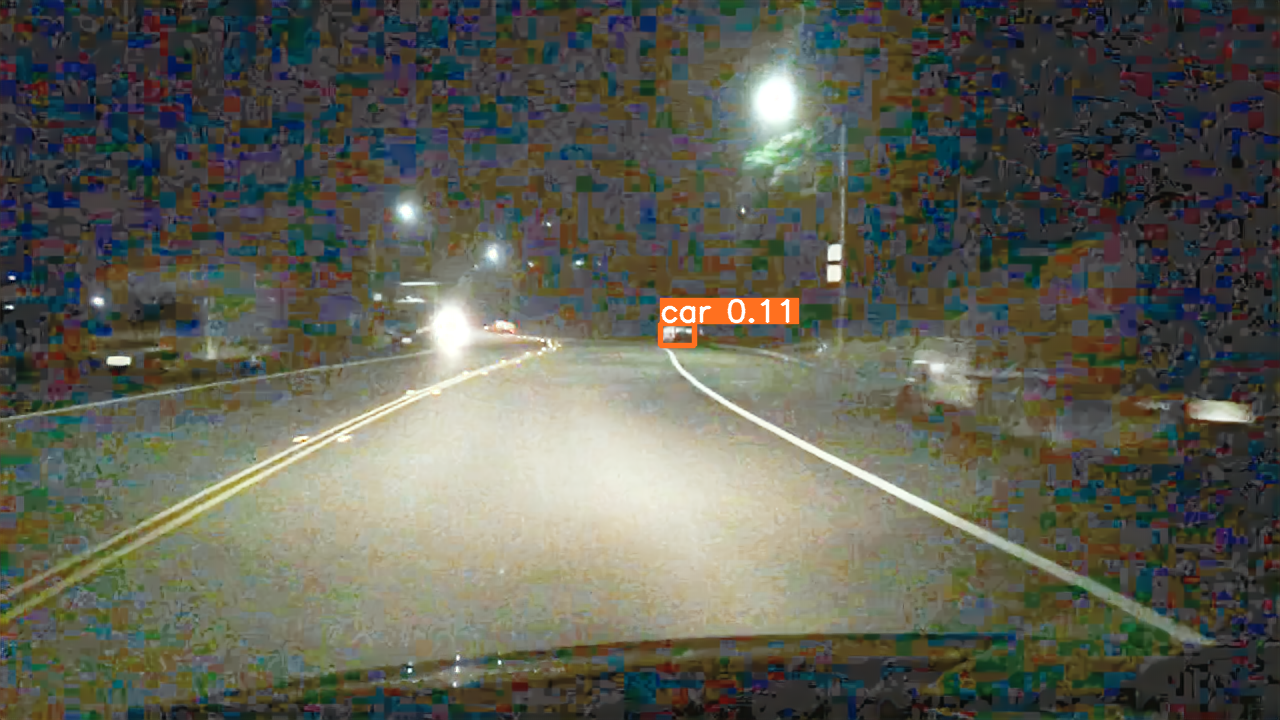} \\

    \multicolumn{5}{c}{UHDFour~\cite{uhdll}} \\
    \includegraphics[width=0.2\linewidth]{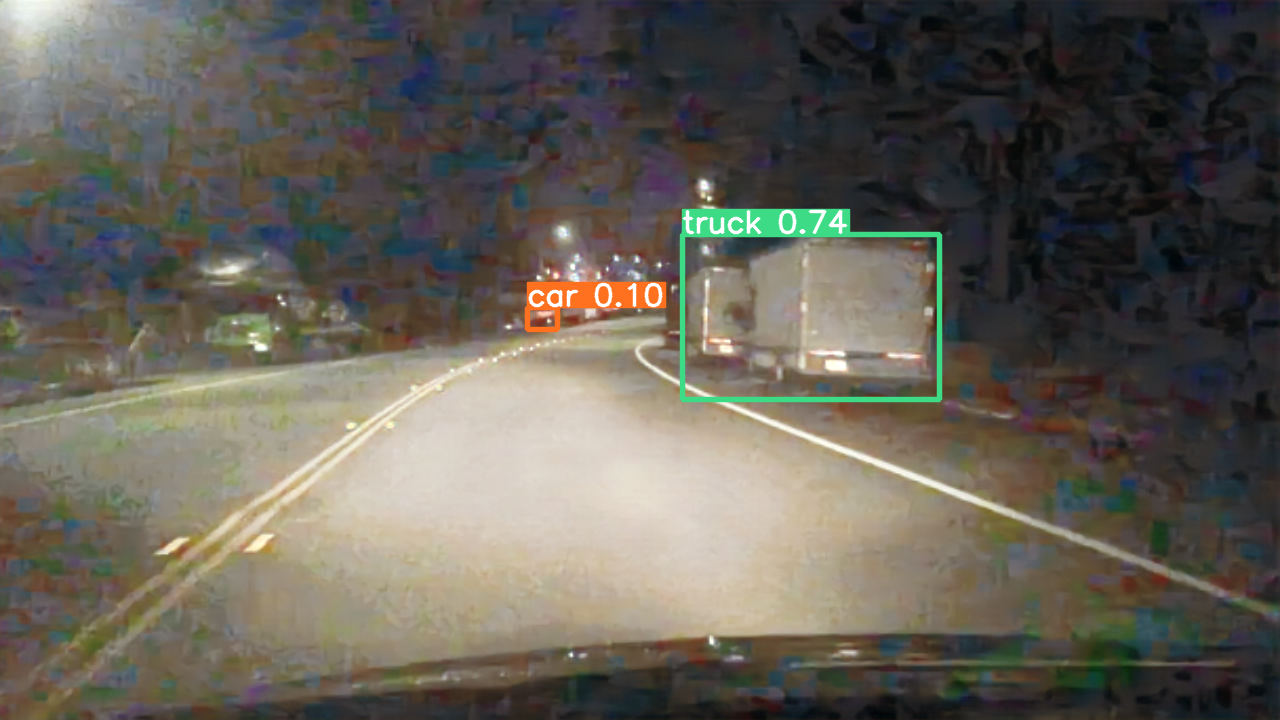} &

    \includegraphics[width=0.2\linewidth]{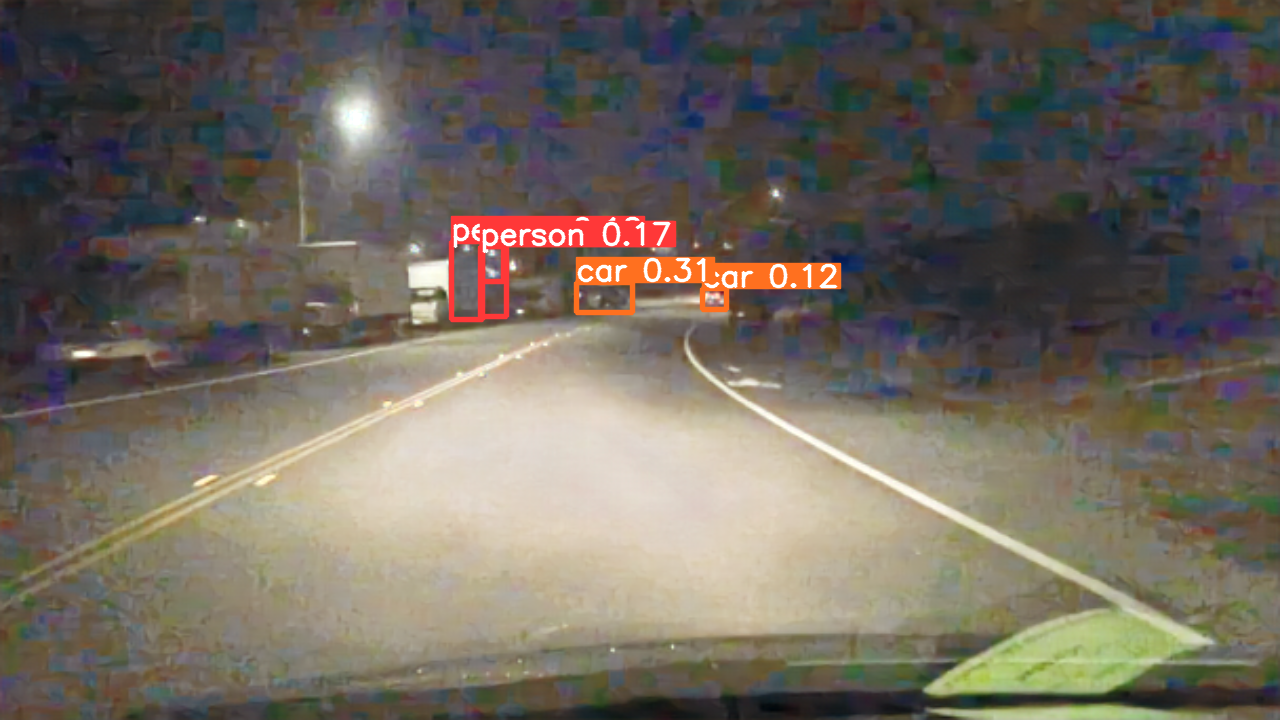} &

    \includegraphics[width=0.2\linewidth]{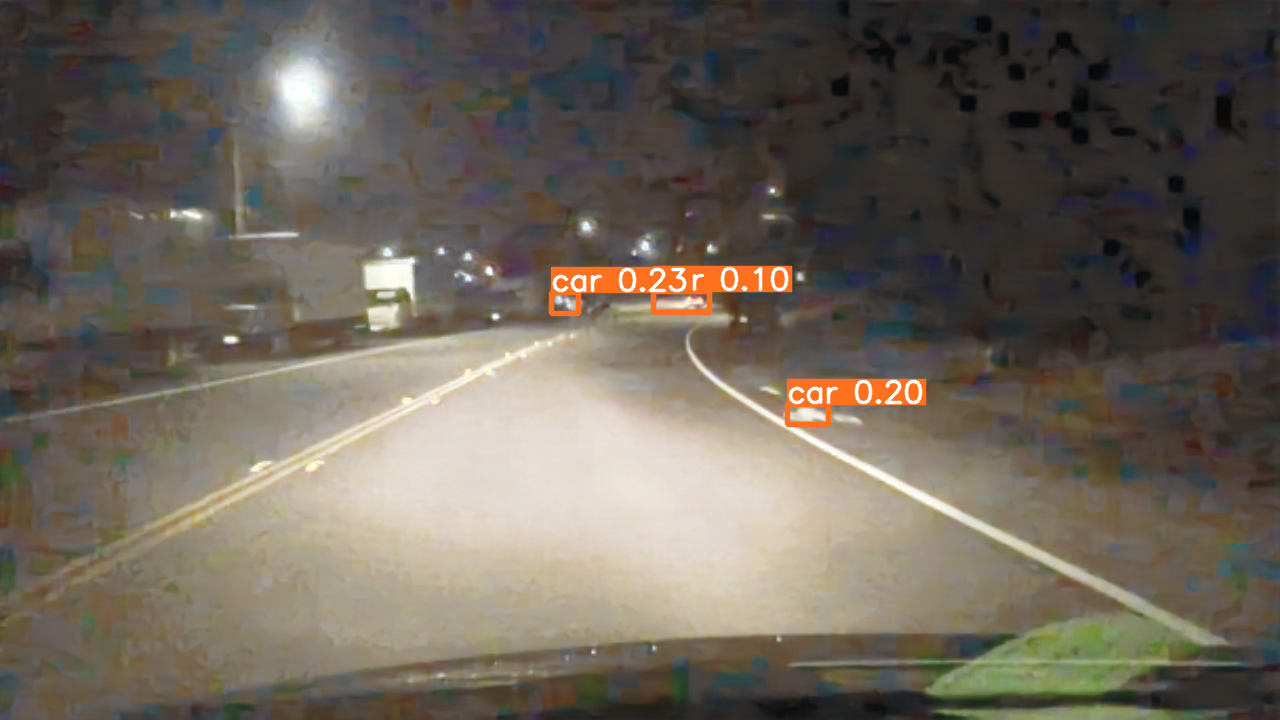} &

    \includegraphics[width=0.2\linewidth]{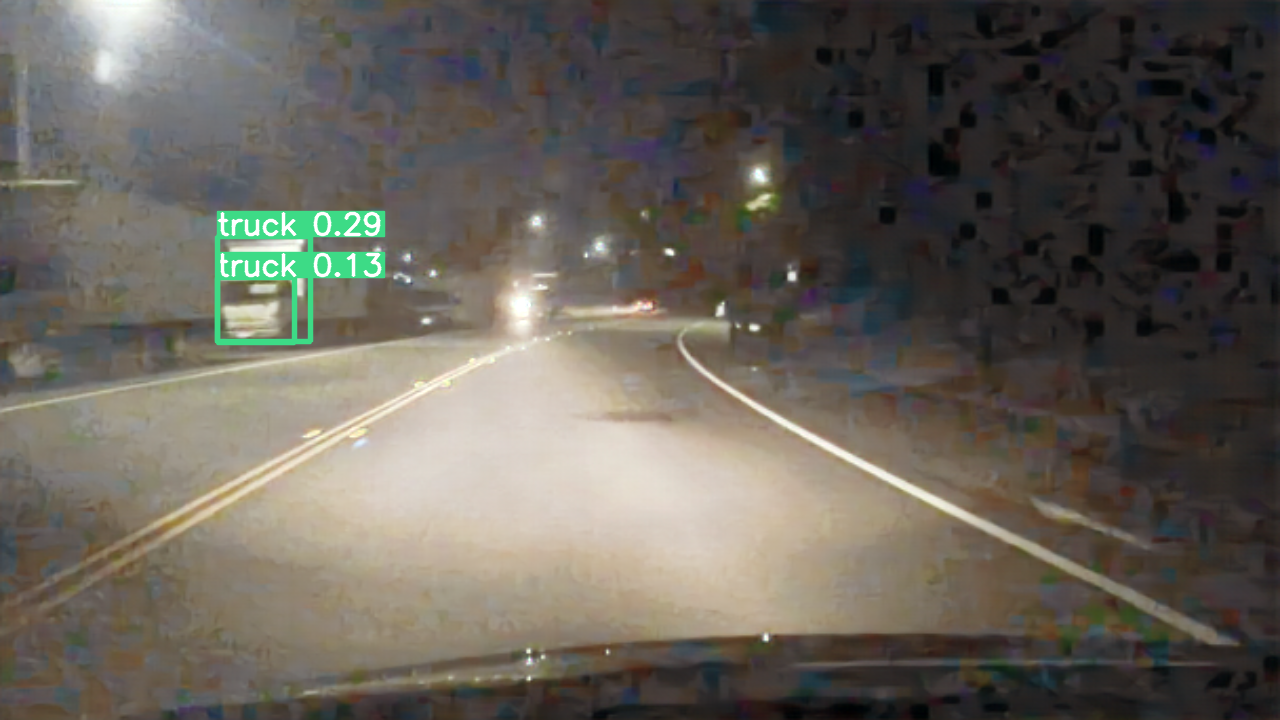} &
    
    \includegraphics[width=0.2\linewidth]{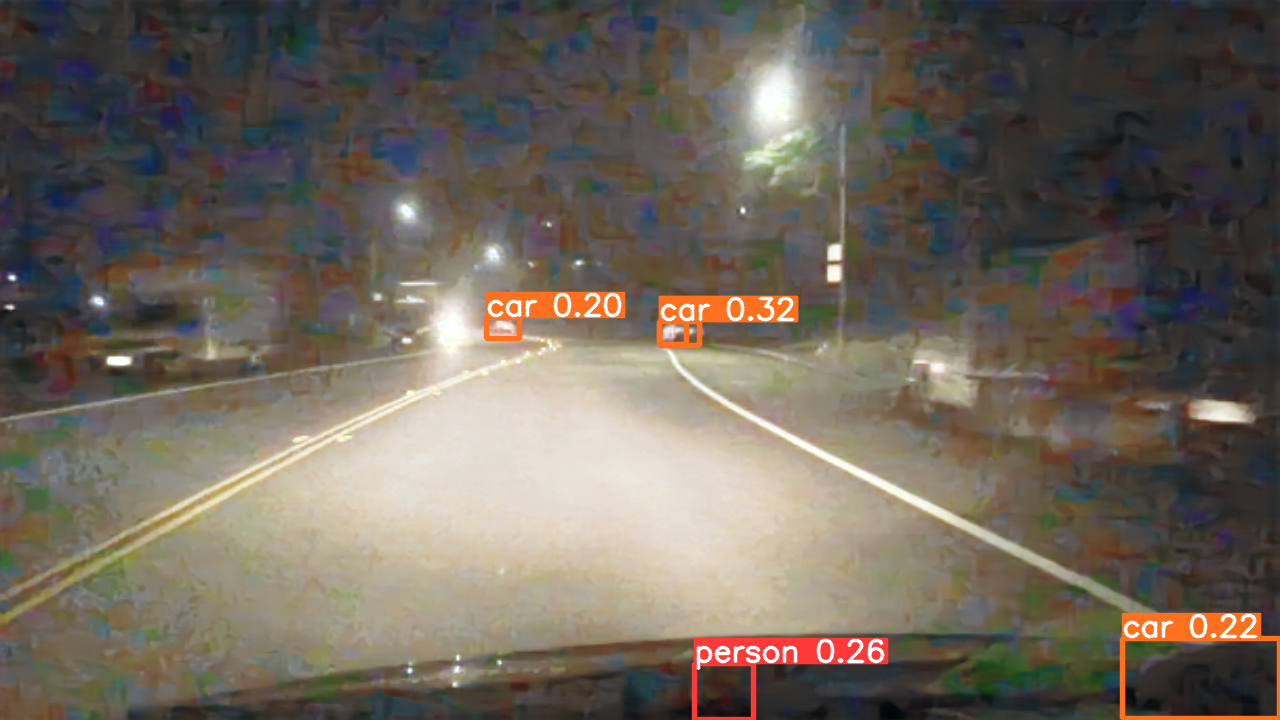} \\

    \multicolumn{5}{c}{SNR-Net~\cite{snr_net}} \\
    \includegraphics[width=0.2\linewidth]{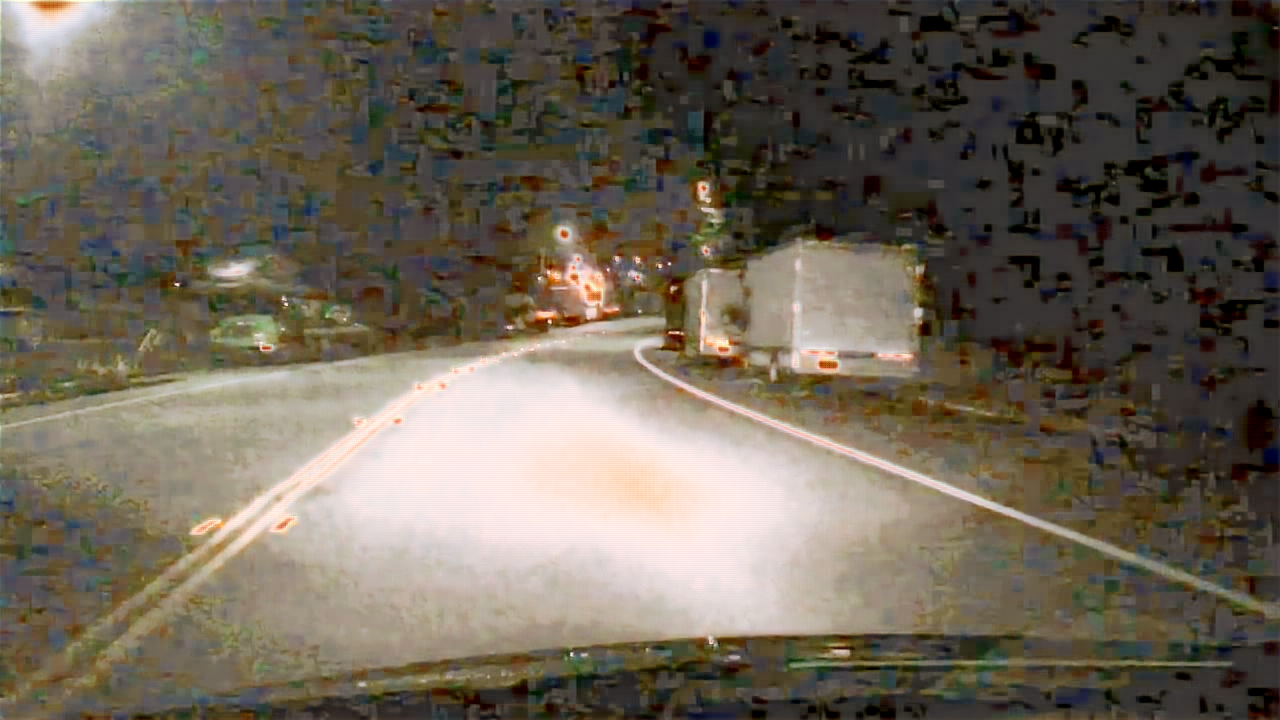} &

    \includegraphics[width=0.2\linewidth]{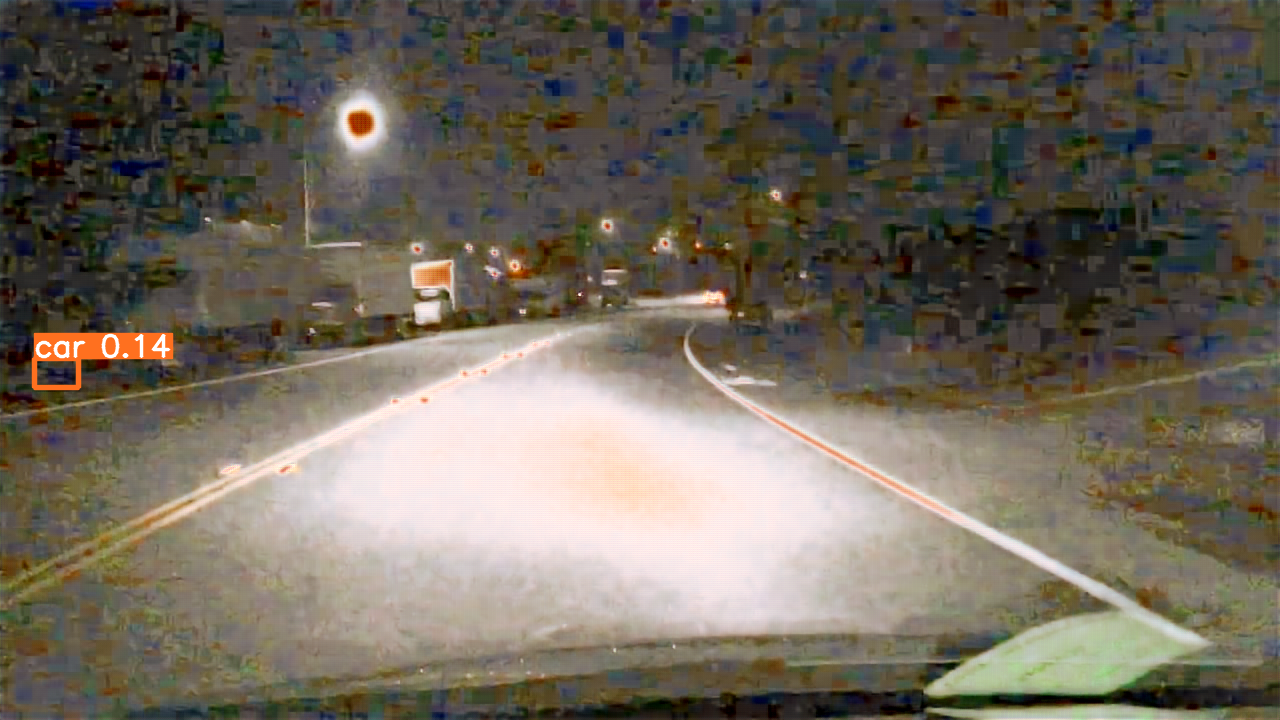} &

    \includegraphics[width=0.2\linewidth]{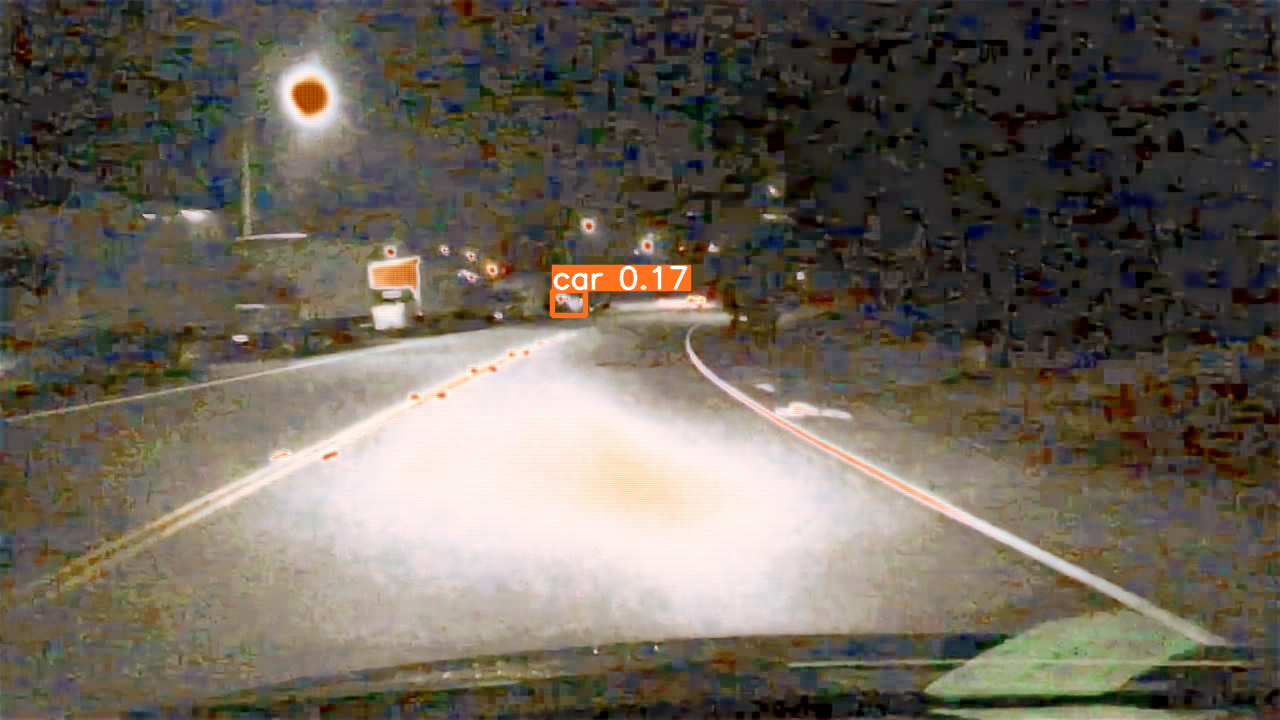} &

    \includegraphics[width=0.2\linewidth]{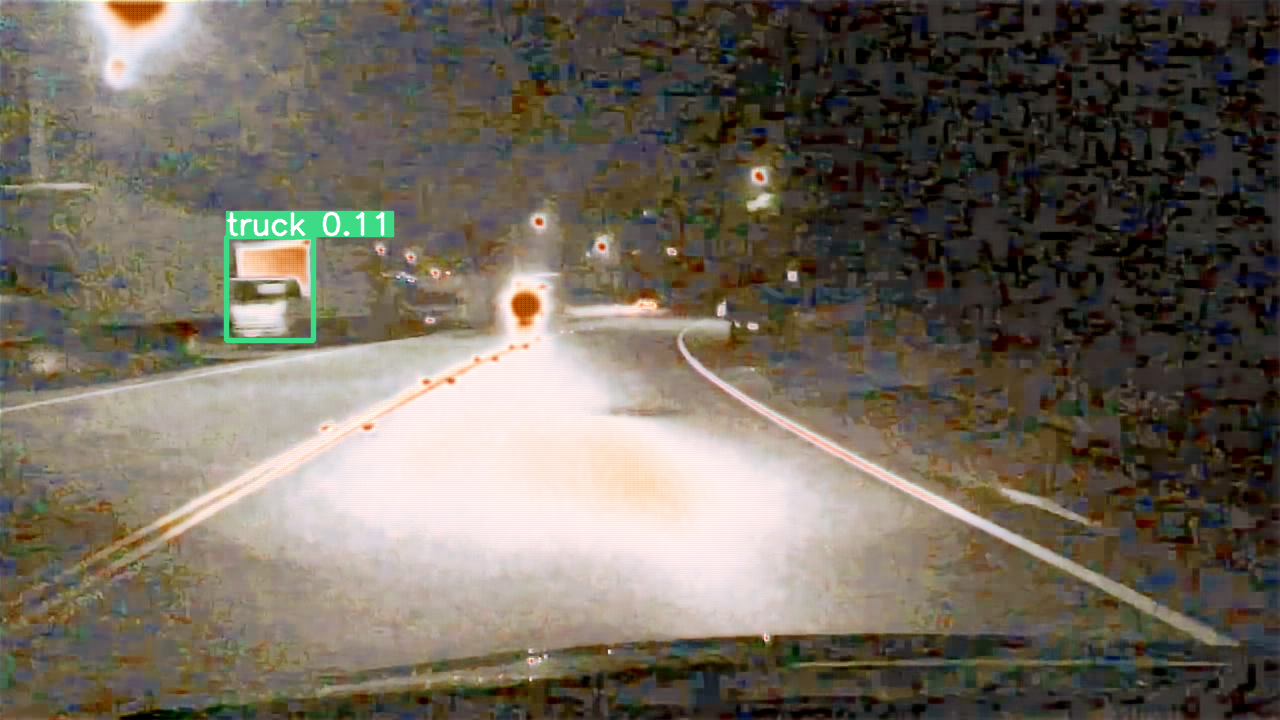} &
    
    \includegraphics[width=0.2\linewidth]{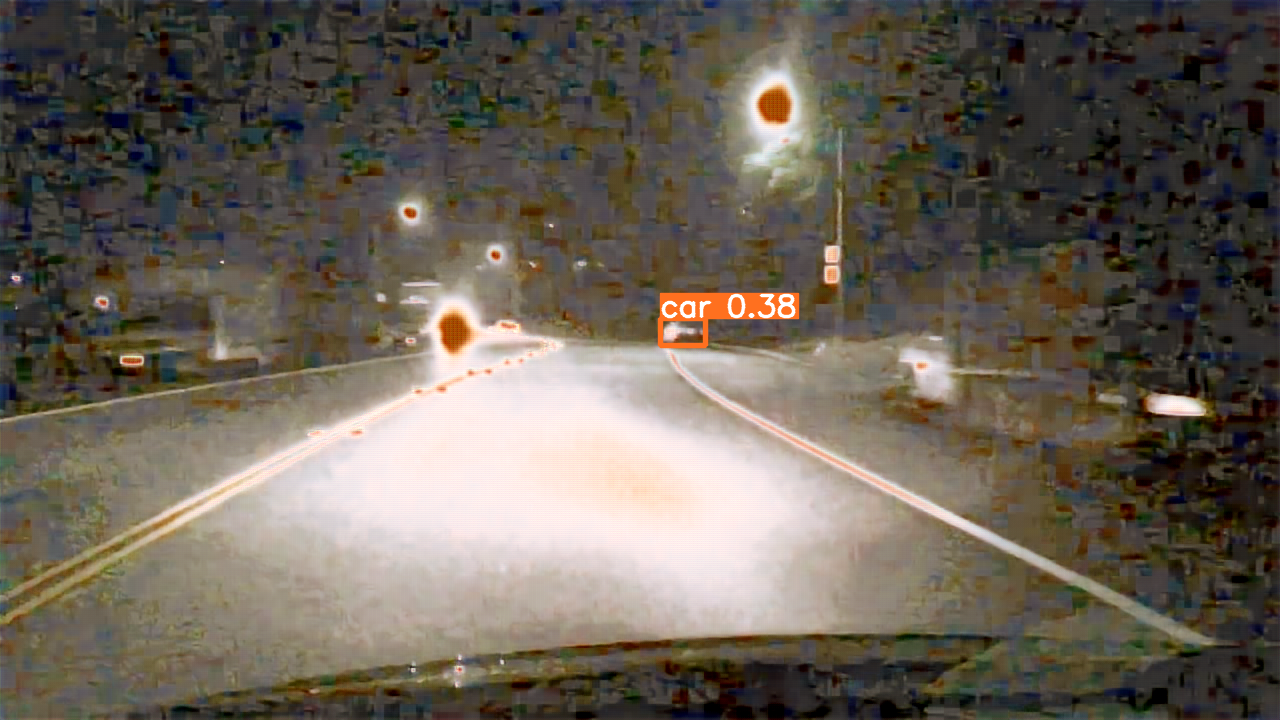} \\
    
    \multicolumn{5}{c}{Retinexformer~\cite{Cai_2023_ICCV}} \\

    \includegraphics[width=0.2\linewidth]{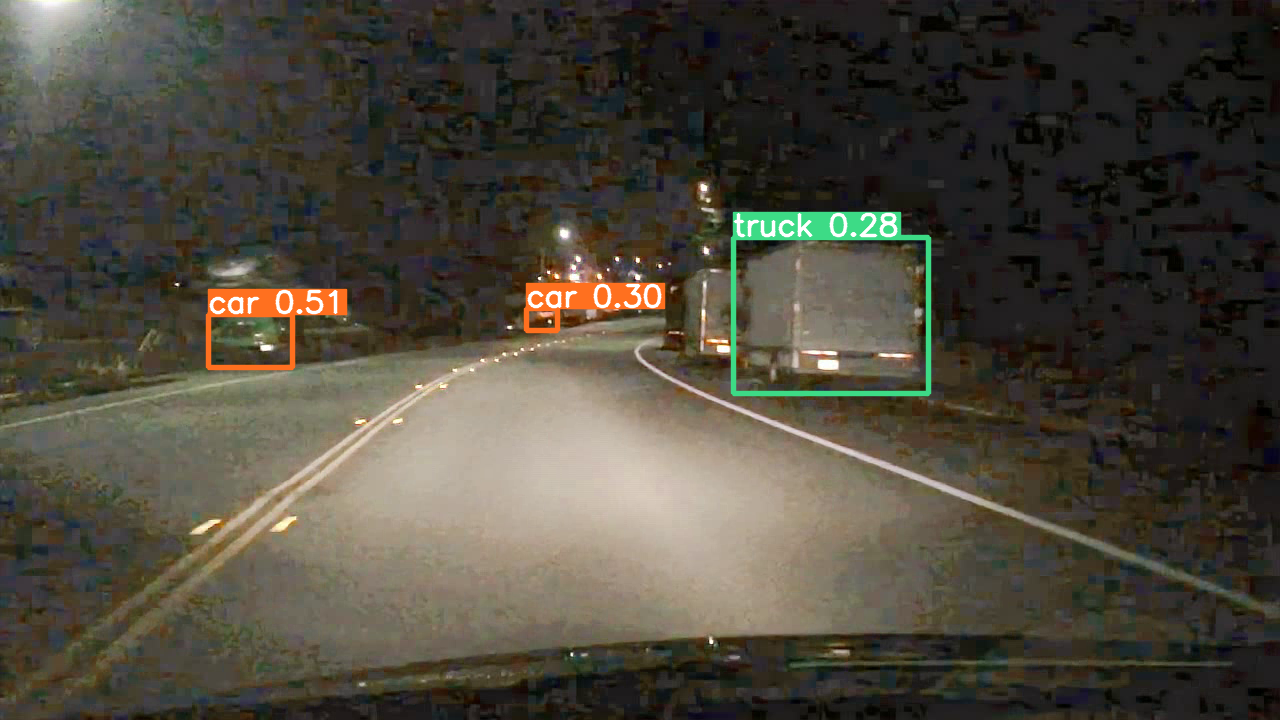} &

    \includegraphics[width=0.2\linewidth]{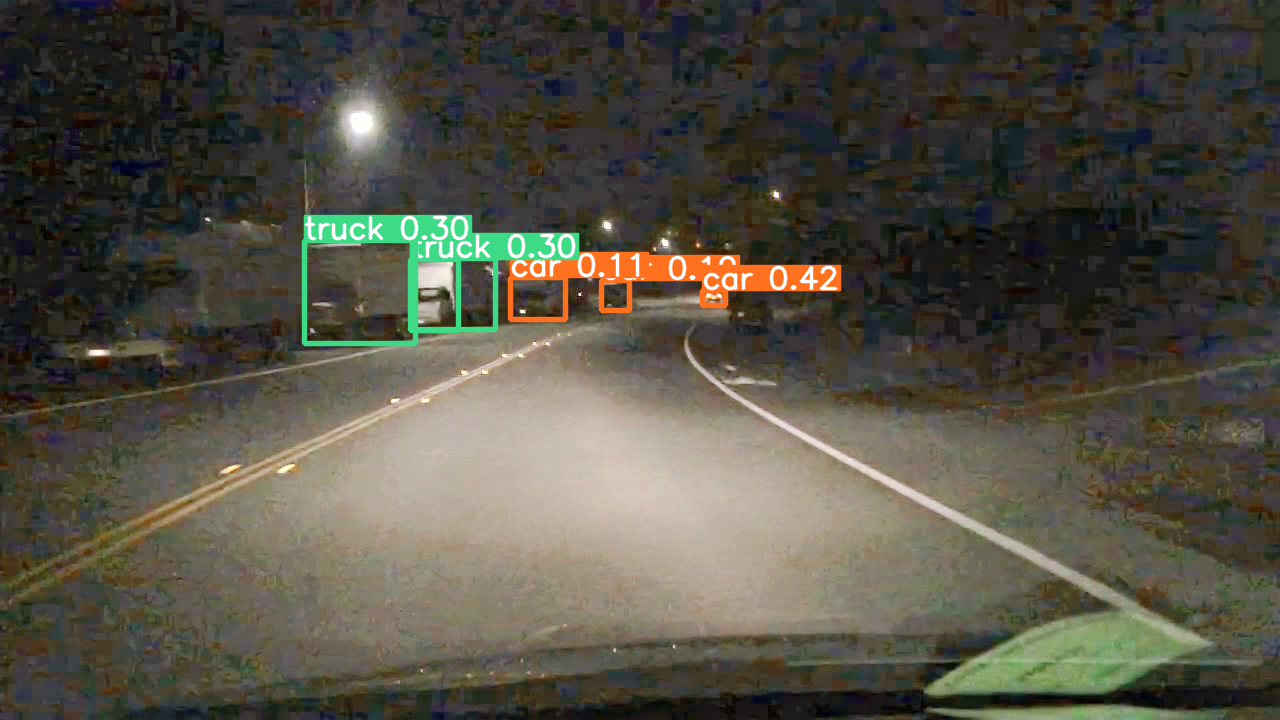} &

    \includegraphics[width=0.2\linewidth]{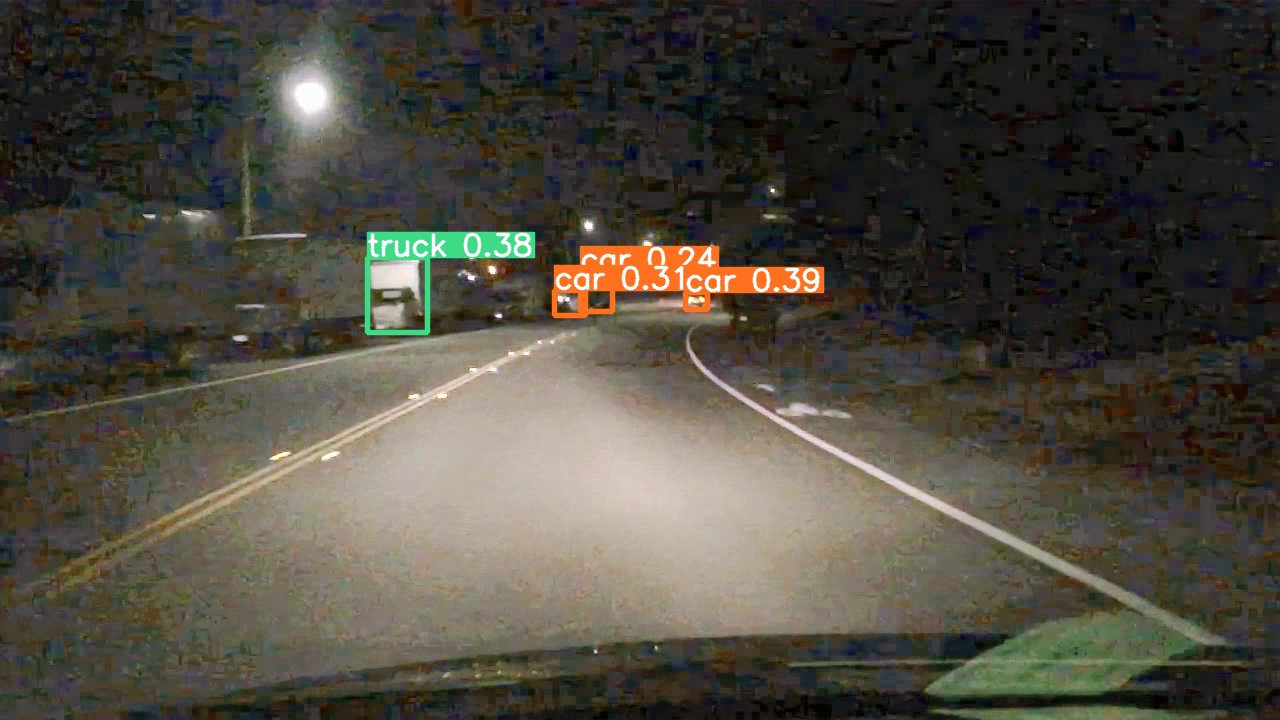} &

    \includegraphics[width=0.2\linewidth]{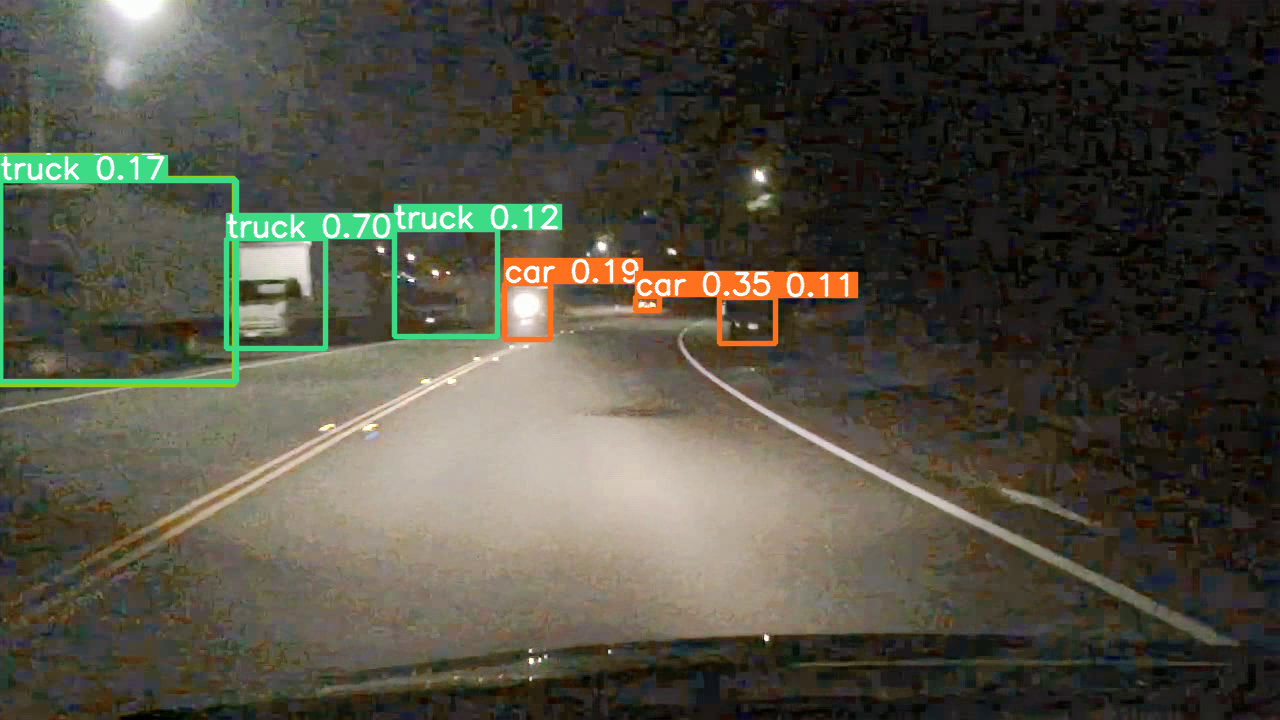} &
    
    \includegraphics[width=0.2\linewidth]{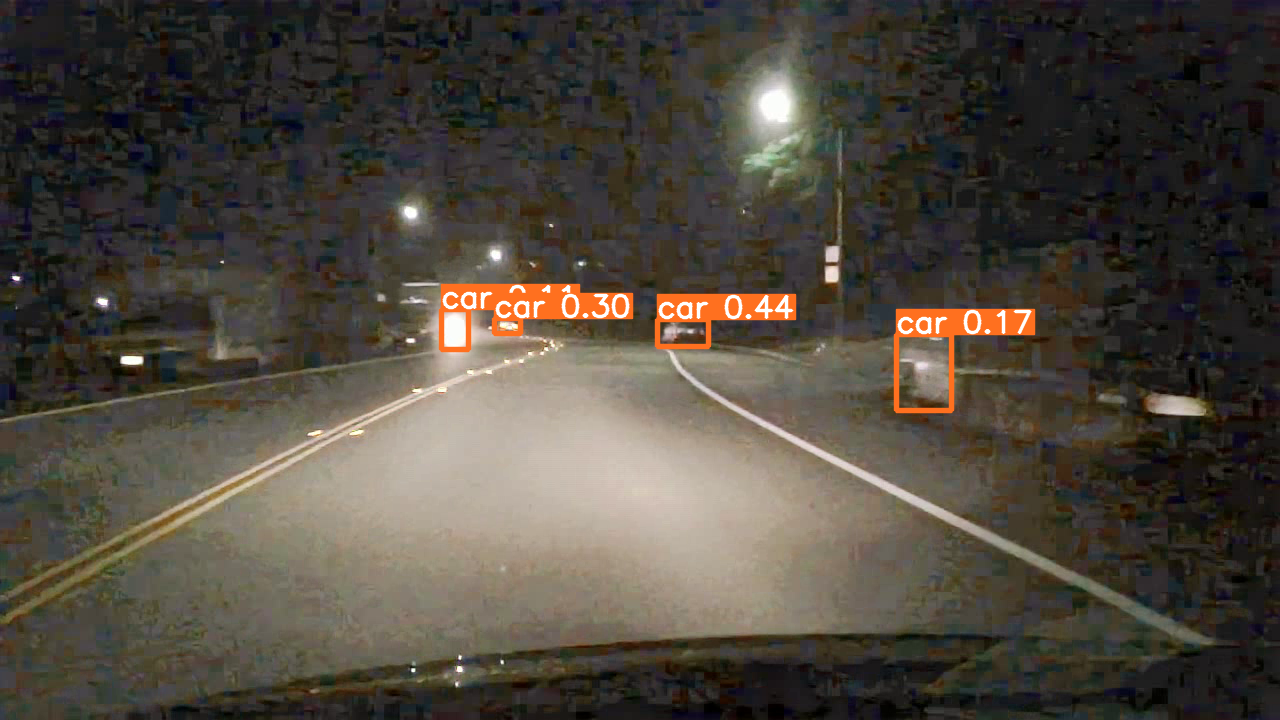} \\

    \multicolumn{5}{c}{\bf{FLOL}} \\

    \end{tabular}
    \caption{Qualitative comparison on the Autonomous Driving Dataset \textbf{BDD100k}~\cite{bdd100k} with HD videos ($1280\times720$). We selected three methods of distinct sizes (see Fig.~\ref{fig:ballgraphic}) to compare with \textbf{FLOL}. We can observe how the detector YOLOv8~\cite{yolo} performs best when we use our method for image enhancement. Recall the bad performance of Retinexformer~\cite{Cai_2023_ICCV} which introduces odd pixels in the overexposed areas of the frames. Note that the methods have not been trained on this dataset. Zoom in for best view.}
    \label{fig:bdd100k2}
\end{figure*}

\begin{table*}[]
\begin{center}
\begin{adjustbox}{width=\textwidth,center}
\begin{tabular}{c | c | cc | cc | c | c | c}
\Xhline{1.5pt}
\multicolumn{1}{c|}{\multirow{2}{*}{Structure ablations}} & \multicolumn{1}{c|}{\multirow{2}{*}{Params(M)$\downarrow$}} & \multicolumn{2}{c|}{LOLv2-Real} & \multicolumn{2}{c|}{LSRW} &\multicolumn{1}{c|}{\multirow{2}{*}{Number of Channels}} &\multicolumn{2}{c}{Complexity}\\

\cline{3-6} \cline{8-9} & &\multicolumn{1}{c}{PSNR$\uparrow$}  & \multicolumn{1}{c|}{SSIM$\uparrow$} &  \multicolumn{1}{c}{PSNR$\uparrow$} & \multicolumn{1}{c|}{SSIM$\uparrow$} &  &\multicolumn{1}{c|}{Params(M)$\downarrow$} &\multicolumn{1}{c}{FLOPs(G)$\downarrow$}\\
\Xhline{1.5pt}
Depth-wise 3x3 convolutions (\textbf{FLOL+})  &\textcolor{red}{0.06}  &\textcolor{red}{22.93}  &\textcolor{blue}{0.832}  &\textcolor{red}{19.52} &\textbf{0.567} &\multirow{2}{*}{16}  &\multirow{2}{*}{\textcolor{red}{0.094}} &\multirow{2}{*}{\textcolor{red}{2.08}} \\

Attention  &0.094  &20.09  &0.815 &19.08 &0.560  &\multirow{2.3}{*}{32} &\multirow{2.3}{*}{\textcolor{blue}{0.364}} &\multirow{2.3}{*}{\textcolor{blue}{7.62}} \\

NAFBlock   &\textcolor{blue}{0.076}  &21.25  &0.823  &\textcolor{blue}{19.32} &\textcolor{blue}{0.568} &\multirow{2.5}{*}{64} &\multirow{2.5}{*}{1.44} &\multirow{2.5}{*}{29.48} \\

\cline{1-6} \textbf{FLOL (reference model)} &\textbf{0.094} &\textcolor{blue}{21.75} &\textcolor{red}{0.847} &\textbf{19.10} &\textcolor{red}{0.583}  & & &\\
\Xhline{1.5pt}
\end{tabular}
\end{adjustbox}
\end{center}
\caption{\textbf{Architecture Ablation Study}. (Left) We show the results of PSNR and SSIM obtained with the different ablations in \textbf{LOLv2-Real}~\cite{lol_v2} and \textbf{LSRW}~\cite{hai2023r2rnet} datasets. (Right) Also, we present complexity values of our reference model FLOL with different number of channels of latent features in LOLv2-Real.}
\label{tab:ablation}
\end{table*}

\begin{table*}[h!]
\begin{adjustbox}{width=1\textwidth,center}
\centering
\begin{tabular}{c | c | c | c | c | c | c | c | c}
\Xhline{1.5pt}
\multicolumn{1}{c|}{\multirow{2}{*}{Number of Channels}} & \multicolumn{4}{c|}{Feature Concatenation} & \multicolumn{4}{c}{Feature Addition} \\
\cline{2-9} & \multicolumn{1}{c|}{PSNR} & \multicolumn{1}{c|}{SSIM} & \multicolumn{1}{c|}{Parameters(M)} & \multicolumn{1}{c|}{FLOPs(G)} &  \multicolumn{1}{c|}{PSNR} & \multicolumn{1}{c|}{SSIM} & \multicolumn{1}{c|}{Parameters(M)} & \multicolumn{1}{c}{FLOPs(G)}\\
\Xhline{1.5pt}
16 & 21.75 & 0.849 & 0.094 & 2.08 & 21.13 & 0.813 & 0.055 & 0.736 \\

32 & 22.33 & 0.833 & 0.364 & 7.62 & 22.64 & 0.823 & 0.206 & 2.34\\

64 & 22.02 & 0.817 & 1.44 & 29.48 & 23.06 & 0.837 & 0.806 & 8.36\\
\Xhline{1.5pt}
\end{tabular}
\end{adjustbox}
\caption{\textbf{Feature Propagation Ablation Study}. We compare two strategies to propagate features from encoder to decoder. ``Number of Channels" stands for the overall number of convolutional layers of latent features. We consider both magnitudes, \textbf{performance} and \textbf{computational cost} of all variants.}
\label{tab:propagation}
\end{table*}



\section{Discussion}
\label{subsec:efficiency}

\paragraph{Efficiency and Runtime.} We conduct a runtime experiment in which we have chosen a video sample from the BDD100k \cite{bdd100k} dataset. We selected some frames by cropping the video sample. We changed the spatial resolution several times to calculate the processing runtime rate when we apply our method FLOL and FLOL+. To demonstrate the efficiency of our solutions, we measure the number of operations (FLOPs) and the runtime at those different resolutions. In Tab. \ref{tab:runtime}, we provide the study and comparison with FourLLIE~\cite{wang2023fourllie}, one of the fastest solutions, and Retinexformer~\cite{Cai_2023_ICCV}. We measure runtimes 1000 times per frame and report the average runtime. Our methods have notably fewer parameters and FLOPs than previous state-of-the-art methods, which translates into real-time performance. We can process Full-HD images at $>80$ FPS on consumer GPUs. Considering the low memory requirements, the model could also be used in mobile devices. We also apply the detection model YOLOv8~\cite{yolo} before and after processing those frames with FLOL and other solutions. The qualitative results are shown in Fig. \ref{fig:bdd100k2}, where we can see the remarkable improvement in the detection task. In conclusion, our method gets similar results, considering its features and limitations, compared to other solutions that have been used on mobile phones \cite{ipienet} with different capabilities.

\paragraph{Ablation Study.}

We perform several ablations (see Tab.~\ref{tab:ablation}) of our solution by changing the inner structure of the network. In FLOL, the spatial blocks are composed of $3\times3$ usual convolutions (please see Fig.~\ref{fig:scheme}). If we use depth-wise convolutions instead of those convolutions we get FLOL+. It is a faster version of FLOL and reaches slightly higher values in PSNR and SSIM than FLOL in some studied datasets (check Tab.~\ref{tab:quantitative} and Fig. \ref{fig:ballgraphic}). Other ablations consist of the usage of NAFNet blocks~\cite{nafnet2022} instead of spatial blocks or residual attention in the \emph{Fre-Process} blocks (pixel-wise multiplication instead of pixel-wise addition). Furthermore, we conduct a study by altering the number of channels of latent features in our base model FLOL. We provide PSNR, SSIM and complexity values (parameters and FLOPs) for our method evaluated on LOLv2-Real~\cite{lol_v2}. In addition, we are also interested in how FLOPs and number of parameters scalate with the number of channels of the inner features (please check Tab. \ref{tab:ablation}). 

A final study about feature propagation is conducted by replacing the concatenations present in FLOL by additions. we discuss the application of concatenation within the decoder structure of our reference model FLOL. We perform skip connections between the encoder and the decoder following the fundamentals of a standard U-net~\cite{unet}. Those concatenations preserve features across the whole structure, making the network aware of the degradations by considering previous information. We can minimize the number of parameters of our network even more by replacing concatenations with feature additions. Thus, the total number of channels in each convolutional layer is decreased and, as consequence, the number of parameters and FLOPs are reduced. We repeat the experiment by changing the overall input number of channels of the latent features as well (please see the Tab. \ref{tab:propagation}). We employ LOLv2-Real~\cite{lol_v2} to perform this study and then extract PSNR and SSIM values. When we fix the overall number of channels at 16, additions always minimize processing times (less FLOPs), however, the achieved performance in this case is lower as they do not consider as many traits as concatenations within the encoder-decoder structure. On the contrary, concatenations preserve information through inner layers as we said, and achieve a better performance, although they always spend more time in recovering images. When the number of channels is increased to 32 or 64, the denoiser gets features which are elaborated enough in this two cases, and only applying additions instead of concatenations yields better results.


\paragraph{Limitations.} As we show in Fig.~\ref{fig:bdd100k2}, our model sometimes struggles in low-light images from unknown sensors. However, training our model using more diverse data --captured from multiple sensors-- should lead to better generalization.

\section{Conclusion}
We focus on solving the low-light image enhancement (LLIE) problem from the efficiency and robustness point of view. We present a new method, FLOL, an end-to-end neural network with two main blocks: frequency (FIE) enhancement, and denoising. Our method achieves results comparable to state-of-the-art in multiple real-world low-light benchmarks. Moreover, it requires $10\times$ less parameters and has $7\times$ less FLOP operations than previous works such as RetinexFormer, which allows to process 1080p images in real-time. Therefore, FLOL represents a new baseline with notable robustness in real scenarios.


\section*{Acknowledgments}
The authors thank Supercomputing of Castile and Leon (SCAYLE, León, Spain) for assistance with the model training and GPU resources.

This work was supported by Spanish funds through Regional Funding Agency Institute for Business Competitiveness of Castile and Leon (MACS.2 project “Investigación en tecnologías del ámbito de la movilidad autónoma, conectada, segura y sostenible”).

\begin{figure*}[!ht]
    \centering
    \setlength{\tabcolsep}{1pt} 
    \begin{tabular}{c c c c}
    
    \includegraphics[width=0.25\linewidth]{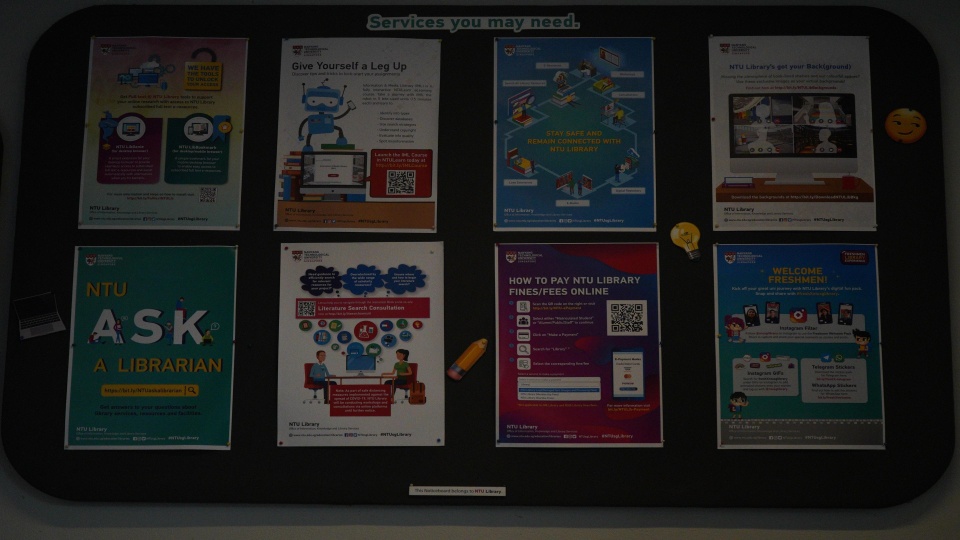} &

    \includegraphics[width=0.25\linewidth]{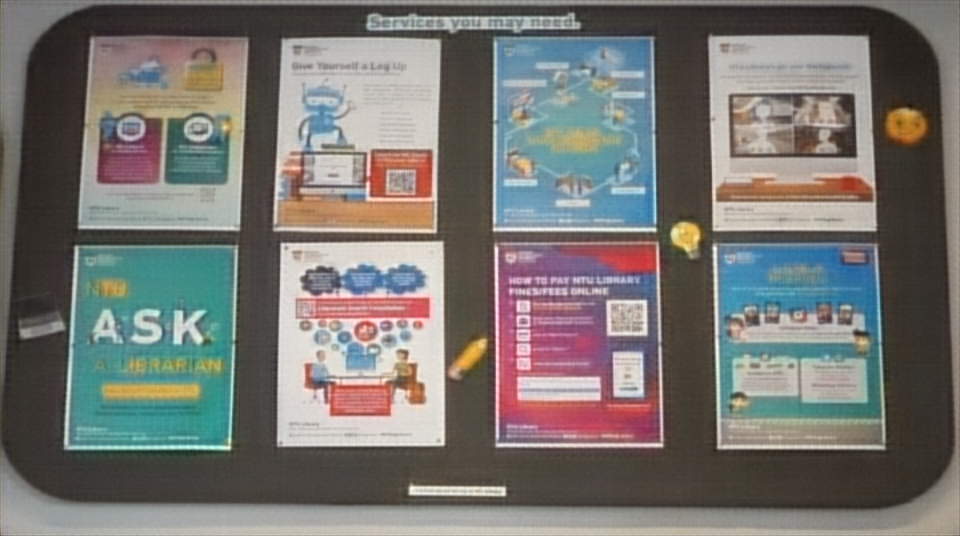} &

    \includegraphics[width=0.25\linewidth]{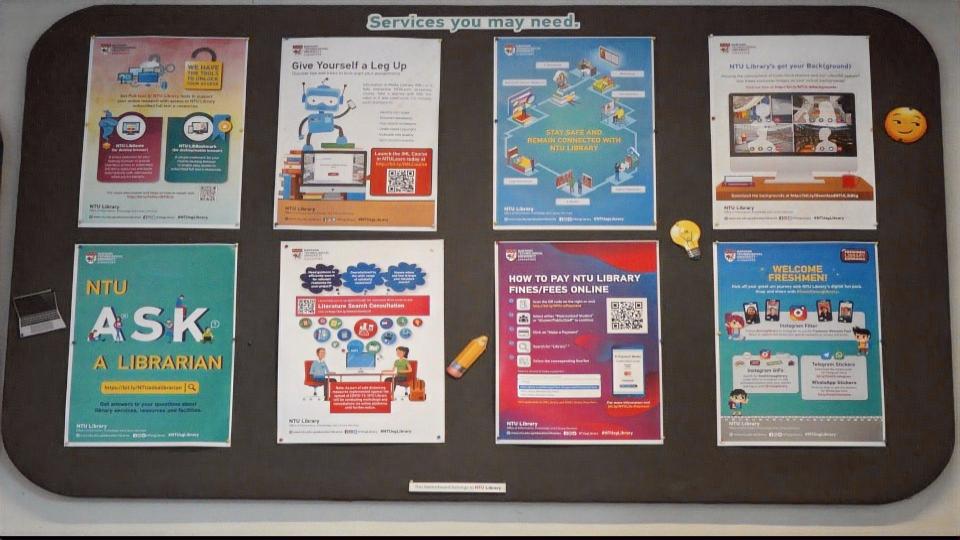} &

    \includegraphics[width=0.25\linewidth]{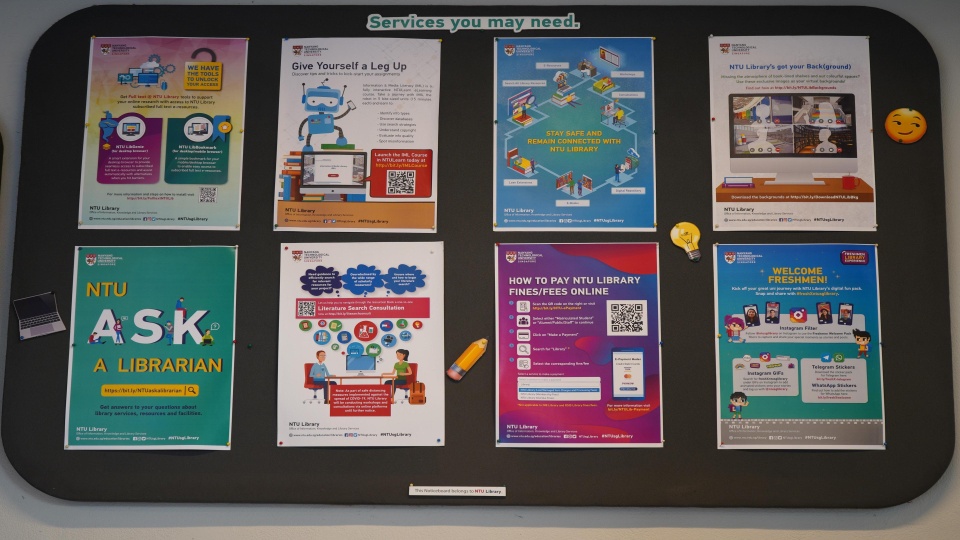} \\
    
    \includegraphics[width=0.25\linewidth]{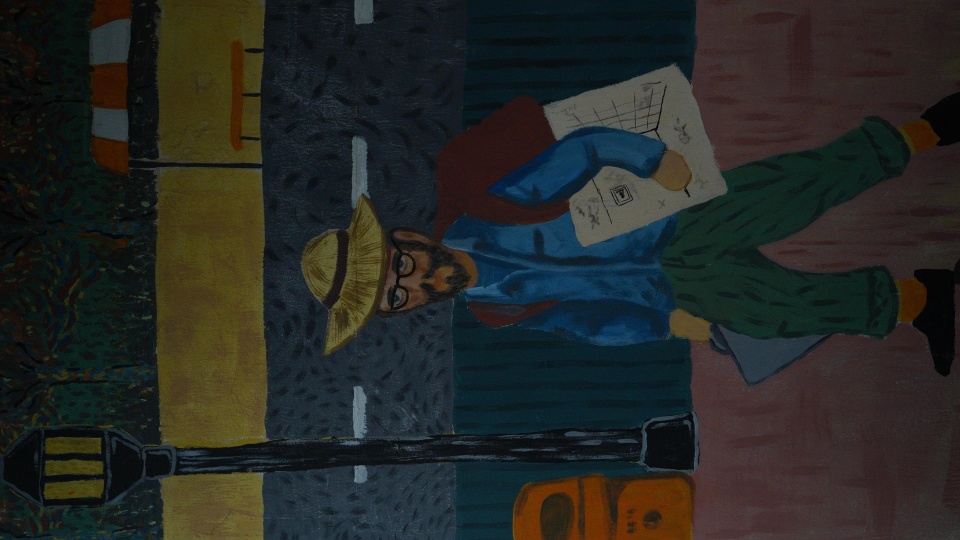} &

    \includegraphics[width=0.25\linewidth]{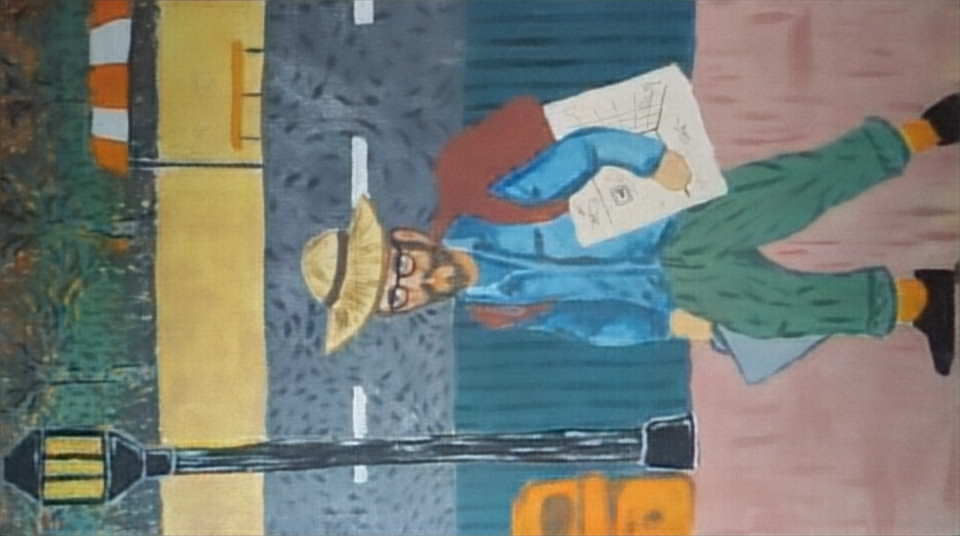} &

    \includegraphics[width=0.25\linewidth]{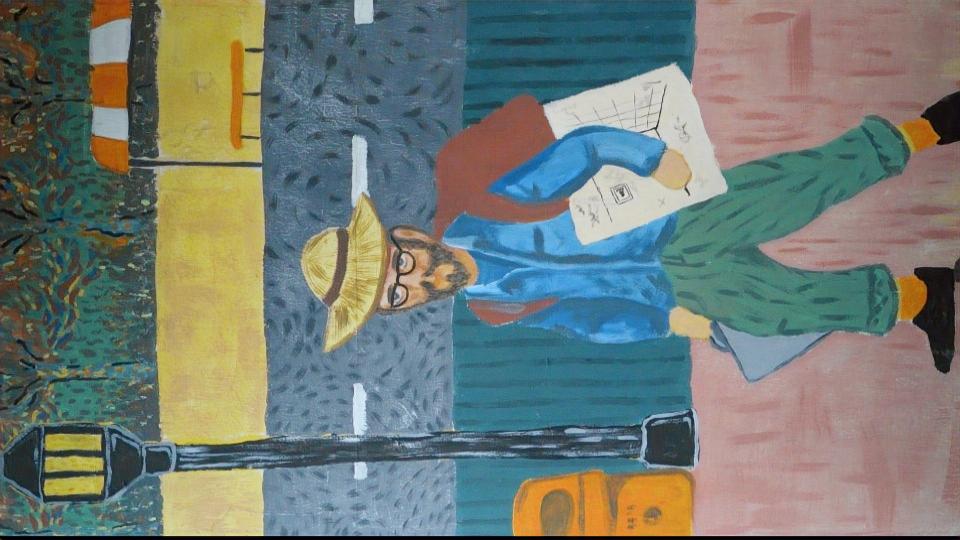} &

    \includegraphics[width=0.25\linewidth]{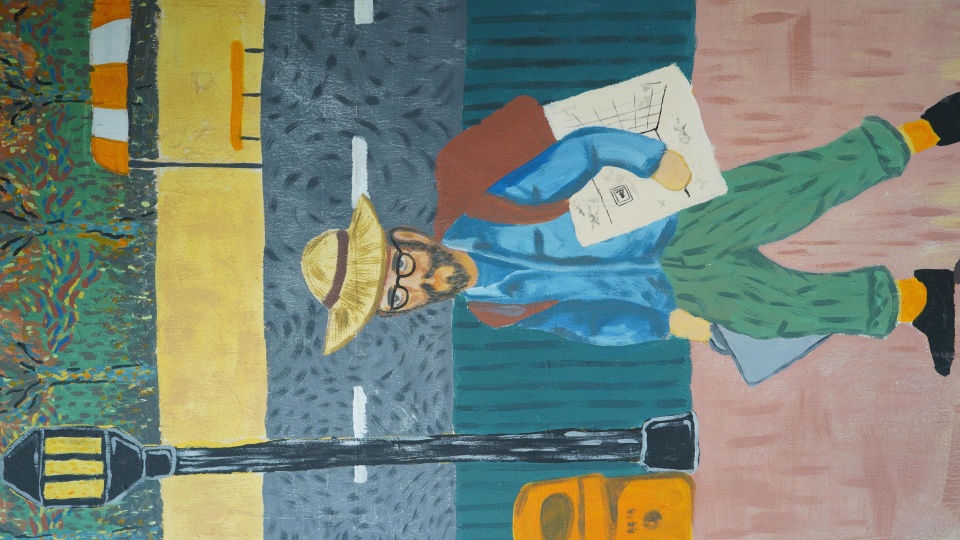} \\

    \includegraphics[width=0.25\linewidth]{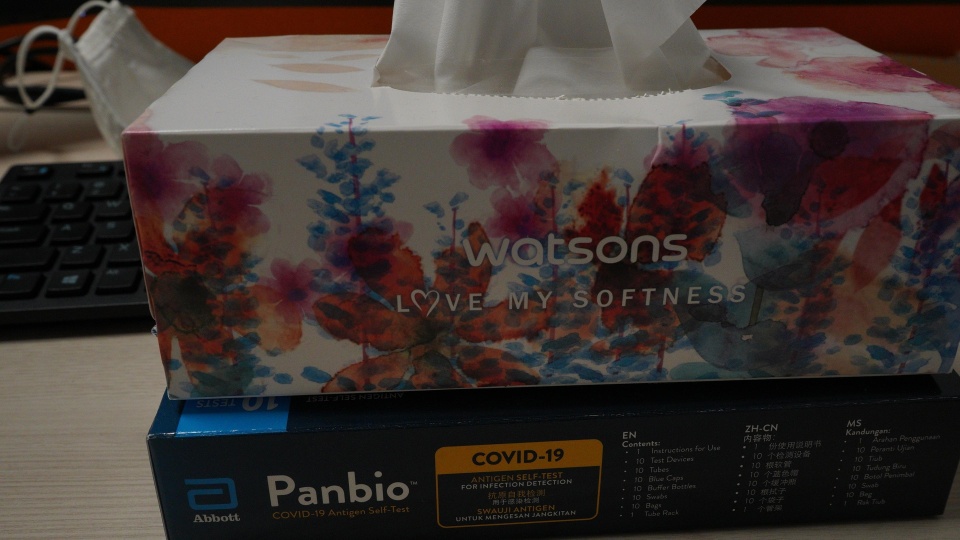} &

    \includegraphics[width=0.25\linewidth]{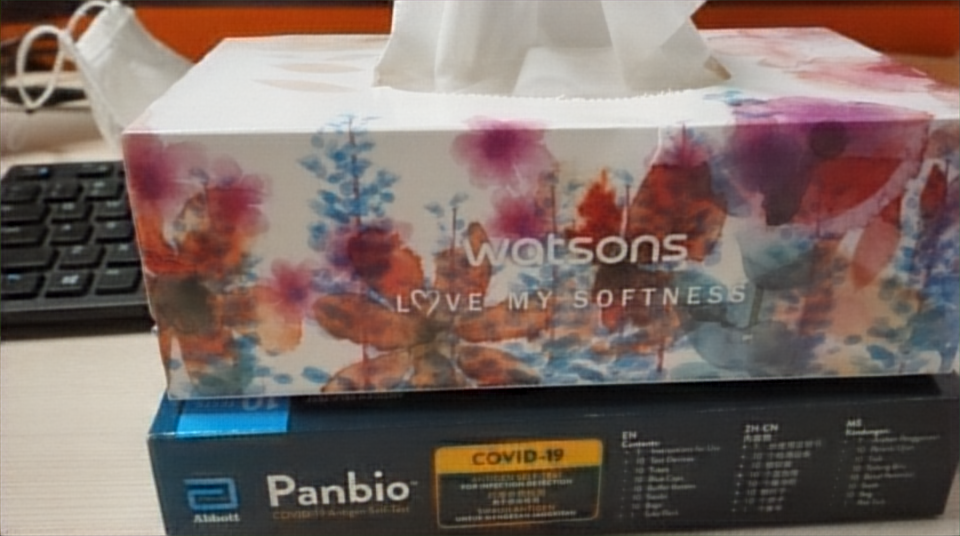} &

    \includegraphics[width=0.25\linewidth]{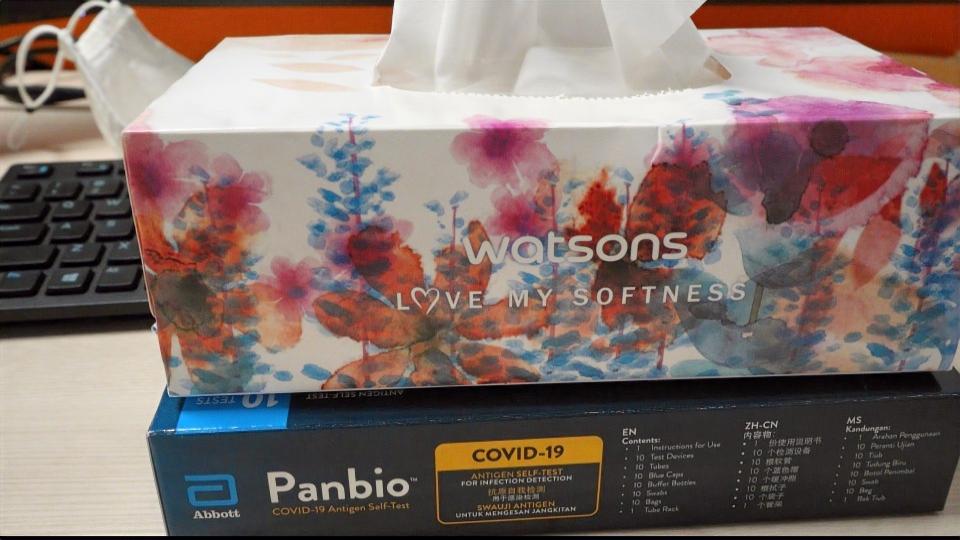} &

    \includegraphics[width=0.25\linewidth]{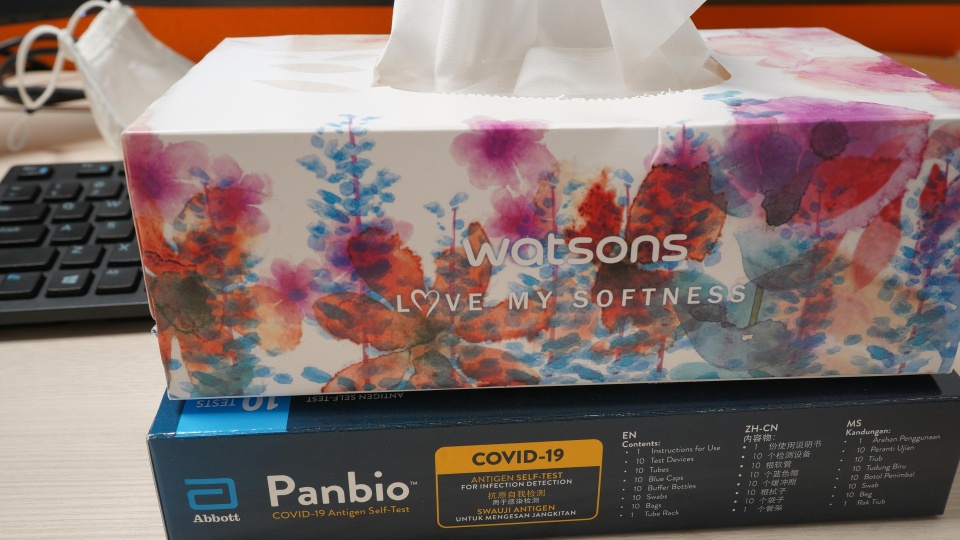} \\

    \includegraphics[width=0.25\linewidth]{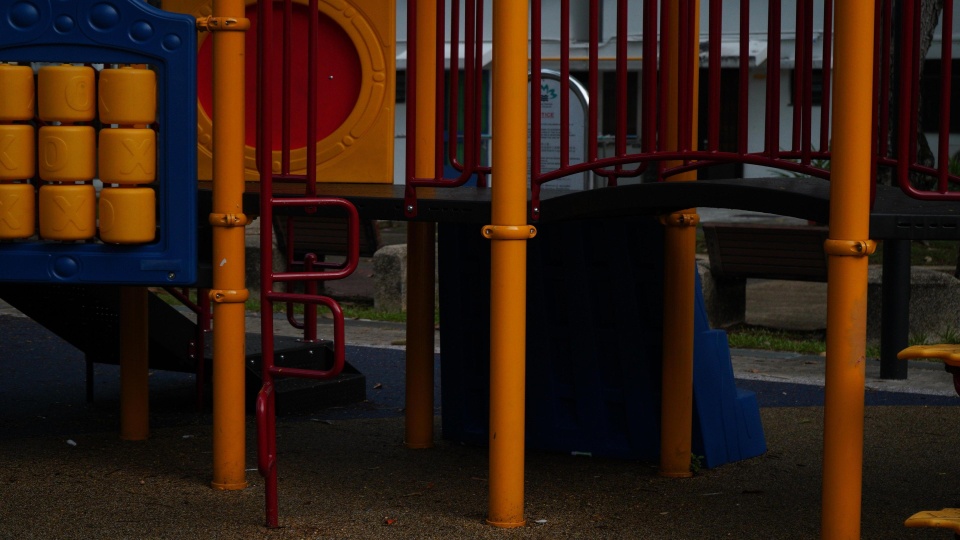} &

    \includegraphics[width=0.25\linewidth]{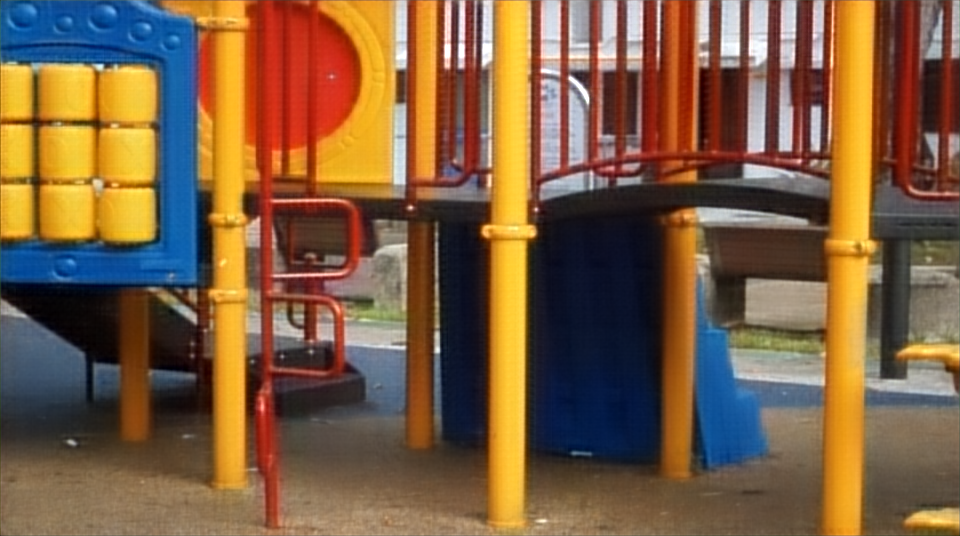} &

    \includegraphics[width=0.25\linewidth]{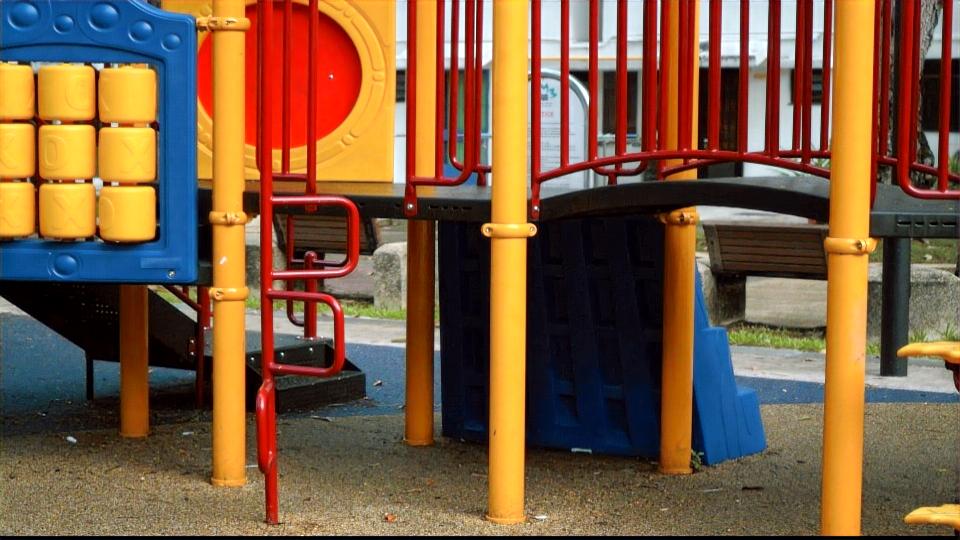} &

    \includegraphics[width=0.25\linewidth]{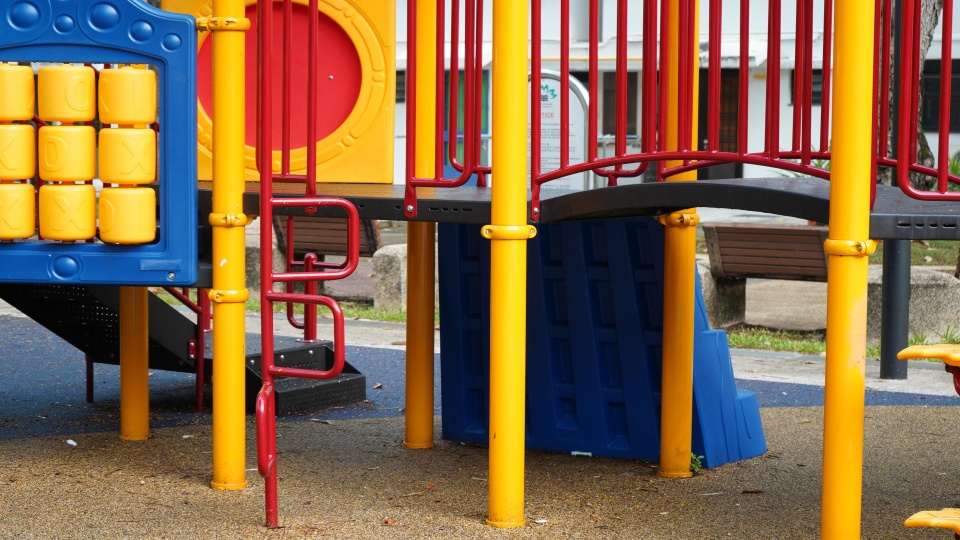}  \\
    
    \includegraphics[width=0.25\linewidth]{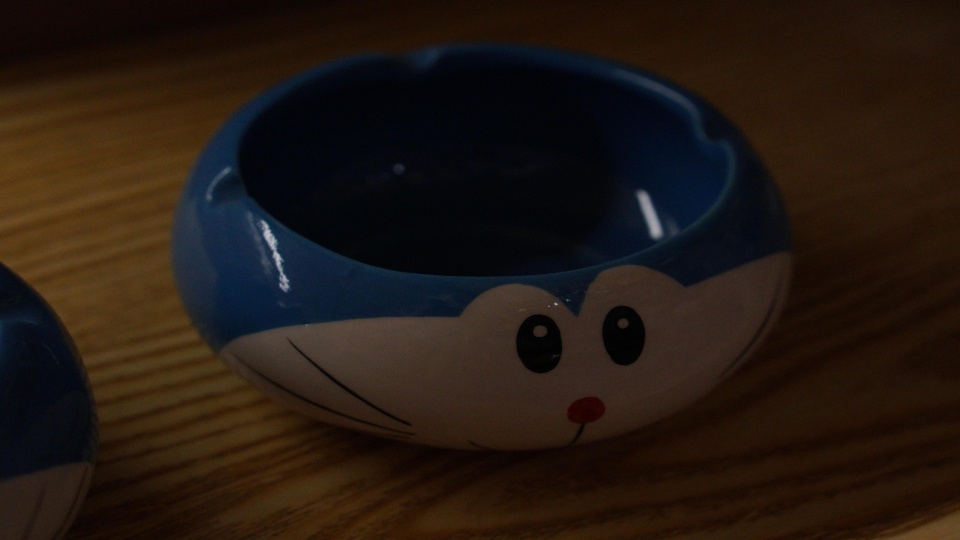} &

    \includegraphics[width=0.25\linewidth]{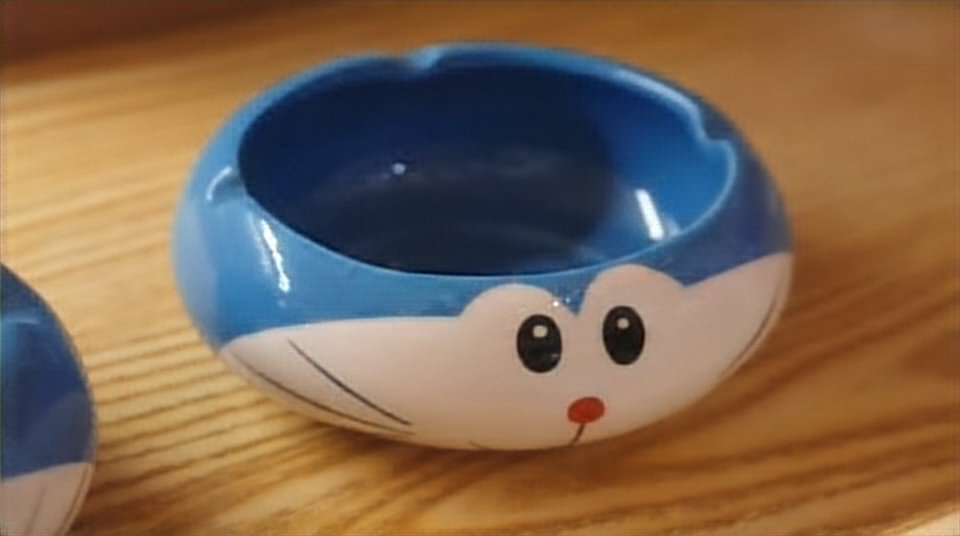} &

    \includegraphics[width=0.25\linewidth]{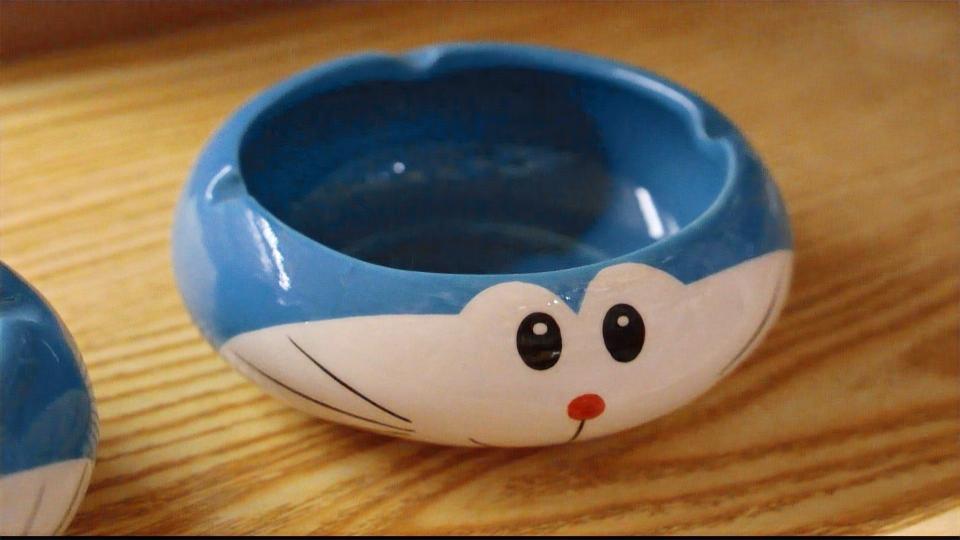} &

    \includegraphics[width=0.25\linewidth]{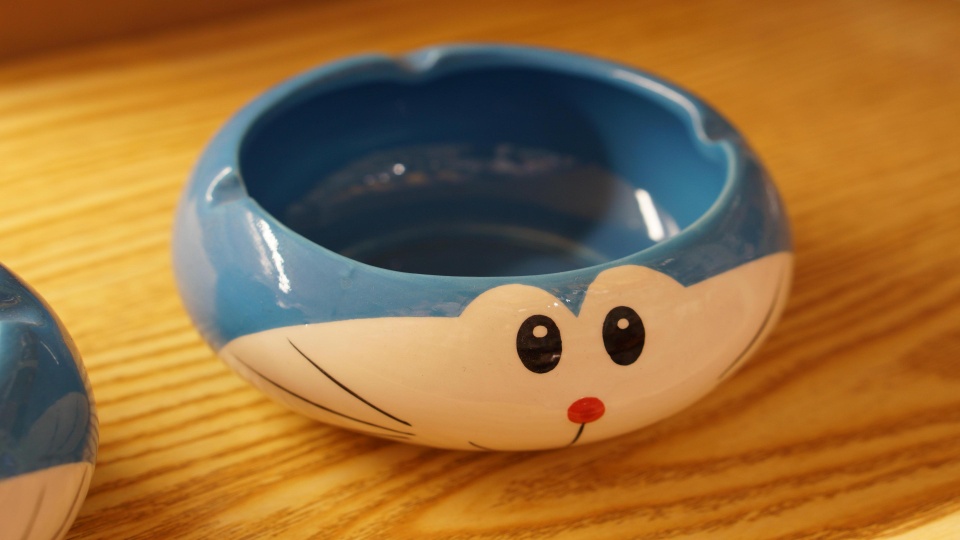}  \\

    \includegraphics[width=0.25\linewidth]{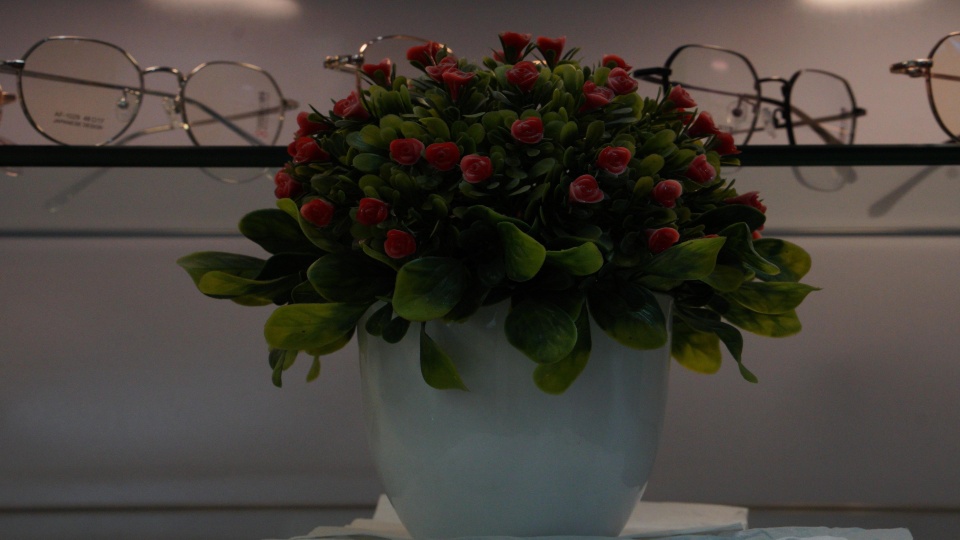} &

    \includegraphics[width=0.25\linewidth]{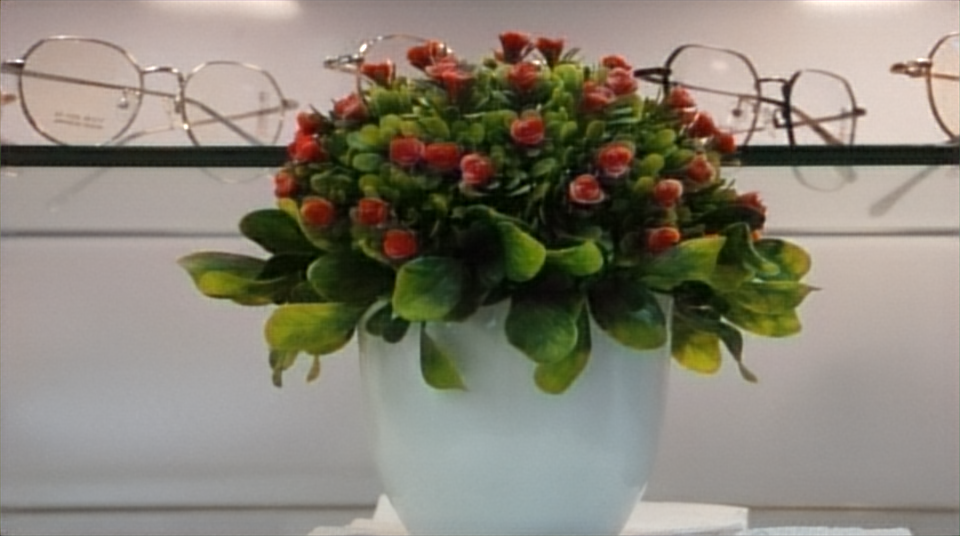} &

    \includegraphics[width=0.25\linewidth]{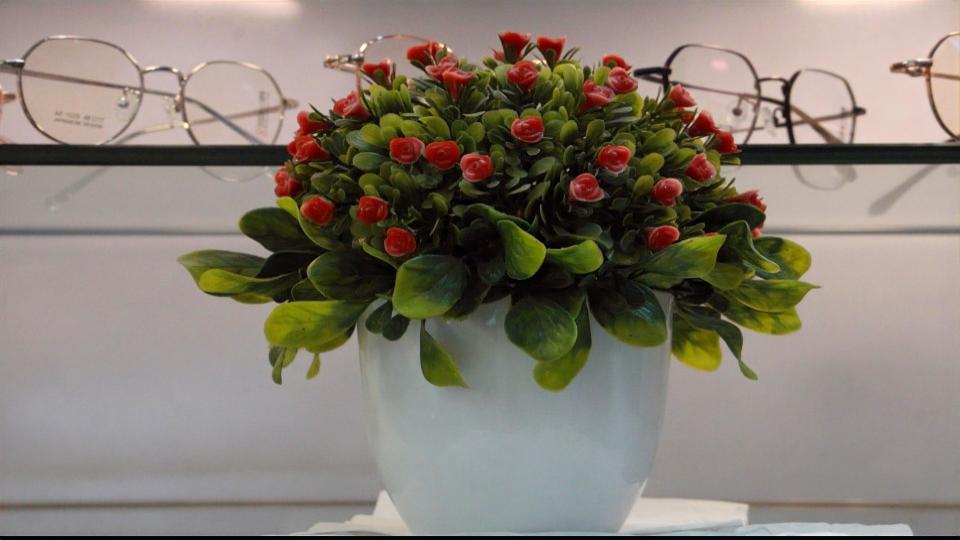} &

    \includegraphics[width=0.25\linewidth]{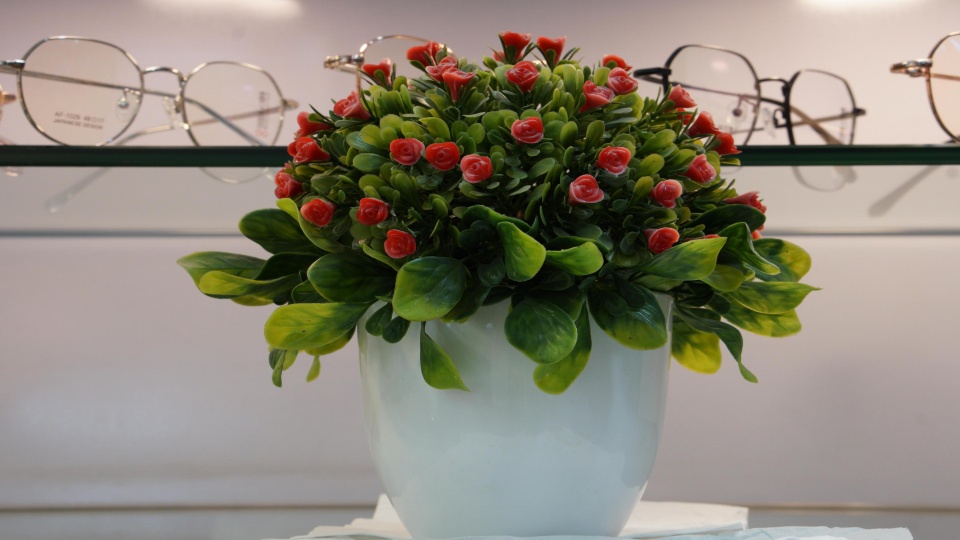}  \\

    \includegraphics[width=0.25\linewidth]{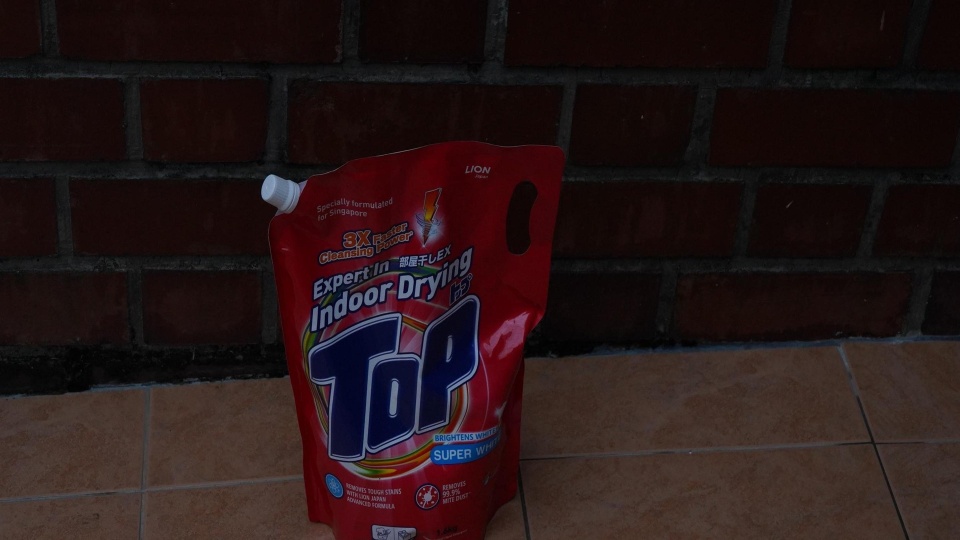} &

    \includegraphics[width=0.25\linewidth]{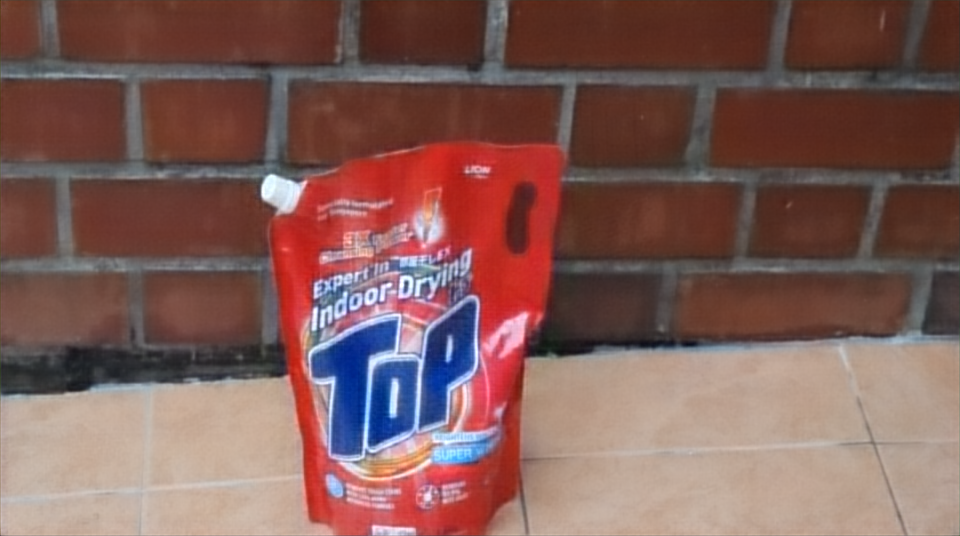} &

    \includegraphics[width=0.25\linewidth]{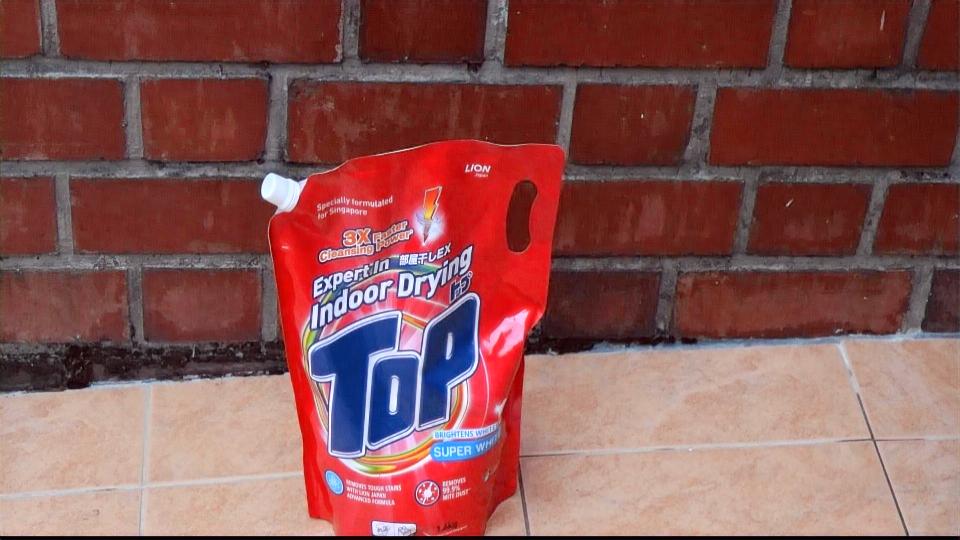} &

    \includegraphics[width=0.25\linewidth]{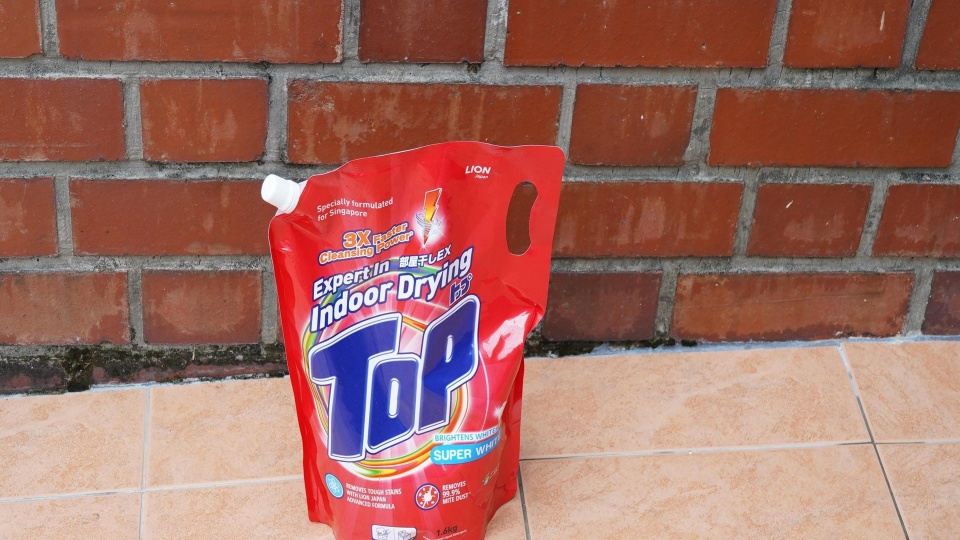}  \\

    \includegraphics[width=0.25\linewidth]{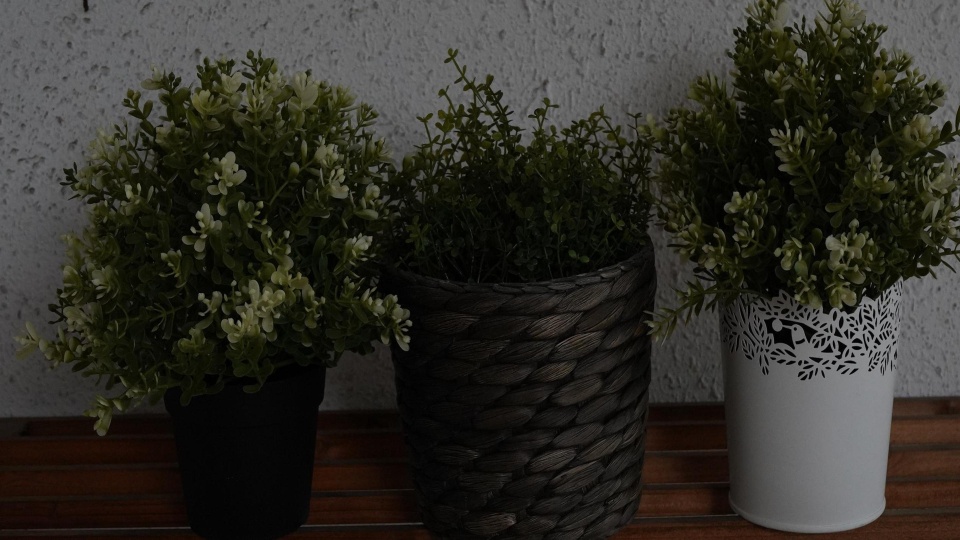} &

    \includegraphics[width=0.25\linewidth]{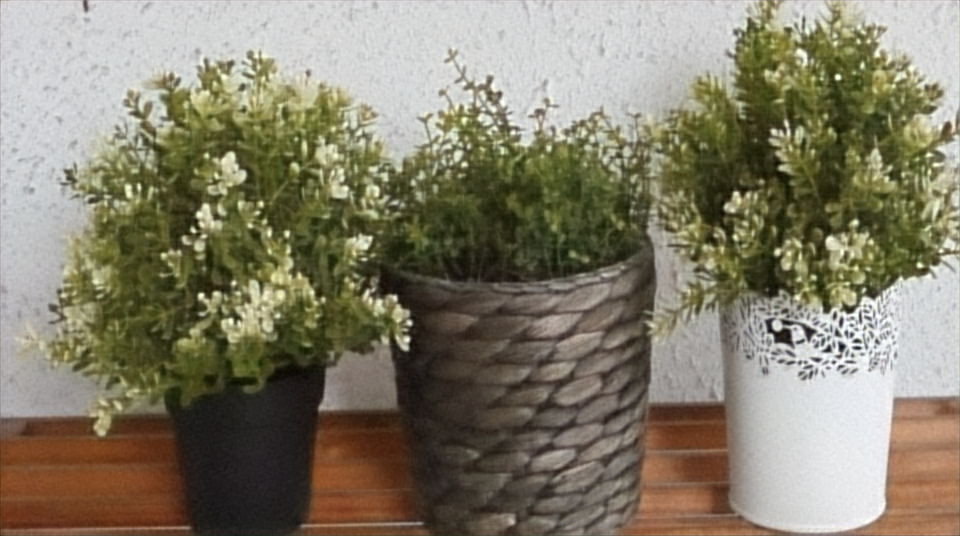} &

    \includegraphics[width=0.25\linewidth]{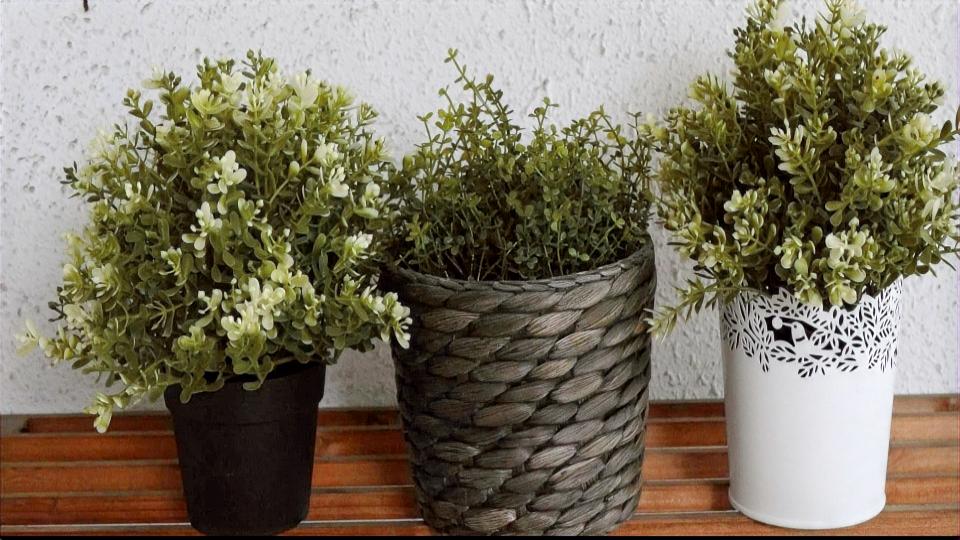} &

    \includegraphics[width=0.25\linewidth]{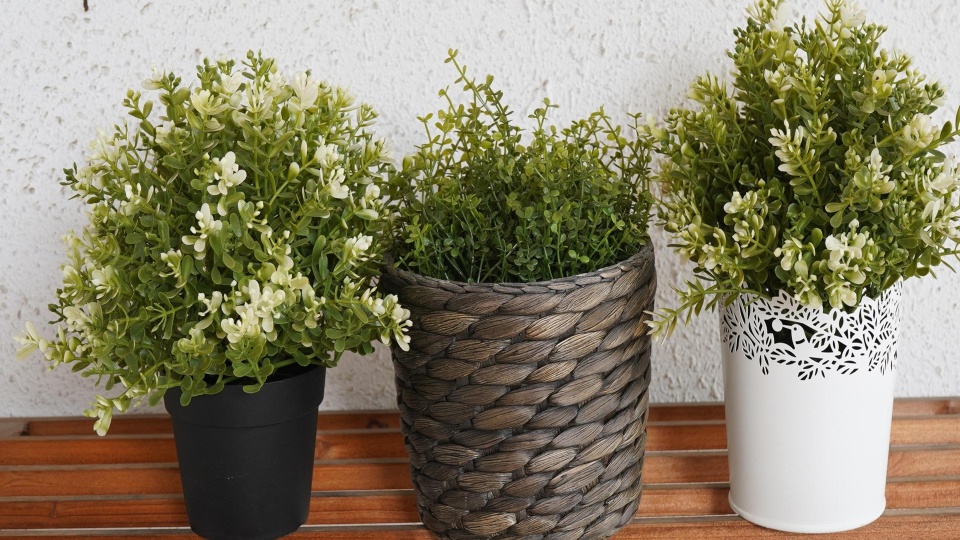}  \\

    Input & UHDFour~\cite{uhdll} & \bf{FLOL}  & Ground Truth \\

    \end{tabular}
    \caption{Qualitative results on \textbf{UHD-LL} dataset~\cite{uhdll} with images of $960\times540$. We compare the performance of our solution \textbf{FLOL} with UHDFour~\cite{uhdll}. Zoom in for best view.}
    \label{fig:uhd2}
\end{figure*}



{
\bibliographystyle{IEEEtran}
\bibliography{main}
}

\end{document}